\documentclass{article}
\pdfoutput=1

\usepackage{PRIMEarxiv}

\usepackage[utf8]{inputenc} 
\usepackage{algorithm}  
\usepackage{algpseudocode}  
\usepackage[T1]{fontenc}    
\usepackage{hyperref}       
\usepackage{url}            
\usepackage{booktabs}       
\usepackage{amsfonts}       
\usepackage{nicefrac}       
\usepackage{microtype}      
\usepackage{lipsum}
\usepackage{fancyhdr}       
\usepackage{graphicx}       
\usepackage{amsmath}
\usepackage{changepage}
\usepackage{color}
\graphicspath{{media/}}     

\title{LLM-PySC2: Starcraft II learning environment for Large Language Models
\thanks{Corresponding author.}
\thanks{Code is available at https://github.com/NKAI-Decision-Team/LLM-PySC2}}

\author{
  $\textbf{Zongyuan Li}^{1}, \textbf{Yanan Ni}^{2}, \textbf{Runnan Qi}^{2},\textbf{Lumin Jiang}^{2},\textbf{Chang Lu}^{1}, \textbf{Xiaojie Xu}^{1},$\\
  $\textbf{Xiangbei Liu}^{1}, \textbf{Pengfei Li}^{1}, \textbf{Yunzheng Guo}^{1}, \textbf{Zhe Ma}^{1}, \textbf{Huanyu Li}^{1}, \textbf{Hui Wu}^{1}$ \\
  $\textbf{Xian Guo}^{1,*}, \textbf{Kuihua Huang}^{2,*}, \textbf{Xuebo Zhang}^{1,*}$ \\
  1 College of Artificial Intelligence, Nankai University\\
  2 Laboratory for Big Data and Decision, National University of Defense Technology\\
}

\begin{document}
\maketitle

\begin{adjustwidth}{0.5cm}{0.5cm}
\begin{abstract}
\vspace{0.4cm}
The tremendous potential has been demonstrated by large language models (LLMs) in intelligent decision-making problems, with unprecedented capabilities shown across diverse applications ranging from gaming AI systems to complex strategic planning frameworks. 
However, the StarCraft II platform, which has been widely adopted for validating decision-making algorithms in the past decade, has not yet provided substantial support for this emerging domain. 
To address issues that LLMs cannot interface with the hundreds of actions of the pysc2 backend and the lack of native support for multi-agent (MA) collaboration, we propose the LLM-PySC2 environment. 
This is the first environment that offers LLMs the complete pysc2 action space with sufficient multi-modal information and game Wiki knowledge. 
With an asynchronous query architecture, the environment efficiently interacts with LLMs that maintain a constant latency regardless of the scale of the agents' population.
In the experiments, we evaluated LLMs' decision-making performance in both the macro-decision and micro-operation scenarios, with traditional StarCraft II Multi-Agent Challenge (SMAC) tasks and a series of new proposed. 
Results indicate that LLMs possess the potential to achieve victories in complex scenarios but cannot constantly generate correct decisions, especially in the recovered pysc2 action space and MA settings. 
Without task-relevant instructions, the pre-trained models suffer from issues such as hallucinations and inefficient collaboration. 
Our findings suggest that StarCraft II still challenges in the era of large models, revealing that there is a lot to do to develop an advanced LLM decision-making system, and the proposed LLM-PySC2 environment will support future development of LLM-based decision-making solutions.

\end{abstract}
\end{adjustwidth}
\vspace{0.2cm}



\section{Introduction}

The remarkable progress of large language models (LLMs) has not only enhanced their reasoning capabilities but also positioned them as multitask strategists, even without post-training on specialized domains.
Unlike reinforcement learning-based decision-making agents, LLMs exhibit advantages in better context understanding, knowledge utilization, and human-AI interactions, acting in a wider range of zero-shot scenarios like gaming
\cite{tan2024towards}-\cite{li2025parallelized}
, robot manipulation/navigation
\cite{liu2024enhancing}-\cite{dorbala2023embodied}
, financial and trading
\cite{xiao2024tradingagents}-\cite{li2023large}

However, there is still a lot to do to release the potential of LLM-based decision systems. Current works are mostly limited to prompt engineering
\cite{2023arXiv231211865M}\cite{li2025hierarchical}
, LLM workflow
\cite{ChatDev}\cite{xiao2024tradingagents}\cite{zeng2023flowmind} 
to dismantle tasks into smaller tasks, and reflection
\cite{hua2024gametheoreticllmagentworkflow}\cite{2025arXiv250213388}\cite{MCghost}\cite{xu2024llm4workflow}
to correct the previous policy. These works enable LLMs to act better in diverse scenarios, but the knowledge-learning problem for a specific domain remains unsolved.

\begin{figure}[h]
  \centering
  \includegraphics[width=0.9\textwidth]{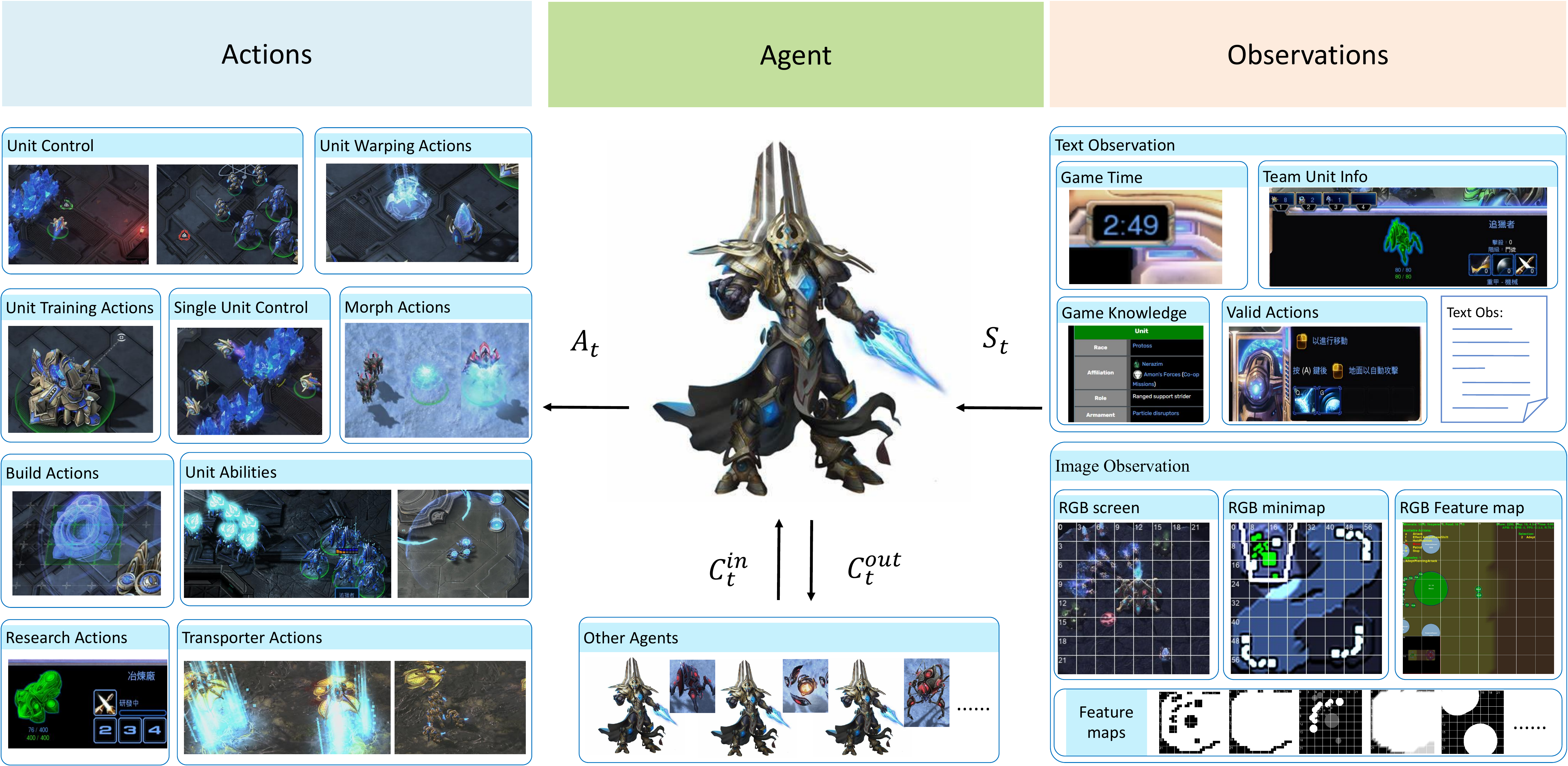}
  \caption{\textbf{Contributions of LLM-PySC2 environment.} LLM-PySC2 is the first LLM decision-making platform that supports the complete pysc2 action space. With multi-modal observation and a native multi-agent system, this environment provides supports for researches such as LLM-based planning, learning and multi-modal information processing, with enough complexity in evaluation scenatios.}
  \label{fig1}
  \vspace{-0.5cm}
\end{figure}

Currently, most LLM decision-making solutions are developed in relatively simple environments, resulting in ignorance of LLM's shortcomings. For example, MineDojo\cite{fan2022minedojo} is relatively comprehensible for LLMs and exhibits a high tolerance for errors, while some other works oversimplify the policy space of the environment\cite{shao2024swarmbrainembodiedagentrealtime}\cite{2023arXiv231211865M}.
Some earlier works, such as StanfordTown\cite{park2023generative}, do not even concern decision-making ability but is more focused on LLMs' behaviors.  

The StarCraft II environment, well known for its complexity, has been widely used as a validation platform for decision algorithms in the past decade, supported multi-agent research such as VDN\cite{sunehag2017valuedecompositionnetworkscooperativemultiagent}, Qmix\cite{rashid2018qmixmonotonicvaluefunction}, MAPPO\cite{yu2022surprisingeffectivenessppocooperative}, the milestone algorithm Alpha-Star\cite{arulkumaran2019alphastar} and DI-Star\cite{contributors2021distar}. It is precisely the extremely high complexity that makes it the most authoritative verification platform for decision-making algorithms. However, since the vector interfaces are not compatible with LLMs, the StarCraft II environment does not support complete interactions with LLMs in the past few years.


Existing LLM Starcraft II environments, such as Swarm Brain\cite{shao2024swarmbrainembodiedagentrealtime}, TextStarCraft II \cite{2023arXiv231211865M}, have the problem of severely limiting observation space and action space. 
They cut off most unit control operations and reduce continuous action space to discrete. 
Although the over-simplified environments attracted attention for LLM decision research in the past years, they hindered further research due to the lack of complexity and incomplete support of refined operations.
Other works like \cite{ma2025vlms}\cite{ma2025smachard} do not support complete games.

At the same time, the support of current platforms for multi-agent systems is insufficient. 
Currently, most LLM multi-agent systems\cite{li2025parallelized}\cite{xiao2024tradingagents} expose only a single agent to interact with the environment, while others act as modules for data processing or aggregating.
Other works focus on conducting social simulations, emphasizing the accuracy of simulation
\cite{tang2024gensim}-\cite{gurcan2024llmaugmentedagentbasedmodellingsocial} 
rather than promoting multi-agent collaboration.

To provide support for LLM decision-making, we developed \textit{LLM-PySC2}, an environment derived from the StarCraft II Learning Environment (SC2LE)\cite{vinyals2017starcraft}.  
As shown in Fig.~\ref{fig1}, this environment expanded the action space to complete pysc2 action space, allowing agents to perform fine-grained operations and unit skills. 
We also provide agents with comprehensive observations, including images and Wiki Knowledge\cite{starcraftwiki}. 




It is worth noting that this is a platform with native multi-agent framework. 
We enable all kinds of multi-agent cooperation such as centralized decision-making and distributed decision-making.
To avoid an increase in waiting time as the number of agents grows, we build an asynchronous query architecture to maintain the latency of multi-agent queries. 

In experiments, eight new scenarios were proposed. Unlike the SMAC\cite{samvelyan2019smac} tasks, these tasks require more on task understanding and usage of unit skills. Mainstream LLMs are evaluated in both the complete StarCraft II games and mini scenarios. Results indicate that pre-trained LLMs \textit{have} possess zero-shot decision-making ability but \textit{lack the ability} to make consistently effective decisions. Without task-specific training, pre-trained LLMs cannot always find the key elements for victories. They fail to identify the important aspect of the situation, making mistakes in analysis and even dealing damage to allies sometimes.

Our contributions can be concluded as follows:

\qquad(1) 
We propose the first LLM StarCraft II framework with a complete pysc2 action space and provide a structured Wiki knowledge database of all units' information.

\qquad(2) 
We provide native support for multi-agent collaboration in our platform, paired with an asynchronous architecture that ensures a stable latency regardless of the population of LLM agents.

\qquad(3) 
We propose several new evaluation scenarios for LLM decision-making and evaluate LLMs' performance in both the macro-decision scenarios and the scenarios for micro-operations.

Problems of the LLM decision system are also discussed in the final sections of our paper.
Results indicate that current LLMs cannot effectively handle complex StarCraft II scenarios due to serious hallucinations and lack of domain knowledge. 
How to increase the ability of LLMs in complex decision-making problems, at an acceptable cost, still poses a challenge in the era of large models and remains an unsolved problem.
\section{LLM-PySC2 environment}

\begin{figure}[t]
  \centering
  \vspace{-0.20cm}
  \includegraphics[width=0.90\textwidth]{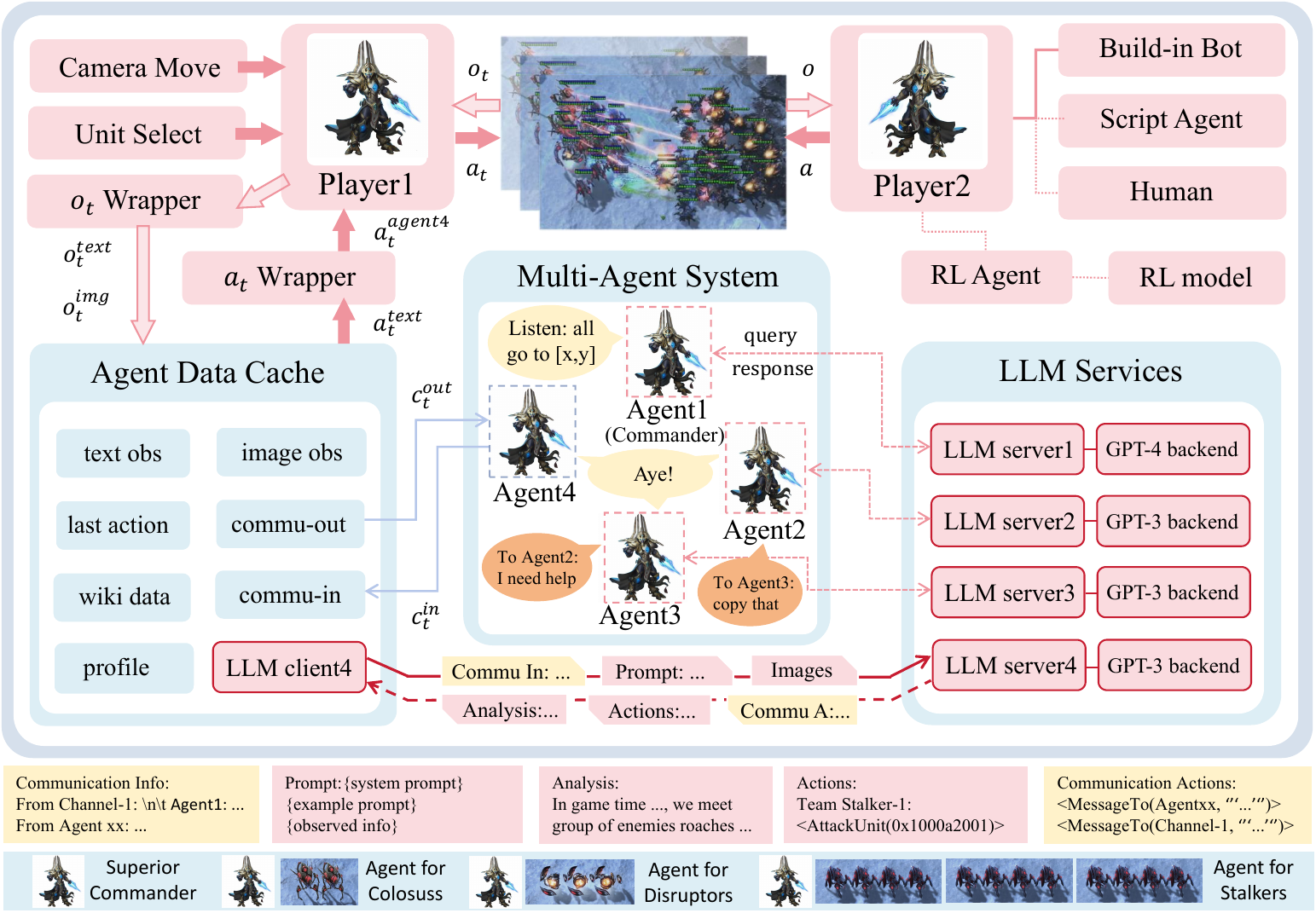}
  \caption{\textbf{LLM-PySC2 framwork.} In LLM-PySC2, the original PySC2 observation will transform into a text-form or multi-modal observation. LLM-generated text action can be recognized and transformed into PySC2 functions, enabling LLMs to interact with the StarCraft II environment and control the units.}
  \label{fig:2}
  \vspace{-0.30cm}
\end{figure}

\subsection{Framework}
The LLM-PySC2 environment is built on the player level of  SC2LE. As shown in Figure \ref{fig:2}, two players fight against each other and play the role of interacting with the pysc2 backend. They directly control the camera, select units, collect observations, and execute actions. 

To precisely control the whole system, a multi-agent framework is designed.
Agents of the system collaborate through natural language communication. At each step time $t$, agent $i$ with profile $p^i$ get the observations $o_t^i$ from the environment, queries remote LLM for analysis $ana_t^{i}$ and strategy $stg_t^{i}$, communication messages $m^{i}_{t}$ and actions $a^{i}_{t}$:
$$(ana_t^{i}, stg_t^{i}, m^{i}_{t}, a^{i}_{t}) = LLM(p^i, o_t^i)$$
Then the player sends the joint action to the environment and transmits messages to assigned agents:
$$(o_{t+1}^{1}, o_{t+1}^{2},\ ...\ , o_{t+1}^{n}) = Env(a^{1}_{t}, a^{2}_{t},\ ...\ ,  a^{n}_{t}; m^{1}_{t}, m^{2}_{t},\ ...\ ,  m^{n}_{t})$$
Pseudo code of the interaction and query process can be seen in Appendix A.


\color{black}



\subsection{Actions}

Actions are the most important part of a decision-making problem.
In LLM-PySC2, textual actions play the role of the interface for large models and the environment. These actions are defined as:
$$<ActionName(args)>$$
where args refer to screen coordinates, minimap coordinates, unit tag or their combination. 
Compared to discrete text actions, these actions avoid clipping the policy space and neglecting StarCraft II complexity.


\begin{figure}[h]
  \centering
  \vspace{-0.1cm}
  \includegraphics[width=0.90\textwidth]{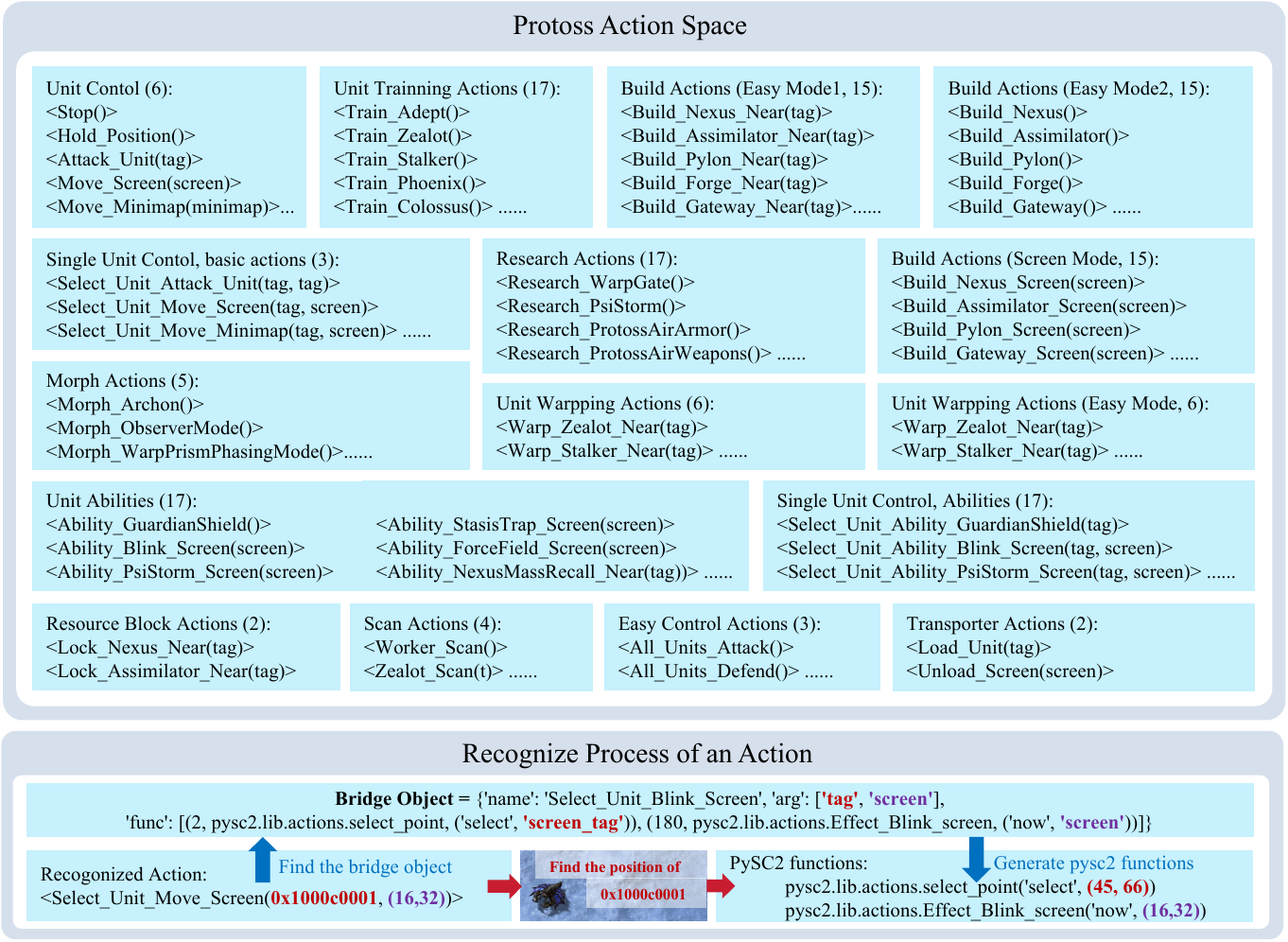}
  \caption{\textbf{Protoss action space and the recognition process.} LLM-PySC2 is the first LLM decision-making environment with complete pysc2 action space. LLM controls units by output actions in the shape of <Action\_Name(args)>. The environment transforms text action into pysc2 functions according to a transform protocol and the relevant bridge object of the action.}
  \label{fig:actions}
  \vspace{-0.2cm}
\end{figure}

\subsubsection{Action Space.}

In SC2LE, there are about 500 original functions for controlling Protoss, Terran, and Zerg. Most of them require additional parameters such as a screen or minimap position. These actions further constitute a huge policy space, making StarCraft II one of the most complex environments for decision-making problems. 


As shown in Fig~\ref{fig:actions}, there are more than 100 text actions for Protoss agents in the LLM-PySC2 environment, which can be classified as unit control, unit skills, building, researching, training, etc. Different from other environments, these actions increase the theoretical performance of optimal policy but also raise the challenge of generating correct actions. More details of the action space can be seen in Appendix B.



\subsubsection{Action Recognition.}

LLMs interact with the environment by generating text actions. After receiving text actions from LLMs, the environment first recognizes valid actions through regular expressions, searching for segments that shape as $<ActionName(args)>$. 



To establish the relationship between textual actions and pysc2 functions, we developed a protocol for text action recognition. This protocol relies on a series of bridge objects that encapsulate both the textual representation and the callback form of the actions, along with the association of action and function arguments. After determining which actions to execute, the LLM-PySC2 environment generates pysc2 functions and sequentially executes these functions in the backend.






\subsection{Observation}

Observation provides fundamental support for decision-making. Given the distinct requirements of different agents, we developed an interface that offers each agent the observations specifically suited to their tasks.
Additionally, with multi-modal observations that convey rich semantic and visual information, we released the potential for a deeper understanding of the situation, solving the problem that the previous environment had only observation of unit quantity information.



\begin{figure}[h]
  \centering
  \includegraphics[width=0.95\textwidth]{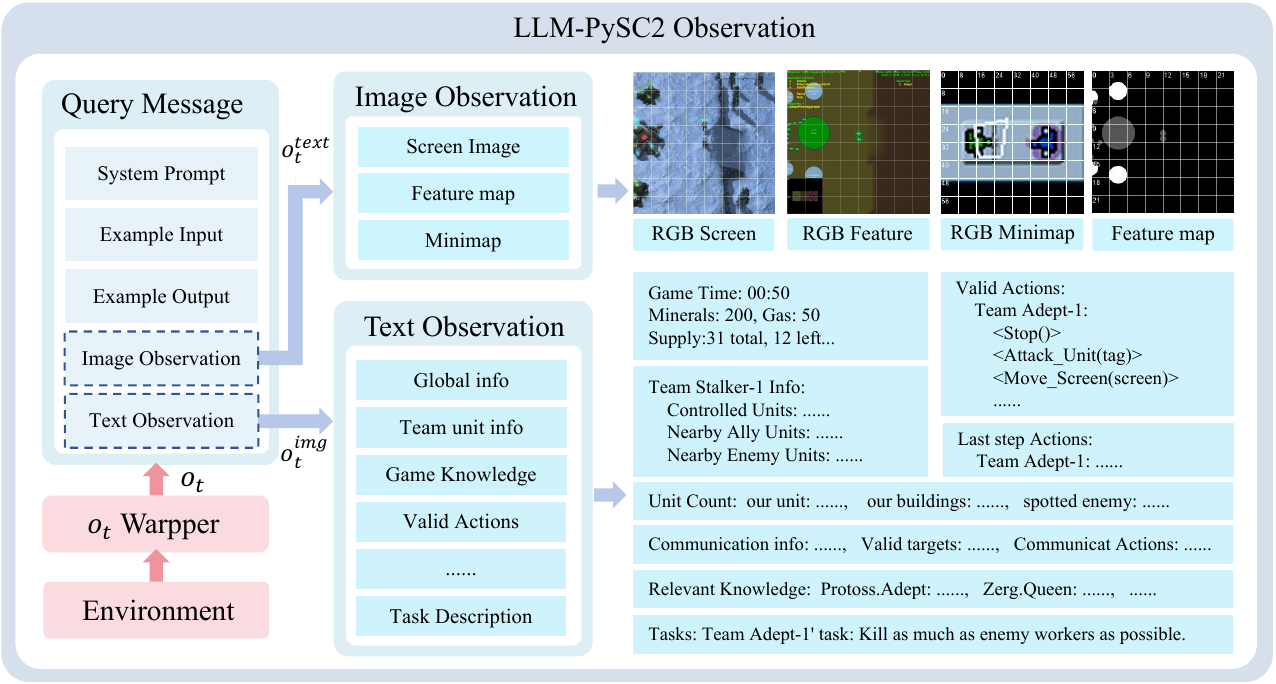}
  \caption{\textbf{LLM-PySC2 observations.} LLM-PySC2 provides multi-modal observation. The observation wrapper generates text and image observations that contain all the important information for decision-making, with access to images of the screen, minimap, and all the pysc2 original feature maps. 
  }
  \vspace{-0.2cm}
  \label{fig:obs}
\end{figure}

\subsubsection{Text Observation.}

As illustrated in Fig.~\ref{fig:obs}, an observation wrapper is implemented to process the relevant text observations for each agent. This wrapper includes a set of functions tailored for handling different types of text information. Once all parts of the text observation are generated, they will be aggregated into OpenAI query messages, which include system prompts, example inputs and outputs, as well as a series of images. These messages are then sent to the LLM server for querying responses.



As an example, we provide functions that encapsulate the following information: (1) Global game information; (2) Unit counts; (3) Screen units information (with each unit's position, health, status); (4) relevant Wiki knowledge of nearby units; (5) important event of last step; (6) valid actions for current step; (7) action explanation; (8) last steps actions; (9) errors of last step actions; (10) received communication message; (11) valid communicate targets and actions; (12) task information.

Considering possible user requirements, we expose all the observation interfaces in the open-source code repository. It is possible to customize the wrapper to generate other kinds of text observations. More detailed examples of text observation can be seen in Appendix C.

\subsubsection{Image Observation.}

In StarCraft II scenarios, textual observations limit agents' comprehension of terrain, relative position, or other aspects. It is almost inevitable to use images, even videos, to describe such higher-dimensional information. 

In LLM-PySC2, we provide four kinds of image observation: RGB-Screen, RGB-Minimap, RGB-Feature, and Original-Feature-Maps. Image observation wrapping functions collect the image from the pysc2 backend, adding auxiliary lines and annotations to facilitate the coordinate recognition by LLMs. The image will be encoded into a base 64 string and will be added to the message to query the LLMs for analysis, actions, and communication behaviors.





\subsection{Multi-Agent System}

Disassembling complex problems into small tasks in a multi-agent system has become a basic solution. Different large models interact through natural language, coordinating their behaviors and managing the massive StarCraft II system together.

\begin{figure}[h]
  \centering
  \includegraphics[width=0.9\textwidth]{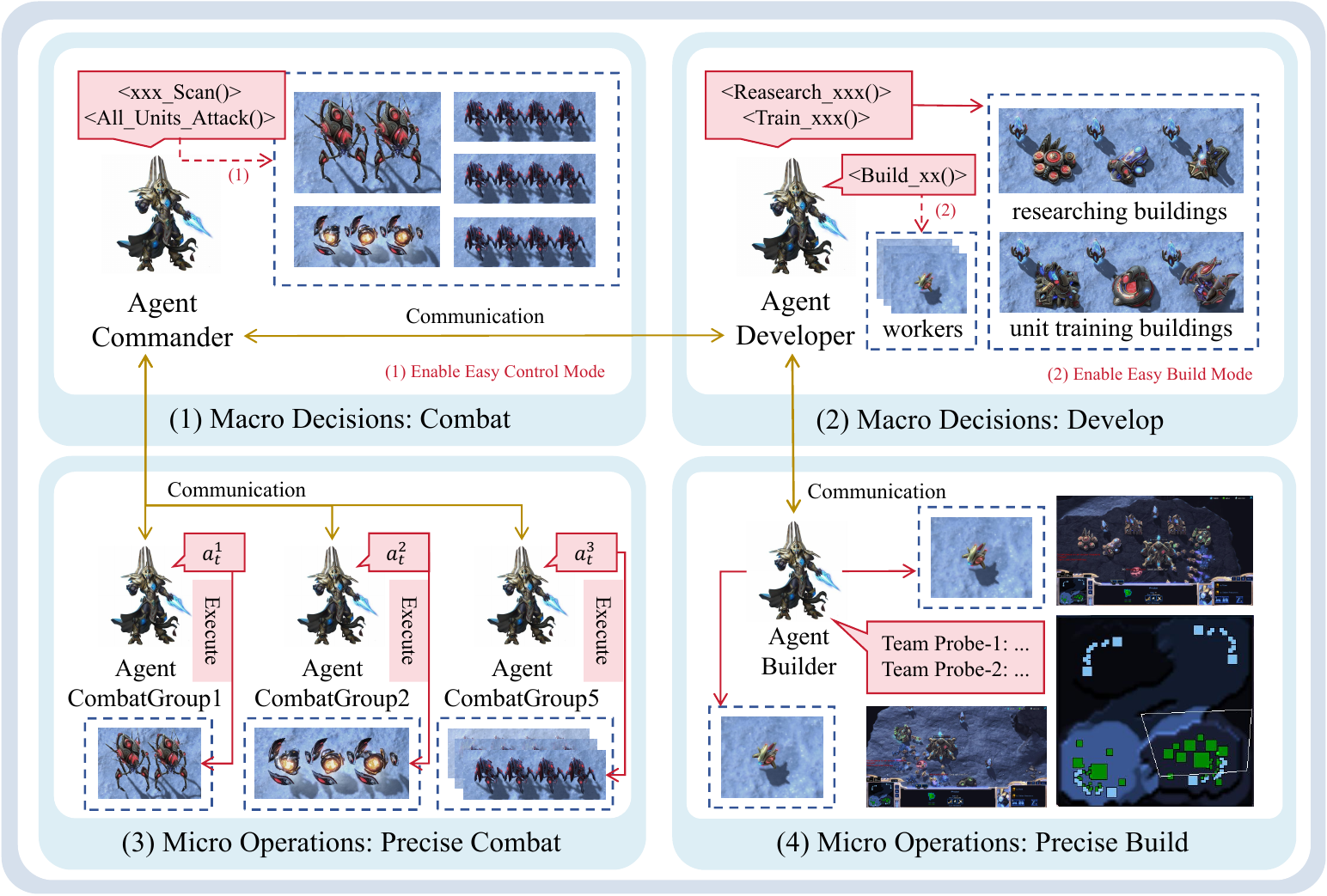}
  \caption{\textbf{LLM-PySC2 multi-agent system.} In LLM-PySC2, game control is divided into combat((1), (3)) and development((2), (4)). In standard unit control mode, the agent Commander sends messages to agents named CombatGroupi, and the CombatGroup agents control their units moving, attacking, or using skills to achieve tasks assigned by superiors. In standard build mode, the agent Developer trains units, updates technologies, and asks the agent Builder to build buildings. Then the Builder controls workers and chooses positions to construct new buildings. }
  \label{fig:commu}
\end{figure}


\subsubsection{Communication}

In LLM-PySC2, agents collaborate by communicating with each other. They can discuss in a channel or directly send messages to another agent. As shown in Fig.~\ref{fig:2}, at each step, received messages will be added to observation, and the agent can respond to others by generating Communication actions shaped as
$<MessageTo(TargetName,\ '''content''')>$.
When the agent receives messages, the received messages will be displayed in their origin form. An Agent should analyze both the observed situation and the requests/information from others, and finally generate actions and reply to their teammates.


\subsubsection{Multi-Agent Settings} 

As shown in Fig.~\ref{fig:commu}, we define four kinds of agents responsible for (1) macro-decisions for combat deployment, (2) macro-decisions for economic development, (3) micro-operations for combat, and (4) micro-operations for building. Agents for macro-decisions organize other agents to work together, while agents for micro-operations execute specific actions. 
Note that, agents of LLM-PySC2 query in independent threads, ensuring a constant waiting time when the number of agents increases.

The multi-agent system supports both centralized and decentralized decisions in the environment. Two 'Easy Modes' are also provided for simplifying some aspects that researchers are not very concerned about, among which 'Easy Build' disables the agent Builder and helps researchers concentrate more on multi-agent collaboration in the combat, while 'Easy Control' disables the agents CombatGroups and helps researchers concentrate more on planning and multi-modal information processing.

\color{black}

\section{Experiments}

In this section, we introduce two series of experiments: (1) Experiments for macro-decisions, i.e. \textbf{complete StarCraft II game}; (2) Experiments for micro-operations, including classic SMAC scenarios and eight new tasks that require units to use their skills and achieve assigned goal. To distinguish micro-operation scenarios from the traditional SMAC environment, we refer to these two groups of experiments the \textbf{LLM-SMAC task group} and the \textbf{LLM-PySC2 task group}. 

Combined with the complete StarCraft II games, these experiment scenarios constitute one of the most comprehensive experiment groups in LLM decision-making and support research on enhancing LLMs’ abilities in reasoning, planning, learning, multi-modal information processing, and multi-agent cooperation.

We use the Kill/Death (KD) ratio and Winning Rate (WR) to evaluate the performance of the LLMs:
$$KD = value(killed\_units) / value(dead\_units), \ WR=num(win)/num(total) $$
The higher the value of these two indices, the better the performance of LLMs.

\subsection{Experiments for Macro-Decisions}

\subsubsection{Experiment Settings}

Complete games demand real decision-making abilities, such as analyzing situations, planning for tactic strategy, deceiving the opponent, and engaging with the enemy at the right time. To evaluate the performance of LLM macro-decisions, we tested the three modes in the Simple64 map: (1) easy control + easy build (ECEB); (2) standard control + easy build (SCEB); and (3) easy control + standard build (ECSB). 

For games with easy control settings, we enable the agent Commander to directly control all units to attack, defend, retreat, and scan for information. For the standard control settings, we enable agents named CombatGroup-i to precisely control different kinds of unit to move, attack, and use skills. 

For games with easy build settings, the agent Developer can build buildings by generating actions $<Build\_BuildingName()>$. For games that enable standard build, the agent Developer can only train units and upgrade technologies, and has to communicate with the agent Builder to build in specific coordinates $[x, y]$ by generating actions $<Build\_BuildingName\_Screen([x, y])>$.

In these experiments, we give each agent a client for querying GPT-4o-mini. For macro-decision agents, we provide relevant text information and minimap images. For standard unit control agents, we provide text observation and both the screen image and minimap images. Examples of observations, responses for different agents, and detailed experimental settings can be seen in Appendices C and D.

\begin{figure}[h]
  \centering
  \includegraphics[width=0.9\textwidth]{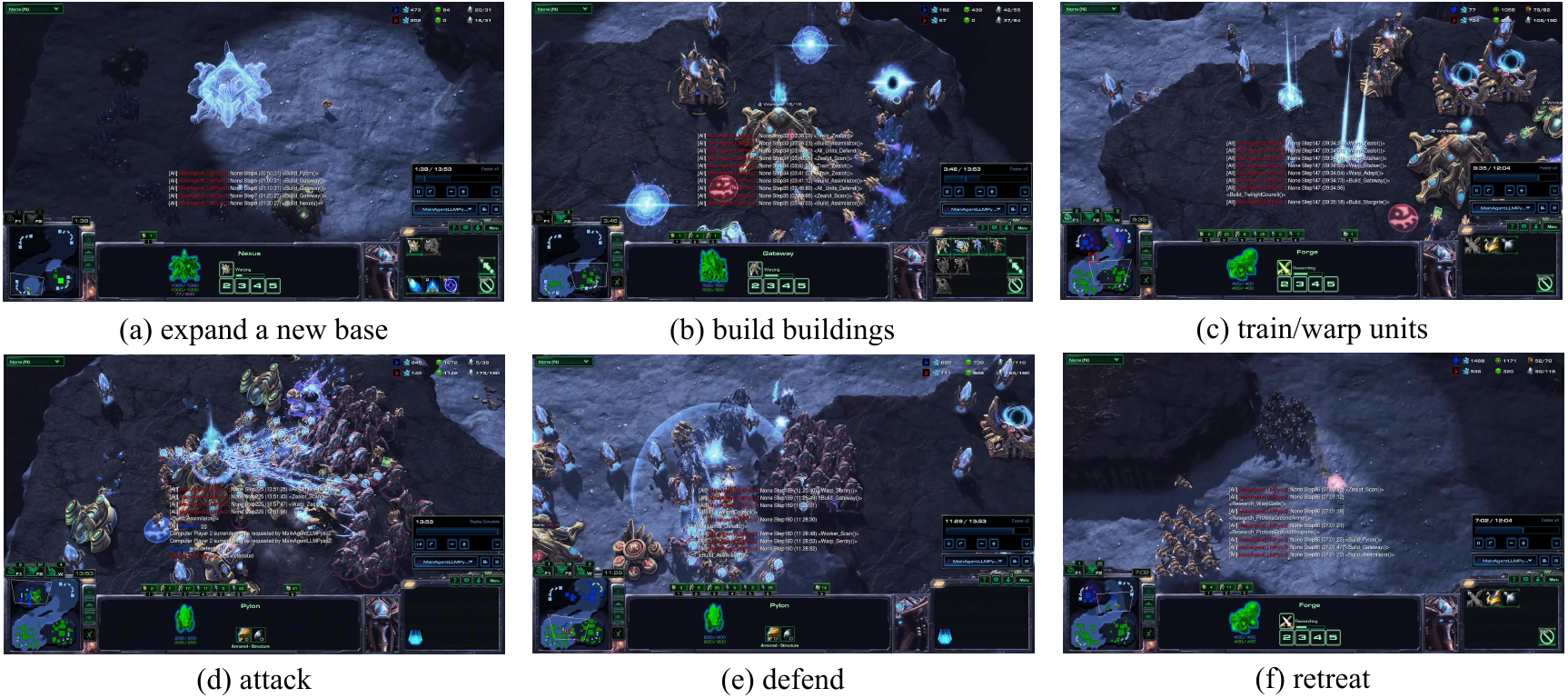}
  \caption{\textbf{StarCraft II complete game in LLM-PySC2.} StarCraft II complete game requires both the macro-decision ability and micro-operation ability. 
  The agent Developer and Builder has to (a) expand new bases, (b) build new buildings, (c)train or warp units for combat, and upgrade technologies. The agent Commander with agents for CombatGroups controls the army (d) defend, (e) attack, (f) retreat, or make complex deployment to deceive and defeat the opponent.}
  \label{fig:task1} 
\end{figure}

\subsubsection{Experiment Results}

\begin{table}[h]
\caption{Winning Rates of GPT-4o-mini in Complete StarCraft II games.}\label{tab1}
\begin{center}
\vspace{-0.2cm}
\small
\renewcommand\arraystretch{1.3}
\begin{tabular}{p{4.9cm} p{0.9cm} p{0.9cm} p{0.9cm} p{0.9cm} p{0.9cm} p{0.9cm} p{0.9cm}}
\toprule
Mode / WR & level-1 & level-2 & level-3 & level-4 & level-5 & level-6 & level-7 \\
\midrule
ECEB (Easy Control + Easy Build)      & 100\% & 100\% & 92\% & 50\% & 42\% & 8\% & 0\%\\
SCEB (Standard Control + Easy Build)  & 100\% & 50\% & 0\% & 0\% & 0\% & 0\% & 0\%\\
ECSB (Easy Control + Standard Build)  & 75\% & 17\% & 0\% & 0\% & 0\% & 0\% & 0\%\\

\bottomrule
\end{tabular}
\end{center}
\vspace{-0.2cm}
\end{table}

In the macro-operation tasks(complete SC2 games), we conducted 12 repeated experiments from level-1 (very easy) to level-7(very hard/elite). As shown in Table.~\ref{tab1}, two agents of ECEB mode control the whole system by discrete actions that perform nearly the same as the works in TextStarCraft2\cite{2023arXiv231211865M}. At level-5, LLMs can only win half of the games in both environments and nearly lost all the games at level-6.

In SCEB and ECSB modes, LLMs perform even worse due to the recovered complexity from complete action space and the higher demand for collaboration. In SCEB mode, the Easy Build part develops the economy and military strength the same as in ECEB mode, but agents for micro-operations frequently make mistakes in command, resulting in a 0\% winning rate in level-3. In ECSB mode, agent Builder takes control of workers, but the Builder frequently builds in dangerous or invalid positions that seriously undermine the development, also resulting in a 0\% winning rate in level-3. Not to mention building a defense line to resist the early attack under high difficulty.

\subsection{Experiments for Micro-Operations}

\subsubsection{Experiment Settings}

SMAC is a well-known benchmark for multi-agent reinforcement learning (MARL) approaches. 
We provide compatible support for SMAC tasks.
Note that, unlike the SMAC tasks that $n$ units are controlled by $n$ agents, the LLM-SMAC units are controlled in groups. It is not recommended to compare the LLM-based method with the MARL-based method in these tasks due to the different control frequencies.

In the LLM-PySC2 task group, eight new experimental scenarios were constructed. These tasks introduce unit skills into the experiments. 
Unlike SMAC, which focuses only on incoming combat, LLM-PySC2 requires an understanding of task description, planning attack routes, and utilizing skills to achieve the goal.
Tasks 1 to 4 are designed as single-agent tasks, while tasks 5 to 8 are designed as multi-agent tasks. Agent settings are the same as for the standard control mode of the complete StarCraft II game. More detailed settings can be seen in Appendix D.

\begin{figure}[h]
  \centering
  \includegraphics[width=0.9\textwidth]{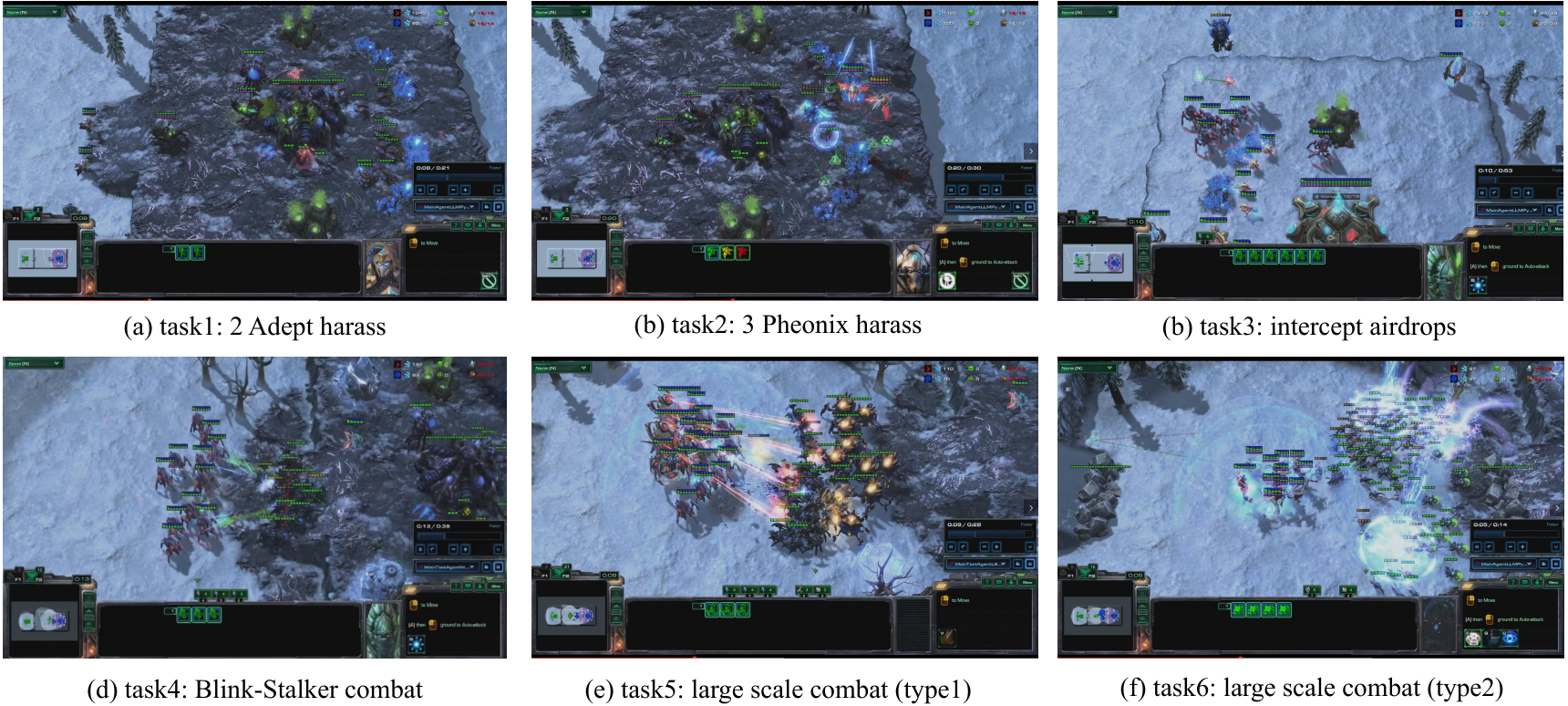}
  \caption{\textbf{Experiments for micro-operations: LLM-PySC task group.} games.
  (a) (b) Controlling 2 Adepts or 3 Pheonix to harass enemy economy, kill more than half of enemy workers; (c) (d) Controlling Stalkers to intercept incoming airdrop or defeat with enemy Roaches using Blink ability; (e) (f) Controlling a mixed combat group of several unit types, use skills especially Area-of-Damage skills to defeat enemies.
  }
  \label{fig:task2} 
\end{figure}

\subsubsection{Experiment Results}

\begin{table}[h]
\vspace{0.1cm}
\caption{Kill/Death Rates and Winning Rates of LLMs in LLM-SMAC Tasks.}\label{tab2}
\begin{center}
\vspace{-0.1cm}
\small
\renewcommand\arraystretch{1.3}
\begin{tabular}{p{2.3cm} p{1.4cm} p{1.4cm} p{1.4cm} p{1.8cm} p{1.6cm} p{1.5cm}}
\toprule
Model/KD(WR) & 2s3z & 3s5z & 1c3s5z & 3s5z\_vs\_3s6z & 2c\_vs\_64zg & 3s\_vs\_3z \\
\midrule
gpt-3.5-turbo   & 0.60 (22\%) & 0.43 (\textbf{4}\%) & 0.91 (44\%) & 0.29 (0\%) & 0.52 (0\%) & 0.05 (0\%) \\
gpt-4o-mini     & 0.66 (20\%) & 0.39 (0\%) & \textbf{1.01} (\textbf{50}\%) & 0.29 (0\%) & 0.54(0\%) & 0.09 (0\%) \\
gpt-4o          & 0.76 (20\%) & 0.47 (0\%) & 0.80 (30\%) & \textbf{0.35} (0\%)  & \textbf{0.56} (0\%) & \textbf{0.15} (0\%) \\
claude3-haiku   & 0.58 (5\%) & \textbf{0.48} (0\%) & 0.48 (0\%) & 0.32 (0\%)  & 0.52 (0\%) & 0.10 (0\%) \\
llama3.1-8b     & 0.19 (0\%) & 0.23 (0\%) & 0.18 (0\%) & 0.14 (0\%)  & 0.49 (0\%) & 0.00 (0\%) \\
glm-4-plus      & \textbf{0.81}(\textbf{25}\%) & 0.46 (0\%) & 0.47 (0\%) & 0.33 (0\%)  & 0.54 (\textbf{5}\%) & \textbf{0.15} (0\%) \\
\bottomrule
\end{tabular}
\end{center}
\vspace{-0.3cm}
\end{table}

\begin{table}[h]
\vspace{0.1cm}
\caption{Kill/Death Rates and Winning Rates of LLMs in LLM-PySC2 Tasks (level-1).}\label{tab3}
\begin{center}
\vspace{-0.2cm}
\small
\renewcommand\arraystretch{1.3}
\begin{tabular}{p{2.0cm} p{1.4cm} p{1.4cm} p{1.5cm} p{1.3cm} p{1.4cm} p{1.3cm} p{1.4cm}}
\toprule
Model/KD(WR) & task1 & task2 & task3 & task4 & task5 & task6 & task7 \\
\midrule
gpt-3.5-turbo   & 1.23 (58\%) & 0.13 (4\%) & 6.63 (38\%) & 0.38 (0\%) & 0.61 (8\%) & 0.28 (0\%) & \textbf{1.29} (\textbf{72}\%) \\
gpt-4o-mini     & 1.67 (70\%) & 0.16 (0\%) & 3.46 (0\%) & 0.39 (0\%) & 0.62 (20\%) & 0.30 (0\%) & 1.02 (40\%) \\
gpt-4o          & \textbf{2.27} (80\%) & 0.16 (\textbf{10}\%) & \textbf{Inf} (\textbf{100}\%) & \textbf{0.46} (0\%) & -- & -- & -- \\
claude3-haiku   & 2.19 (\textbf{90}\%) & 0.19 (\textbf{10}\%) & 5.25 (40\%) & 0.34 (0\%) & \textbf{0.75} (\textbf{25}\%) & \textbf{0.33} (0\%) & 0.93 (45\%) \\
llama3.1-8b     & 0.28 (5\%) & 0.12 (5\%) & 14.9 (75\%) & 0.18 (0\%) & 0.48 (5\%) & 0.14 (0\%) & 0.71 (25\%) \\
glm-4-plus      & 0.78 (30\%) & \textbf{0.21} (5\%) & 153 (\textbf{100}\%) & 0.38 (0\%) & 0.60 (10\%) & 0.30 (0\%) & 1.03 (55\%) \\
\midrule
llama3.1-70b    & 0.36 (15\%) & 0.14 (0\%) & 58.9 (95\%) & 0.33 (0\%) & 0.59 (15\%) & 0.31 (0\%) & 0.71 (30\%) \\
llama3.1-405b   & 0.70 (30\%) & 0.10 (0\%) & 3.0k(\textbf{100}\%) & 0.28 (0\%) & 0.56 (10\%) & 0.32 (0\%) & 0.47 (15\%) \\
gpt-o1-mini     & 1.36 (60\%) & 0.04 (0\%) & -- & -- & -- & -- & -- \\
\bottomrule
\end{tabular}
\end{center}
\vspace{-0.2cm}
\end{table}

\begin{table}[h]
\vspace{0.1cm}
\caption{Kill/Death Rates and Winning Rates of Gpt-3.5-turbo in LLM-PySC2 Tasks (level-1/2/3).}\label{tab4}
\begin{center}
\vspace{-0.1cm}
\small
\renewcommand\arraystretch{1.3}
\begin{tabular}{p{1.8cm} p{1.4cm} p{1.3cm} p{1.4cm} p{1.3cm} p{1.3cm} p{1.3cm} p{1.4cm}}
\toprule
Task/KD(WR) & task1 & task2 & task3 & task4 & task5 & task6 & task7 \\
\midrule
task-level-1   & 1.23 (58\%) & 0.13 (4\%) & 6.63 (38\%) & 0.38 (0\%) & 0.61 (8\%) & 0.28 (0\%) & 1.29 (72\%) \\
task-level-2   & 0.56 (5\%) & 0.04 (0\%) & 3.31 (5\%) & 0.34 (0\%) & 0.52 (0\%) & 0.20 (0\%) & 0.98 (25\%) \\
task-level-3   & 0.39 (0\%) & 0.05 (0\%) & 1.99 (0\%) & 0.31 (0\%) & 0.40 (0\%) & 0.26 (0\%) & 0.62 (0\%) \\
\bottomrule
\end{tabular}
\end{center}
\vspace{-0.2cm}
\end{table}

In the micro-operation tasks, we conducted 20 repeated experiments for each LLM (except GPT-3.5-turbo which evaluates 50 games for each task). As shown in Table.~\ref{tab2}, all the tested LLMs act poorly in LLM-SMAC scenarios, similar to works such as \cite{ma2025vlms}. LLMs make obvious mistakes that do not move their long-range combat units, even when attacked by melee units in 3s\_vs\_3z. In 2s3z, the agent for Stalkers sometimes escapes from the battlefield, resulting in the quicker death of ally Zealots.

In the LLM-PySC2 task group, we evaluate the performance of 9 models. Results in Table.~\ref{tab3} demonstrate that the proposed tasks evaluate the decision-making ability of different models more effectively than LLM-SMAC tasks.
Some LLMs achieve the task goal with a high success rate, while others cannot. We find that LLM suffers from hallucinations. More details are provided in the following discussions.
Two additional findings derive from these results: (1) Reasoning models such as GPT-o1-mini cannot significantly improve the decision-making ability in an environment never seen before; (2) Scaling law does not work well in decision-making problems that Llama3.1-405b does not significantly outperform Llama3.1-70b (but enough parameters is crucial for basic decision-making ability). These problems are possibly due to a lack of relevant knowledge and instructions in the pre-training stage.

To avoid the situation that all tasks achieve 100\% winning rates several years after its proposal, we set three difficulty levels for the LLM-PySC2 task group. As the level grows, it will be more difficult for the LLMs to reach the goal due to additional enemy units or upgrades. We evaluate the performance of GPT-3.5-turbo in these tasks, as shown in Table.~\ref{tab4}, and serve it as a baseline for future research.

\section{Discussion}

In this section, we discuss three challenges identified in the experiments. These challenges significantly reduce performance, severely hindering the application of the LLM-based decision-making system.




\textbf{Lack of domain knowledge}. Correct and sufficient knowledge is the prerequisite for correct decisions. However, there is no guarantee that all knowledge across all fields are introduced in the pre-training phase
As a result, LLMs may not realize that 49 additional supplies are far beyond the demand for StarCraft II games, or know that shields recharge in the 2s\_vs\_1sc scenario. 

\textbf{Hallucinations and mistakes}. Hallucinations and mistakes have an inevitable impact on the decision-making process. LLM suffers from (1) input-conflicting hallucinations that generate invalid actions; (2) fact-conflicting hallucinations that mistake ally units for enemy units; and (3) context-conflicting hallucinations that mistake screen coordinates for minimap coordinates. 

\textbf{Inefficient collaboration}. Effective information exchange is critical for multi-agent collaboration. However, LLMs generate communication messages with a lot of non-essential and incorrect information. At the same time, they tend to unconditionally trust their teammates and ignore possible errors in the incoming information, which severely damages the performance in StarCraft II games.

These problems hinder the further application of LLM-based intelligent decision-making systems, waiting for further research and solutions. Examples of these problems are provided in Appendix E.

\section{Conclusion}

In this paper, we introduce a new environment for LLM decision-making, the first environment that accommodates the complete continuous PySC2 actions, and the first LLM StarCraft II environment with a multi-agent framework and communication system. In experiments, we evaluated the performance of mainstream LLMs in complete StarCraft II games and both the LLM-SMAC and LLM-PySC2 task groups, among which the LLM-PySC2 task group is a brand-new experimental scenario that we designed for large models. Results show that LLMs can make decisions and generate valid actions but cannot make effective decisions consistently. Still, the quality of the decision is relatively low and there are several problems such as hallucinations, poor utilization of knowledge of the game, and lack of understanding of the world. Results indicate that learning in the deployment environment is necessary for LLM-based decision-making solutions. We hope the LLM-PySC2 environment can promote research on LLM learning methods, helping LLM-based decision-making methods better adapt to task scenarios.




\begin{thebibliography}{10}
%

%



\bibitem{tan2024towards}  
Tan, W., Ding, Z., Zhang, W., Li, B., Zhou, B., Yue, J., Xia, H., Jiang, J., Zheng, L., Xu, X., Bi, Y., Gu, P., Wang, X., Karlsson, B. F., An, B.,\emph{et al.}, Lu, Z.
\newblock Towards General Computer Control: A Multimodal Agent for Red Dead Redemption II as a Case Study.
\newblock In {\em ICLR 2024 Workshop on Large Language Model (LLM) Agents}, January 2024.


\bibitem{xu2023exploring}  
Xu, Y., Wang, S., Li, P., Luo, F., Wang, X., Liu, W.,\emph{et al.}, Liu, Y.
\newblock Exploring Large Language Models for Communication Games: An Empirical Study on Werewolf.
\newblock {\em arXiv preprint arXiv:2309.04658}, 2023.


\bibitem{ChatDev}  
Qian, C., Liu, W., Liu, H., Chen, N., Dang, Y., Li, J., Yang, C., Chen, W., Su, Y., Cong, X., Xu, J.,\emph{et al.}, Sun, M.
\newblock ChatDev: Communicative Agents for Software Development.
\newblock {\em arXiv preprint arXiv:2307.07924}, 2023.


\bibitem{hua2024gametheoreticllmagentworkflow}  %
Hua, W., Liu, O., Li, L., Amayuelas, A., Chen, J., Jiang, L., Jin, M., Fan, L., Sun, F., Wang, W., \emph{et al.}, Zhang, Y.
\newblock Game-Theoretic LLM: Agent Workflow for Negotiation Games.
\newblock {\em arXiv preprint arXiv:2411.05990}, 2024.

\bibitem{shao2024swarmbrainembodiedagentrealtime}  
Shao, X., Jiang, W., \emph{et al.},Liu, M. 
\newblock SwarmBrain: Embodied Agent for Real-Time Strategy Game StarCraft II via Large Language Models.
\newblock In {\em arXiv preprint arXiv:2401.17749}, 2024. URL: \url{https://arxiv.org/abs/2401.17749}.



\bibitem{2023arXiv231211865M}  
Ma, W., Mi, Q., Zeng, Y., Yan, X., Wu, Y., Lin, R., \emph{et al.}, Wang, J.
\newblock Large Language Models Play StarCraft II: Benchmarks and a Chain of Summarization Approach.
\newblock In {\em Advances in Neural Information Processing Systems}, volume 37, pages 133386--133442, 2024.


\bibitem{li2025hierarchical}  
Z. Li, C. Lu, X. Xu, R. Qi, Y. Ni, L. Jiang, \emph{et al.}, X. Guo.
\newblock Hierarchical Expert Prompt for Large-Language-Model: An Approach Defeat Elite AI in TextStarCraft II for the First Time.
\newblock {\em arXiv preprint arXiv:2502.11122}, 2025.


\bibitem{2025arXiv250213388}   
Xu, X., Li, Z., Lu, C., Qi, R., Ni, Y., Jiang, L., Liu, X., Zhang, X., Fang, Y., Huang, \emph{et al.}, Li, Z.
\newblock Reflection of Episodes: Learning to Play Game from Expert and Self Experiences
\newblock arXiv preprint arXiv:2502.13388, 2025.




\bibitem{MCghost}  
Zhu, X., Chen, Y., Tian, H., Tao, C., Su, W., Yang, C., Huang, G., Li, B., Lu, L., Wang, X., Qiao, Y.,\emph{et al.}, Dai, J .
\newblock Ghost in the minecraft: Generally capable agents for open-world environments via large language models with text-based knowledge and memory
\newblock arXiv preprint arXiv:2305.17144, 2023.

\bibitem{MCVoyager}    
Wang, G., Xie, Y., Jiang, Y., Mandlekar, A., Xiao, C.,\emph{et al.}, Anandkumar, A.
\newblock Voyager: An open-ended embodied agent with large language models. 
\newblock arXiv preprint arXiv:2305.16291, 2023.


\bibitem{MCLlamaRider}   
Feng, Y., Wang, Y., \emph{et al.}, Lu, Z.
\newblock Llama rider: Spurring large language models to explore the open world. 
\newblock arXiv preprint arXiv:2310.08922, 2023.

\bibitem{li2025parallelized}  
Y. Li, S. Liu, T. Zheng, M. Song.
\newblock Parallelized Planning-Acting for Efficient LLM-based Multi-Agent Systems.
\newblock {\em arXiv preprint arXiv:2503.03505}, 2025.




\bibitem{liu2024enhancing}
Liu, H., Zhu, Y., Kato, K., Tsukahara, A., Kondo, I., \emph{et al.}, Hasegawa, Y.
\newblock Enhancing the LLM-Based Robot Manipulation Through Human-Robot Collaboration.
\newblock {\em IEEE Robotics and Automation Letters}, 2024.

\bibitem{shah2023lmnav}
D. Shah, B. Osiński, S. Levine.
\newblock LM-Nav: Robotic Navigation with Large Pre-Trained Models of Language, Vision, and Action.
\newblock In {\em Conference on Robot Learning}, pages 492--504, March 2023.

\bibitem{doma2024llm}
P. Doma, A. Arab, X. Xiao.
\newblock LLM-Enhanced Path Planning: Safe and Efficient Autonomous Navigation with Instructional Inputs.
\newblock {\em arXiv preprint arXiv:2412.02655}, 2024.

\bibitem{jin2024robotgpt}
Jin, Y., Li, D., Shi, J., Hao, P., Sun, F.,\emph{et al.},  Fang, B.
\newblock RobotGPT: Robot Manipulation Learning From ChatGPT.
\newblock {\em IEEE Robotics and Automation Letters}, vol. 9, no. 3, pp. 2543--2550, 2024.

\bibitem{dorbala2023embodied}
Dorbala, V. S., Mullen, \emph{et al.},Manocha, D.
\newblock Can an Embodied Agent Find Your ``Cat-Shaped Mug''? LLM-Based Zero-Shot Object Navigation.
\newblock {\em IEEE Robotics and Automation Letters}, vol. 9, no. 5, pp. 4083--4090, 2023.



\bibitem{xiao2024tradingagents}
Xiao, Y., Sun, E., \emph{et al.}, Wang, W.
\newblock TradingAgents: Multi-Agents LLM Financial Trading Framework.
\newblock {\em arXiv preprint arXiv:2412.20138}, 2024.

\bibitem{ouyang2024modal} 
Ouyang, K., Liu, Y., Li, S., Bao, R., \emph{et al.}, Sun, X.
\newblock Modal-adaptive Knowledge-enhanced Graph-based Financial Prediction from Monetary Policy Conference Calls with LLM.
\newblock In {\em Proceedings of the Joint Workshop of the 7th Financial Technology and Natural Language Processing, the 5th Knowledge Discovery from Unstructured Data in Financial Services, and the 4th Workshop on Economics and Natural Language Processing}, pages 59--69. Association for Computational Linguistics, Torino, Italia, 2024.

\bibitem{li2023large}
Y. Li, S. Wang, H. Ding, H. Chen.
\newblock Large Language Models in Finance: A Survey.
\newblock In {\em Proceedings of the Fourth ACM International Conference on AI in Finance}, pages 374--382, November 2023.



\bibitem{zeng2023flowmind}
Z. Zeng, W. Watson, N. Cho, S. Rahimi, S. Reynolds, \emph{et al.}, M. Veloso.
\newblock FlowMind: Automatic Workflow Generation with LLMs.
\newblock In {\em Proceedings of the Fourth ACM International Conference on AI in Finance}, pages 73--81, November 2023.

\bibitem{xu2024llm4workflow}
J. Xu, W. Du, X. Liu, X. Li.
\newblock LLM4Workflow: An LLM-Based Automated Workflow Model Generation Tool.
\newblock In {\em Proceedings of the 39th IEEE/ACM International Conference on Automated Software Engineering}, pages 2394--2398, October 2024.




\bibitem{fan2022minedojo}
Fan, L., Wang, G., Jiang, Y., Mandlekar, A., Yang, Y., Zhu, H., Tang, A., Huang, D., \emph{et al.}, Anandkumar, A.
\newblock MineDojo: Building Open-Ended Embodied Agents with Internet-Scale Knowledge.
\newblock {\em Advances in Neural Information Processing Systems}, vol. 35, pages 18343--18362, 2022.







\bibitem{park2023generative}
Park, J. S., O'Brien, J., Cai, C. J., Morris, M. R., \emph{et al.}, Bernstein, M.
\newblock Generative Agents: Interactive Simulacra of Human Behavior.
\newblock In {\em Proceedings of the 36th Annual ACM Symposium on User Interface Software and Technology}, pages 1--22, October 2023.










\bibitem{sunehag2017valuedecompositionnetworkscooperativemultiagent}
Sunehag, P., Lever, G., Gruslys, A., Czarnecki, W., Zambaldi, V., Jaderberg, M., Lanctot, M., Sonnerat, N., Leibo,  \emph{et al.}, Graepel, T.
\newblock Value-Decomposition Networks for Cooperative Multi-Agent Learning.
\newblock In {\em arXiv preprint arXiv:1706.05296}, 2017. URL: \url{https://arxiv.org/abs/1706.05296}.

\bibitem{rashid2018qmixmonotonicvaluefunction}
Rashid, T., Samvelyan, M., De Witt, C. S., Farquhar, G., Foerster, \emph{et al.}J.,Whiteson, S.
\newblock Monotonic Value Function Factorisation for Deep Multi-Agent Reinforcement Learning.
\newblock In {\em Journal of Machine Learning Research}, volume 21, number 178, pages 1--51, 2020.

\bibitem{yu2022surprisingeffectivenessppocooperative}
Yu, C., Velu, A., Vinitsky, E., Gao, J., Wang, Y., Bayen, A.,\emph{et al.},Wu, Y .
\newblock The Surprising Effectiveness of PPO in Cooperative Multi-Agent Games.
\newblock In {\em Advances in Neural Information Processing Systems}, volume 35, pages 24611--24624, 2022.


\bibitem{arulkumaran2019alphastar}
Vinyals, O., Babuschkin, I., Czarnecki, W., Mathieu, M., Dudzik, A., Chung, J., Choi, D., Powell, R., Ewalds, T., Georgiev, P., Oh, J., Horgan, D., Kroiss, M., Danihelka, I., Huang, A., Sifre, L., Cai, T., Agapiou, J., Jaderberg, M., Vezhnevets, A., Leblond, R., Pohlen, T., Dalibard, V., Budden, D., Sulsky, Y., Molloy, J., Paine, T., Gulcehre, C., Wang, Z., Pfaff, T., Wu, Y., Ring, R., Yogatama, D., Wünsch, D., McKinney, K., Smith, O., Schaul, T., Lillicrap, T., Kavukcuoglu, \emph{et al.}, C., Silver, D.
\newblock Grandmaster Level in StarCraft II Using Multi-Agent Reinforcement Learning.
\newblock {\em Nature}, vol. 575, no. 7782, pages 350--354, 2019.




\bibitem{contributors2021distar}
DI-star Contributors.
\newblock DI-star: An Open-source Reinforcement Learning Framework for StarCraft II.
\newblock 2021.

%

\bibitem{ma2025vlms}
Ma, W., Fu, Y., Zhang, Z., Li, G.,\emph{et al.} Ghanem, B.
\newblock VLMs Play StarCraft II: A Benchmark and Multimodal Decision Method.
\newblock {\em arXiv e-prints}, arXiv:2503, 2025.










\bibitem{ma2025smachard}
Y Deng, Y Yu, W Ma, Z Wang, W Zhu, J Zhao, Y Zhang
\newblock SMAC-Hard: Enabling Mixed Opponent Strategy Script and Self-play on SMAC
\newblock arXiv preprint arXiv:2412.17707, 2024.




\bibitem{tang2024gensim}
Tang, J., Gao, H., Pan, X., Wang, L., Tan, H., Gao, D., Chen, Y., Chen, X., Lin, Y., Li, Y., Ding, B., Zhou, J., Wang, J., \emph{et al.} ,Wen, J.
\newblock GenSim: A General Social Simulation Platform with Large Language Model Based Agents.
\newblock {\em arXiv preprint arXiv:2410.04360}, 2024.



\bibitem{hua2023war}
Hua, W., Fan, L., Li, L., Mei, K., Ji, J., Ge, Y., Hemphill, L., Zhang, Y.
\newblock War and Peace (WarAgent): Large Language Model-Based Multi-Agent Simulation of World Wars.
\newblock {\em arXiv preprint arXiv:2311.17227}, 2023.

\bibitem{gurcan2024llmaugmentedagentbasedmodellingsocial}
 Gürcan,Ö.
\newblock LLM-Augmented Agent-Based Modelling for Social Simulations: Challenges and Opportunities.
\newblock In {\em HHAI 2024: Hybrid Human AI Systems for the Social Good}, pages 134--144, 2024.
%

\bibitem{vinyals2017starcraft}  
Vinyals, O., Ewalds, T., Bartunov, S., Georgiev, P., Vezhnevets, A., Yeo, M., Makhzani, A., Küttler, H., Agapiou, J., Schrittwieser, J., Quan, J., Gaffney, S., Petersen, S., Simonyan, K., Schaul, T., van Hasselt, H., Silver, D., Lillicrap, T., Calderone, K., Keet, P., Brunasso, A., Lawrence, D., Ekermo, A., Repp, J.,  \emph{et al.}  Tsing, R.
\newblock StarCraft II: A New Challenge for Reinforcement Learning.
\newblock {\em arXiv preprint arXiv:1708.04782}, 2017.


\bibitem{starcraftwiki}   
Blizzard Entertainment.
\newblock StarCraft II.
\newblock {\em StarCraft Wiki}, [Online]. Available: \url{https://starcraft.fandom.com/wiki/StarCraft_II}.


\bibitem{samvelyan2019smac}  
Samvelyan, M., Rashid, T., de Witt, C., Farquhar, G., Nardelli, N., Rudner, T., Hung, C., Torr, P., Foerster, J., \emph{et al.} Whiteson, S.
\newblock The StarCraft Multi-Agent Challenge.
\newblock In {\em arXiv preprint arXiv:1902.04043}, 2019. URL: \url{https://arxiv.org/abs/1902.04043}.










\end{thebibliography}

\appendix
\clearpage
\section*{Appendix A. Pseudo Code}
\setcounter{figure}{0}
\setcounter{table}{0}
\renewcommand{\thefigure}{A\arabic{figure}}
\renewcommand{\thetable}{A\arabic{table}}

\subsection*{A.1 LLM-PySC2 Rollout Process}

\begin{algorithm}
\caption{LLM-PySC2 Rollout Process}
\begin{algorithmic}

\Require Map name. Max waiting time $T$ for a step. Profiles for each agent, with model-name, api-key, api-url for remote LLMs. 

\State 
\State Initialize environment Env 
\State Initialize player with its agents according to the profile
\State Initialize LLM client for agents

\While{not Env.is\_terminated()}

    \State Add the tags of new units to relevant agent's data cache
    \State Remove the tags of dead units from relevant agent's data cache

    \While{current step waiting time < max waiting time}
        \State time.sleep(0.05s)
        
        \If {any agent is waiting for querying remote LLMs}
            \State Collect observations for these agents and wrap the observation into text form
            \State Generate independent threads and query remote LLMs in the threads
        \EndIf
        
        \If {all agents have already got responses}
            \For {$i$ in agents' indexes}
                \State Recognize text actions and generate pysc2 actions into agent\_i's data cache
                \State Recognize communication messages and send to assigned agents
                \State Move camera to the agent's unit team, execute generated pysc2 actions
            \EndFor
        \EndIf
        
    \EndWhile
    \State num\_step += 1

\EndWhile
\State Store the final state and the game result(win/draw/lose)

\end{algorithmic}
\end{algorithm}

\subsection*{A.2 Query Process for an Agent}

\begin{algorithm}
\caption{Query Process for an Agent}
\begin{algorithmic}

\Require Max retry times $n$, max waiting time $T^{'}$ for query.

\State Generate OpenAI message using collected image observation and text information

\State current retry time $i$ = 0
\While{$i$ < $n$}

    \State Reset current waiting time $t$ to 0
    \State Initialize an independent thread and query remote LLM in the thread

    \While{$t$ < $T^{'}$}
        \State time.sleep(0.05s), $t$ += 0.05s
        \If {response received successfully}
            \State Recognize valid actions and generate pysc2 functions for the agent
            \State Break the query process.
        \EndIf
    \EndWhile

    \State waiting for $2^{i}$ seconds to avoid remote service error
    \State $i$ += 1
    
\EndWhile

\State return default action if no valid response received

\end{algorithmic}
\end{algorithm}

\clearpage
\section*{Appendix B. Action Space}
\setcounter{figure}{0}
\setcounter{table}{0}
\renewcommand{\thefigure}{B\arabic{figure}}
\renewcommand{\thetable}{B\arabic{table}}

\begin{table}[h]
\vspace{0.2cm}
\caption{Default Protoss Action Space, Basic Actions}\label{tabb1}
\begin{center}
\vspace{-0.2cm}
\small
\renewcommand\arraystretch{1.2}
\begin{tabular}{p{1.2cm} p{5.0cm} p{7.5cm}}

\toprule
Unit & Text action  & pysc2 functions (id, function, args)\\

\midrule
All unit    & <No\_Operation()> & (0, F.no\_op, ())\\
            & <Hold\_Position()> & (274, F.HoldPosition\_quick, ('queued'))\\
            & <Move\_Screen(screen)> & (331, F.Move\_screen, ('queued', 'screen'))\\
            & <Move\_Minimap(minimap)> & (332, F.Move\_minimap, ('queued', 'minimap'))\\
    & <Select\_Unit\_Move\_Screen(screen)> & (3, F.select\_rect, ('select', 'screen1\_tag', 'screen2\_tag')) \\
    &   & (331, F.Move\_screen, ('now', 'screen')) \\
    & <Select\_Unit\_Move\_Minimap(minimap)> & (3, F.select\_rect, ('select', 'screen1\_tag', 'screen2\_tag')) \\
    &   & (332, F.Move\_minimap, ('queued', 'minimap')) \\
    
\midrule
Attackable & <Attack\_Unit(tag)> & (12, F.Attack\_screen, ('queued', 'screen\_tag'))\\
& <Select\_Unit\_Attack\_Unit(tag, tag)> & (3, F.select\_rect, ('select', 'screen1\_tag', 'screen2\_tag')) \\
& & (12, F.Attack\_screen, ('queued', 'screen\_tag')) \\

\bottomrule
\end{tabular}
\end{center}
\end{table}

Most units share the actions above, so we listed here to avoid mention repetitive unit control actions in subsequent appendices. Here, $F$ refers to $pysc2.lib.actions.FUNCTION$, the same as in the following.

\begin{table}[h]
\vspace{0.2cm}
\caption{Default Protoss Action Space, Standard Building Actions}\label{tabb2}
\begin{center}
\vspace{-0.2cm}
\small
\renewcommand\arraystretch{1.2}
\begin{tabular}{p{1.2cm} p{5.0cm} p{7.5cm}}
\toprule
Unit & Text action  & pysc2 functions (id, function, args)\\
\midrule
Probe  & <Build\_Nexus\_Near(tag)> & (573, F.llm\_pysc2\_move\_camera, ('world\_tag')) \\
& & (65, F.Build\_Nexus\_screen, ('queued', 'screen\_tag')) \\
& <Build\_Assimilator\_Near(tag)> & (573, F.llm\_pysc2\_move\_camera, ('world\_tag'))\\
& & (40, F.Build\_Assimilator\_screen, ('queued', 'screen\_tag')) \\
& <Build\_Nexus\_Screen(screen)> & (65, F.Build\_Nexus\_screen, ('queued', 'screen\_tag'))\\
& <Build\_Assimilator\_Screen(screen)> & (40, F.Build\_Assimilator\_screen, ('queued', 'screen\_tag'))\\
& <Build\_Pylon\_Screen(screen)> & (70, F.Build\_Pylon\_screen, ('queued', 'screen'))\\
& <Build\_Gateway\_Screen(screen)> & (57, F.Build\_Gateway\_screen, ('queued', 'screen'))\\
& <Build\_CyberneticsCore\_Screen(screen)> & (48, F.Build\_CyberneticsCore\_screen, ('queued', 'screen'))\\
& <Build\_Forge\_Screen(screen)> & (55, F.Build\_Forge\_screen, ('queued', 'screen'))\\
& <Build\_PhotonCannon\_Screen(screen)> & (69, F.Build\_PhotonCannon\_screen, ('queued', 'screen'))\\
& <Build\_ShieldBattery\_Screen(screen)> & (525, F.Build\_ShieldBattery\_screen, ('queued', 'screen'))\\
& <Build\_TwilightCouncil\_Screen(screen)>& (101, F.Build\_TwilightCouncil\_screen, ('queued', 'screen'))\\
& <Build\_TemplarArchive\_Screen(screen)>& (100, F.Build\_TemplarArchive\_screen, ('queued', 'screen'))\\
& <Build\_DarkShrine\_Screen(screen)>& (49, F.Build\_DarkShrine\_screen, ('queued', 'screen'))\\
& <Build\_Stargate\_Screen(screen)>& (88, F.Build\_Stargate\_screen, ('queued', 'screen'))\\
& <Build\_FleetBeacon\_Screen(screen)>& (54, F.Build\_FleetBeacon\_screen, ('queued', 'screen'))\\
& <Build\_RoboticsBay\_Screen(screen)>& (81, F.Build\_RoboticsBay\_screen, ('queued', 'screen'))\\
& <Build\_RoboticsFacility\_Screen(screen)>& (82, F.Build\_RoboticsFacility\_screen, ('queued', 'screen'))\\
& <Lock\_Nexus\_Near(tag)> & (70, F.Build\_Pylon\_screen, ('queued', 'screen\_tag'))\\
& <Lock\_Assimilator\_Near(tag)> & (40, F.Build\_Assimilator\_screen, ('queued', 'screen\_tag')) \\
\bottomrule
\end{tabular}
\end{center}
\end{table}

In standard building mode, a worker will be chosen as building-worker at the beginning of the game or the time the worker dead. The 'Builder' agent will control the worker to build buildings using the actions mentioned above.

\begin{table}[h]
\vspace{0.2cm}
\caption{Default Protoss Action Space, Building Actions (easy mode1, for Builder)}\label{tabb3}
\begin{center}
\vspace{-0.2cm}
\small
\renewcommand\arraystretch{1.2}
\begin{tabular}{p{0.8cm} p{4.8cm} p{8.1cm}}
\toprule
Unit & Text action  & pysc2 functions (id, function, args)\\
\midrule
Probe  & <Build\_Nexus\_Near(tag)> & (573, F.llm\_pysc2\_move\_camera, ('world\_tag')) \\
& & (65, F.Build\_Nexus\_screen, ('queued', 'screen\_tag')) \\
& <Build\_Assimilator\_Near(tag)> & (573, F.llm\_pysc2\_move\_camera, ('world\_tag'))\\
& & (40, F.Build\_Assimilator\_screen, ('queued', 'screen\_tag')) \\
& <Build\_Pylon\_Near(tag)> & (70, F.Build\_Pylon\_screen, ('queued', 'screen\_tag'))\\
& <Build\_Gateway\_Near(tag)> & (57, F.Build\_Gateway\_screen, ('queued', 'screen\_tag'))\\
& <Build\_CyberneticsCore\_Near(tag)> & (48, F.Build\_CyberneticsCore\_screen, ('queued', 'screen\_tag'))\\
& <Build\_Forge\_Near(tag)> & (55, F.Build\_Forge\_screen, ('queued', 'screen\_tag'))\\
& <Build\_PhotonCannon\_Near(tag)> & (69, F.Build\_PhotonCannon\_screen, ('queued', 'screen\_tag'))\\
& <Build\_ShieldBattery\_Near(tag)> & (525, F.Build\_ShieldBattery\_screen, ('queued', 'screen\_tag'))\\
& <Build\_TwilightCouncil\_Near(tag)>& (101, F.Build\_TwilightCouncil\_screen, ('queued', 'screen\_tag'))\\
& <Build\_TemplarArchive\_Near(tag)>& (100, F.Build\_TemplarArchive\_screen, ('queued', 'screen\_tag'))\\
& <Build\_DarkShrine\_Near(tag)>& (49, F.Build\_DarkShrine\_screen, ('queued', 'screen\_tag'))\\
& <Build\_Stargate\_Near(tag)>& (88, F.Build\_Stargate\_screen, ('queued', 'screen\_tag'))\\
& <Build\_FleetBeacon\_Near(tag)>& (54, F.Build\_FleetBeacon\_screen, ('queued', 'screen\_tag'))\\
& <Build\_RoboticsBay\_Near(tag)>& (81, F.Build\_RoboticsBay\_screen, ('queued', 'screen\_tag'))\\
& <Build\_RoboticsFacility\_Near(tag)>& (82, F.Build\_RoboticsFacility\_screen, ('queued', 'screen\_tag'))\\
& <Lock\_Nexus\_Near(tag)> & (70, F.Build\_Pylon\_screen, ('queued', 'screen\_tag'))\\
& <Lock\_Assimilator\_Near(tag)> & (40, F.Build\_Assimilator\_screen, ('queued', 'screen\_tag')) \\
\bottomrule
\end{tabular}
\end{center}
\end{table}

In easy-build mode-1, the agent 'Builder' does not need to provide precision position, but a tag of nearby buildings. The LLM-PySC2 program will autonomously find a position near the unit with given tag and build new buildings there.

\begin{table}[h]
\vspace{0.2cm}
\caption{Default Protoss Action Space, Building Actions (easy mode2, for Developer)}\label{tabb4}
\begin{center}
\vspace{-0.2cm}
\small
\renewcommand\arraystretch{1.2}
\begin{tabular}{p{0.8cm} p{4.8cm} p{8.1cm}}
\toprule
Unit & Text action  & pysc2 functions (id, function, args)\\
\midrule
Probe  & <Build\_Nexus()> & (65, F.Build\_Nexus\_screen, ('queued', 'auto')) \\
& <Build\_Assimilator()> & (40, F.Build\_Assimilator\_screen, ('queued', 'auto')) \\
& <Build\_Pylon()> & (70, F.Build\_Pylon\_screen, ('queued', 'auto'))\\
& <Build\_Gateway()> & (57, F.Build\_Gateway\_screen, ('queued', 'auto'))\\
& <Build\_CyberneticsCore()> & (48, F.Build\_CyberneticsCore\_screen, ('queued', 'auto'))\\
& <Build\_Forge()> & (55, F.Build\_Forge\_screen, ('queued', 'auto'))\\
& <Build\_PhotonCannon()> & (69, F.Build\_PhotonCannon\_screen, ('queued', 'auto'))\\
& <Build\_ShieldBattery()> & (525, F.Build\_ShieldBattery\_screen, ('queued', 'auto'))\\
& <Build\_TwilightCouncil()>& (101, F.Build\_TwilightCouncil\_screen, ('queued', 'auto'))\\
& <Build\_TemplarArchive()>& (100, F.Build\_TemplarArchive\_screen, ('queued', 'auto'))\\
& <Build\_DarkShrine()>& (49, F.Build\_DarkShrine\_screen, ('queued', 'auto'))\\
& <Build\_Stargate()>& (88, F.Build\_Stargate\_screen, ('queued', 'auto'))\\
& <Build\_FleetBeacon()>& (54, F.Build\_FleetBeacon\_screen, ('queued', 'auto'))\\
& <Build\_RoboticsBay()>& (81, F.Build\_RoboticsBay\_screen, ('queued', 'auto'))\\
& <Build\_RoboticsFacility()>& (82, F.Build\_RoboticsFacility\_screen, ('queued', 'auto'))\\
\bottomrule
\end{tabular}
\end{center}
\end{table}

In easy-build mode-2, the agent 'Builder' does not need to provide any additional information of where to build the building. the program will automatically find a position for above actions. Experiments of ECEB mode and SCEB mode use these actions as the Developer's building actions.

\begin{table}[h]
\vspace{0.2cm}
\caption{Default Protoss Action Space, Researching Actions}\label{tabb5}
\begin{center}
\vspace{-0.2cm}
\small
\renewcommand\arraystretch{1.2}
\begin{tabular}{ p{5.5cm} p{8.2cm}}
\toprule
Text action  & pysc2 functions (id, function, args)\\
\midrule
<Research\_ProtossAirArmor()> & (381, F.Research\_ProtossAirArmor\_quick, ('queued'))\\
<Research\_ProtossAirWeapons()> & (385, F.Research\_ProtossAirWeapons\_quick, ('queued'))\\
<Research\_WarpGate()> & (428, F.Research\_WarpGate\_quick, ('queued'))\\
<Research\_ProtossGroundArmor()> & (389, F.Research\_ProtossGroundArmor\_quick, ('queued'))\\
<Research\_ProtossGroundWeapons()> & (393, F.Research\_ProtossGroundWeapons\_quick, ('queued'))\\
<Research\_ProtossShields()>& (397, F.Research\_ProtossShields\_quick, ('queued'))\\
<Research\_Charge()>& (359, F.Research\_Charge\_quick, ('queued'))\\
<Research\_Blink()>& (356, F.Research\_Blink\_quick, ('queued'))\\
<Research\_AdeptResonatingGlaives()>& (351, F.Research\_AdeptResonatingGlaives\_quick, ('queued'))\\
<Research\_PhoenixAnionPulseCrystals()>& (379, F.Research\_PhoenixAnionPulseCrystals\_quick, ('queued'))\\
<Research\_ExtendedThermalLance()>& (364, F.Research\_ExtendedThermalLance\_quick, ('queued'))\\
<Research\_GraviticBooster()>& (366, F.Research\_GraviticBooster\_quick, ('queued'))\\
<Research\_GraviticDrive()>& (367, F.Research\_GraviticDrive\_quick, ('queued'))\\
<Research\_PsiStorm()>& (401, F.Research\_PsiStorm\_quick, ('queued'))\\
<Research\_ShadowStrike()>& (404, F.Research\_ShadowStrike\_quick, ('queued'))\\
\bottomrule
\end{tabular}
\end{center}
\end{table}

Researching actions are actually complex actions of combination pysc2 functions, they require first selecting a research building and then starting the research.
The program will autonomously find idle building for these functions, select the building and execute the pysc2 functions for technology upgrades.

\begin{table}[h]
\vspace{0.2cm}
\caption{Default Protoss Action Space, Unit Training Actions}\label{tabb6}
\begin{center}
\vspace{-0.2cm}
\small
\renewcommand\arraystretch{1.2}
\begin{tabular}{p{1.4cm} p{4.0cm} p{8.3cm}}
\toprule
Unit & Text action  & pysc2 functions (id, function, args)\\
\midrule
Nexus & <Train\_Mothership()> & (541, F.Train\_Mothership\_quick, ('queued'))\\
\midrule
Gateway & <Train\_Adept()>& (457, F.Train\_Adept\_quick, ('queued'))\\
& <Train\_DarkTemplar()>& (465, F.Train\_DarkTemplar\_quick, ('queued')) \\
& <Train\_HighTemplar()>& (471, F.Train\_HighTemplar\_quick, ('queued'))\\
& <Train\_Sentry()>& (491, F.Train\_Sentry\_quick, ('queued'))\\
& <Train\_Stalker()>& (493, F.Train\_Stalker\_quick, ('queued'))\\
& <Train\_Zealot()>& (503, F.Train\_Zealot\_quick, ('queued'))\\
\midrule
Stargate & <Train\_Oracle()>& (482, F.Train\_Oracle\_quick, ('queued'))\\
& <Train\_Phoenix()>& (484, F.Train\_Phoenix\_quick, ('queued'))\\
& <Train\_VoidRay()>& (500, F.Train\_VoidRay\_quick, ('queued'))\\
& <Train\_Tempest()>& (495, F.Train\_Tempest\_quick, ('queued'))\\
& <Train\_Carrier()>& (461, F.Train\_Carrier\_quick, ('queued'))\\
\midrule
RoboticBay & <Train\_Observer()>& (481, F.Train\_Observer\_quick, ('queued'))\\
& <Train\_WarpPrism()>& (501, F.Train\_WarpPrism\_quick, ('queued'))\\
& <Train\_Immortal()>& (473, F.Train\_Immortal\_quick, ('queued'))\\
& <Train\_Colossus()>& (462, F.Train\_Colossus\_quick, ('queued'))\\
& <Train\_Disruptor()>& (466, F.Train\_Disruptor\_quick, ('queued'))\\
\bottomrule
\end{tabular}
\end{center}
\end{table}

Unit training actions share the same pre-process of researching actions. To avoid stocking a lot of resources in the training queue, the program only trains units in idle buildings. To avoid spending too many tokens on finding suitable buildings, the program autonomously searches idle buildings for these actions.

\begin{table}[h]
\vspace{0.2cm}
\caption{Default Protoss Action Space, Unit Warp Actions and Warp Actions in easy mode}\label{tabb7}
\begin{center}
\vspace{-0.2cm}
\small
\renewcommand\arraystretch{1.2}
\begin{tabular}{p{1.2cm} p{4.2cm} p{8.3cm}}
\toprule
Unit & Text action  & pysc2 functions (id, function, args)\\
\midrule
WarpGate    & <Warp\_Adept\_Near(tag)> & (8, F.select\_warp\_gates, ('select'))\\
& & (573, F.llm\_pysc2\_move\_camera, ('world\_tag'))\\
& & (505, F.TrainWarp\_Adept\_screen, ('queued', 'screen\_tag'))\\

& <Warp\_DarkTemplar\_Near(tag)> & (8, F.select\_warp\_gates, ('select'))\\
& & (573, F.llm\_pysc2\_move\_camera, ('world\_tag'))\\
& & (506, F.TrainWarp\_DarkTemplar\_screen, ('queued', 'screen\_tag'))\\

& <Warp\_HighTemplar\_Near(tag)> & (8, F.select\_warp\_gates, ('select'))\\
& & (573, F.llm\_pysc2\_move\_camera, ('world\_tag'))\\
& & (507, F.TrainWarp\_HighTemplar\_screen, ('queued', 'screen\_tag')) \\

& <Warp\_Sentry\_Near(tag)> & (8, F.select\_warp\_gates, ('select'))\\
& & (573, F.llm\_pysc2\_move\_camera, ('world\_tag'))\\
& & (505, F.TrainWarp\_Sentry\_screen, ('queued', 'screen\_tag'))\\

& <Warp\_Stalker\_Near(tag)> & (8, F.select\_warp\_gates, ('select'))\\
& & (573, F.llm\_pysc2\_move\_camera, ('world\_tag'))\\
& & (506, F.TrainWarp\_Stalker\_screen, ('queued', 'screen\_tag'))\\

& <Warp\_Zealot\_Near(tag)> & (8, F.select\_warp\_gates, ('select'))\\
& & (573, F.llm\_pysc2\_move\_camera, ('world\_tag'))\\
& & (507, F.TrainWarp\_Zealot\_screen, ('queued', 'screen\_tag')) \\

\midrule

WarpGate    & <Warp\_Zealot()> & (510, F.TrainWarp\_Zealot\_screen, ('queued', 'auto'))\\
            & <Warp\_Stalker()> & (509, F.TrainWarp\_Stalker\_screen, ('queued', 'auto'))\\
            & <Warp\_Sentry()> & (508, F.TrainWarp\_Sentry\_screen, ('queued', 'auto'))\\
            & <Warp\_Adept()> & (505, F.TrainWarp\_Adept\_screen, ('queued', 'auto'))\\
    & <Warp\_HighTemplar(screen)> & (507, F.TrainWarp\_HighTemplar\_screen, ('queued', 'auto')) \\
    & <Warp\_DarkTemplar(minimap)> & (506, F.TrainWarp\_DarkTemplar\_screen, ('queued', 'auto')) \\

\bottomrule
\end{tabular}
\end{center}
\end{table}

For protoss, some of the unit training actions will change to unit warpping actions after WarpGate technology upgrades. They need to choose a tag for power field provider (such as Pylon) to warp unit there. In easy-warp mode, the program will autonomously find valid position for unit warping actions.

\begin{table}[h]
\vspace{0.2cm}
\caption{Default Protoss Action Space, Easy Control Actions}\label{tabb8}
\begin{center}
\vspace{-0.2cm}
\small
\renewcommand\arraystretch{1.2}
\begin{tabular}{p{5.0cm} p{9.0cm}}

\toprule
Text action  & pysc2 functions (id, function, args)\\

\midrule
<All\_Units\_Attack()> & (13, F.Attack\_minimap, ('auto'))\\
<All\_Units\_Defend()> & (331, F.Move\_screen, ('queued', 'auto'))\\
<All\_Units\_Retreat()> & (331, F.Move\_screen, ('now', 'auto'))\\

\midrule
<Worker\_Scan()> & (332, F.Move\_minimap, ('queued', 'auto'))\\
<Zealot\_Scan()> & (332, F.Move\_minimap, ('queued', 'auto')) \\
<Adept\_Scan()> & (332, F.Move\_minimap, ('queued', 'auto')) \\
<Pheonix\_Scan()> & (332, F.Move\_minimap, ('queued', 'auto')) \\
<Oracle\_Scan()> & (332, F.Move\_minimap, ('queued', 'auto')) \\
<Observer\_Scan()> & (332, F.Move\_minimap, ('queued', 'auto')) \\

\bottomrule
\end{tabular}
\end{center}
\end{table}

We provide easy control actions, a series of actions similar to TextStarCraft-II unit control actions. For researchers who focus on studying LLM-based planning (develop the economy) or VLM-based decision making(precisely build buildings), simplifying unit control actions can provide great convenience.

\begin{table}[h]
\tiny
\vspace{0.2cm}
\caption{Default Protoss Action Space, Unit Skills (Part1, control a unit team)}\label{tabb9}
\begin{center}
\vspace{-0.2cm}
\small
\renewcommand\arraystretch{1.2}
\begin{tabular}{p{1.2cm} p{5.4cm} p{7.5cm}}

\toprule
Unit & Text action  & pysc2 functions (id, function, args)\\

\midrule
Adept & <Ability\_AdeptPhaseShift\_Minimap(& (547, F.Effect\_AdeptPhaseShift\_minimap, \\
&minimap)> &('now', 'minimap')) \\
& <Ability\_AdeptPhaseShift\_Screen(screen)> & (177, F.Effect\_AdeptPhaseShift\_screen, ('now', 'screen'))\\
& <Ability\_CancelPhaseShift>& (141, F.Cancel\_AdeptPhaseShift\_quick, ('now'))\\

\midrule
Stalker & <Ability\_Blink\_Screen(screen)> & (180, F.Effect\_Blink\_screen, ('now', 'screen'))\\

\midrule
Sentry & <Ability\_ForceField\_Screen(screen)> & (193, F.Effect\_ForceField\_screen, ('queued', 'screen'))\\
& <Ability\_GuardianShield()>& (197, F.Effect\_GuardianShield\_quick, ('queued')) \\

\midrule
HighTeplar & <Ability\_PsiStorm\_Screen(screen)>& (218, F.Effect\_PsiStorm\_screen, ('queued', 'screen'))\\
& <Ability\_PsiStorm\_Attack\_Unit(tag)> & (218, F.Effect\_PsiStorm\_screen, ('queued', 'screen\_tag'))\\
& <Morph\_Archon()> & (296, F.Morph\_Archon\_quick, ('queued'))\\
& <Select\_Two\_Units\_Morph\_Archon( & (3, F.select\_rect, ('select', 'screen1\_tag', 'screen2\_tag'))\\
& tag, tag)> & (3, F.select\_rect, ('select', 'screen1\_tag2', 'screen2\_tag2'))\\
&  & (296, F.Morph\_Archon\_quick, ('queued'))\\

\midrule
DarkTeplar& <Ability\_ShadowStride\_Unit(tag)> & (182, F.Effect\_ShadowStride\_screen, \\
& & ('queued', 'screen\_tag')) \\
& <Morph\_Archon()> & (296, F.Morph\_Archon\_quick, ('queued'))\\

\midrule
Observer & <Morph\_SurveillanceMode()>& (538, F.Morph\_SurveillanceMode\_quick, ('queued'))\\
& <Morph\_ObserverMode()>& (535, F.Morph\_ObserverMode\_quick, ('queued'))\\

\midrule
Disruptor & <Ability\_PurificationNova\_Attack(tag)> & (219, F.Effect\_PurificationNova\_screen, \\
& & ('queued', 'screen\_tag')) \\

\midrule
Oracle & <Ability\_PulsarBeamOn()>& (38, F.Behavior\_PulsarBeamOn\_quick, ('queued'))\\
& <Ability\_OracleRevelation\_Screen(screen)>& (214, F.Effect\_OracleRevelation\_screen,\\
& & ('queued', 'screen')) \\
& <Build\_StasisTrap\_Screen(screen)>& (90, F.Build\_StasisTrap\_screen, ('queued', 'screen'))\\

\midrule
Pheoenix& <Ability\_GravitonBeam>& (196, F.Effect\_GravitonBeam\_screen\\
&<Cancel\_GravitonBeam\_For\_All()> & (140, F.Cancel\_quick, ('now')) \\

\midrule
WarpPrism& <Morph\_WarpPrismPhasingMode()>& (329, F.Morph\_WarpPrismPhasingMode\_quick, ('queued'))\\
& <Load\_Unit(tag)>& (287, F.Load\_screen, ('queued', 'screen\_tag'))\\
& <Unload\_Screen(screen)>& (516, F.UnloadAllAt\_screen, ('queued', 'screen'))\\
& <Morph\_WarpPrismTransportMode>& (330, F.Morph\_WarpPrismTransportMode\_quick\\
& &, ('queued')) \\

\midrule
MotherShip& <Ability\_TimeWarp\_Attack(tag)>& (241, F.Effect\_TimeWarp\_screen, ('queued', 'screen\_tag'))\\
& <Ability\_TimeWarp\_Screen(screen)>& (241, F.Effect\_TimeWarp\_screen, ('queued', 'screen'))\\

\bottomrule
\end{tabular}
\end{center}
\end{table}

In LLM-PySC2 environment, agent controls its unit in a group, using above actions. The program will select the units belong to the agent, and then execute LLM-generated actions.

\begin{table}[h]
\tiny
\vspace{0.2cm}
\caption{Default Protoss Action Space, Unit Skills  (Part2, control specific unit)}\label{tabb10}
\begin{center}
\vspace{-0.2cm}
\small
\renewcommand\arraystretch{1.2}
\begin{tabular}{p{1.2cm} p{5.4cm} p{7.5cm}}

\toprule
Unit & Text action  & pysc2 functions (id, function, args)\\

\midrule
Adept & <Select\_Unit\_Ability\_AdeptPhaseShift & (3, F.select\_rect, ('select', 'screen1\_tag', 'screen2\_tag'))\\
&\_Minimap(minimap)> & (547, F.Effect\_AdeptPhaseShift\_minimap,\\
& &('now', 'minimap')) \\

& <Select\_Unit\_Ability\_AdeptPhaseShift & (3, F.select\_rect, ('select', 'screen1\_tag', 'screen2\_tag'))\\
& \_Screen(screen)> & (177, F.Effect\_AdeptPhaseShift\_screen, ('now', 'screen'))\\

& <Select\_Unit\_Ability\_CancelPhaseShift(& (3, F.select\_rect, ('select', 'screen1\_tag', 'screen2\_tag'))\\
&tag)> & (141, F.Cancel\_AdeptPhaseShift\_quick, ('now'))\\

\midrule
Stalker & <Select\_Unit\_Blink\_Screen(tag, screen)> & (3, F.select\_rect, ('select', 'screen1\_tag', 'screen2\_tag'))\\
& & (180, F.Effect\_Blink\_screen, ('now', 'screen'))\\

\midrule
Sentry & <Select\_Unit\_Ability\_ForceField\_Screen(  & (3, F.select\_rect, ('select', 'screen1\_tag', 'screen2\_tag'))\\
 & tag, screen)> & (193, F.Effect\_ForceField\_screen, ('queued', 'screen'))\\

 & <Select\_Unit\_Ability\_GuardianShield( & (3, F.select\_rect, ('select', 'screen1\_tag', 'screen2\_tag'))\\
 & tag)> & (197, F.Effect\_GuardianShield\_quick, ('queued'))\\

\midrule
HighTeplar & <Select\_Two\_Units\_Morph\_Archon( & (3, F.select\_rect, ('select', 'screen1\_tag', 'screen2\_tag'))\\
& tag, tag)>& (3, F.select\_rect, ('add', 'screen1\_tag2', 'screen2\_tag2'))\\
& & (296, F.Morph\_Archon\_quick, ('queued'))\\

& <Select\_Unit\_Ability\_PsiStorm\_Screen & (3, F.select\_rect, ('select', 'screen1\_tag', 'screen2\_tag'))\\
& (tag, screen)> & (218, F.Effect\_PsiStorm\_screen, ('queued', 'screen'))\\

& <Select\_Unit\_Ability\_PsiStorm\_Attack\_Unit( & (3, F.select\_rect, ('select', 'screen1\_tag', 'screen2\_tag'))\\
& (tag, tag)> & (218, F.Effect\_PsiStorm\_screen, ('queued', 'screen\_tag'))\\

\midrule
Disruptor & <Select\_Unit\_Ability\_PurificationNova 
& (3, F.select\_rect, ('add', 'screen1\_tag2', 'screen2\_tag2'))\\
&\_Attack(tag)> & (219, F.Effect\_PurificationNova\_screen, \\
& & ('queued', 'screen\_tag'))\\

\midrule
DarkTeplar& <Select\_Two\_Units\_Morph\_Archon( & (3, F.select\_rect, ('select', 'screen1\_tag', 'screen2\_tag'))\\
& tag, tag)> & (3, F.select\_rect, ('add', 'screen1\_tag2', 'screen2\_tag2'))\\
& & (296, F.Morph\_Archon\_quick, ('queued'))\\

\midrule
Oracle & <Select\_Unit\_Ability\_PulsarBeamOn(tag)>& (3, F.select\_rect, ('select', 'screen1\_tag', 'screen2\_tag'))\\
& & (38, F.Behavior\_PulsarBeamOn\_quick, ('queued'))\\
& <Select\_Unit\_OracleRevelation\_Screen(& (3, F.select\_rect, ('select', 'screen1\_tag', 'screen2\_tag'))\\
&screen)> &(214, F.Effect\_OracleRevelation\_screen, \\
& &  ('queued', 'screen'))\\

& <Select\_Unit\_Build\_StasisTrap\_Screen(& (3, F.select\_rect, ('select', 'screen1\_tag', 'screen2\_tag'))\\
&tag, screen)> & (90, F.Build\_StasisTrap\_screen, ('queued', 'screen')) \\

\midrule
Pheoenix& <Select\_Phoenix\_Ability\_GravitonBeam>& (3, F.select\_rect, ('select', 'screen1\_tag', 'screen2\_tag'))\\
&\_Unit(tag) & (196, F.Effect\_GravitonBeam\_screen, \\
& &('queued', 'screen\_tag2')) \\

&<Cancel\_GravitonBeam\_For\_Phoenix(tag)> & (3, F.select\_rect, ('select', 'screen1\_tag', 'screen2\_tag'))\\
& & (140, F.Cancel\_quick, ('now')) \\

\bottomrule
\end{tabular}
\end{center}
\end{table}

In some senerios, precisely controlling single unit is key to the victory, especially in SMAC tasks and early stage of the game (against high level opponent). We provide single unit control actions for this senarios, and LLM can use the actions whenever they need to improve performance of micro-operations.

\clearpage
\section*{Appendix C. Query message, Prompt, Examples of Observations and Responses}
\setcounter{figure}{0}
\setcounter{table}{0}
\renewcommand{\thefigure}{C\arabic{figure}}
\renewcommand{\thetable}{C\arabic{table}}

\subsection*{C1. Query message and Prompt}

\begin{figure}[h]
  \centering
  \includegraphics[width=0.92\textwidth]{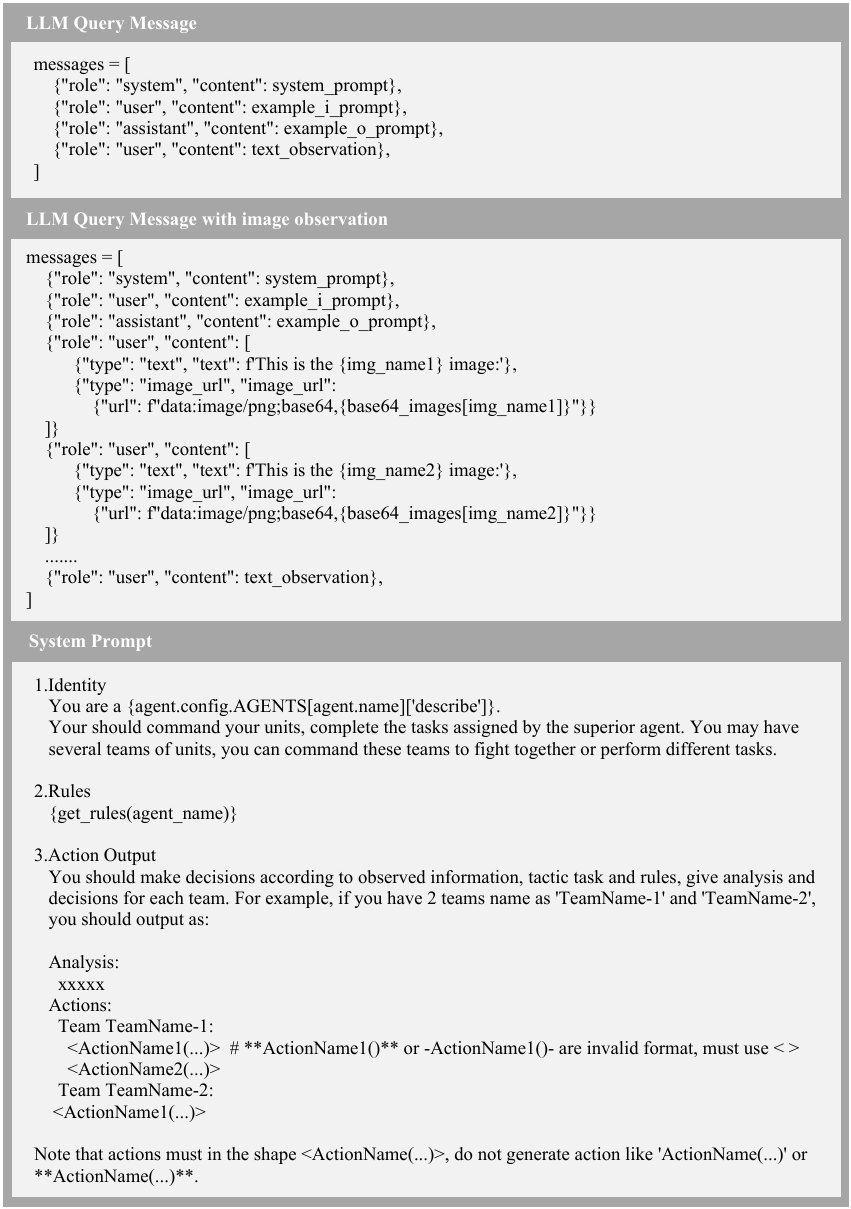}
    \caption{\textbf{OpenAI LLM query message and system prompt in LLM-PySC2.}}
  \label{fig_prompt_m_s}
\end{figure}

\begin{figure}[h]
  \centering
  \includegraphics[width=0.95\textwidth]{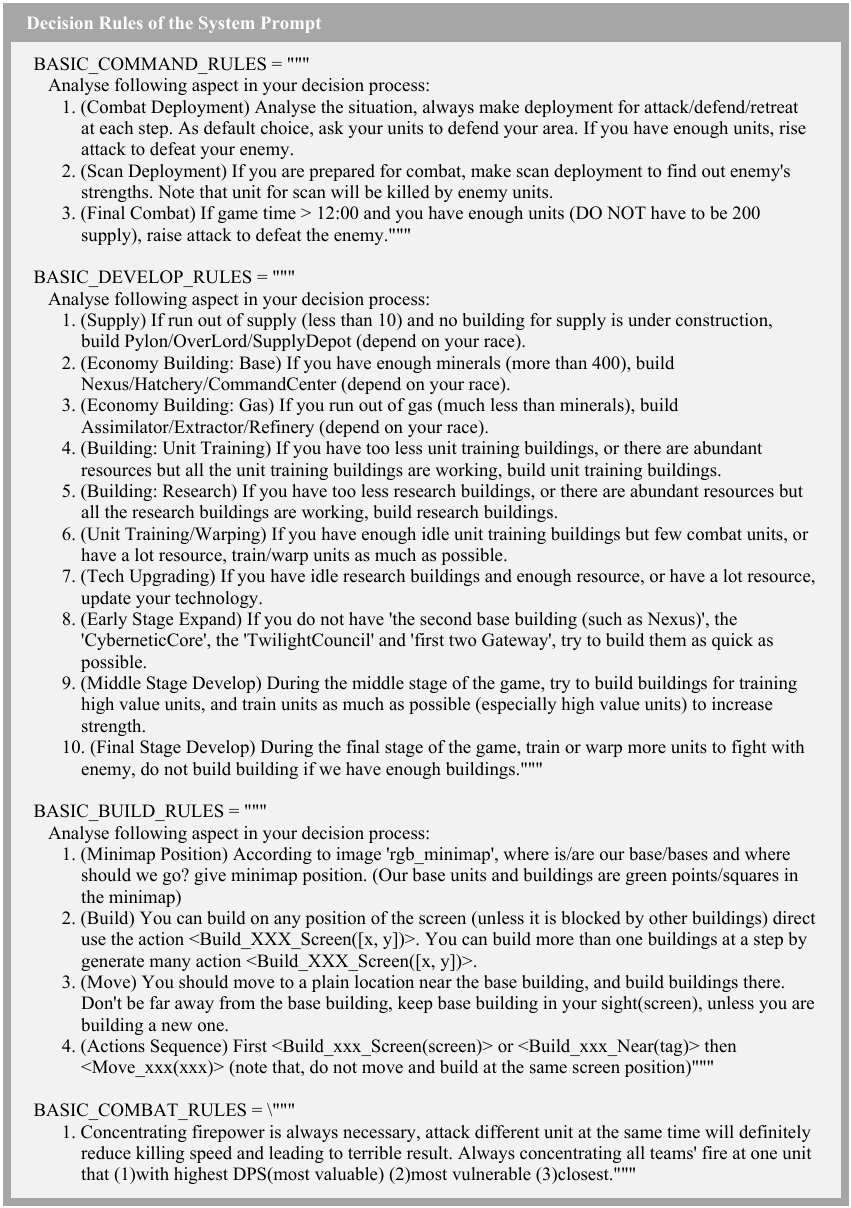}
    \caption{\textbf{Basic Rules for agent Commander, Developer, Builder, CombaGroups.}}
  \label{fig-sys-rules}
\end{figure}

\clearpage
\subsection*{C2. Examples of Textual Observations}

\begin{figure}[h]
  \centering
  \includegraphics[width=0.92\textwidth]{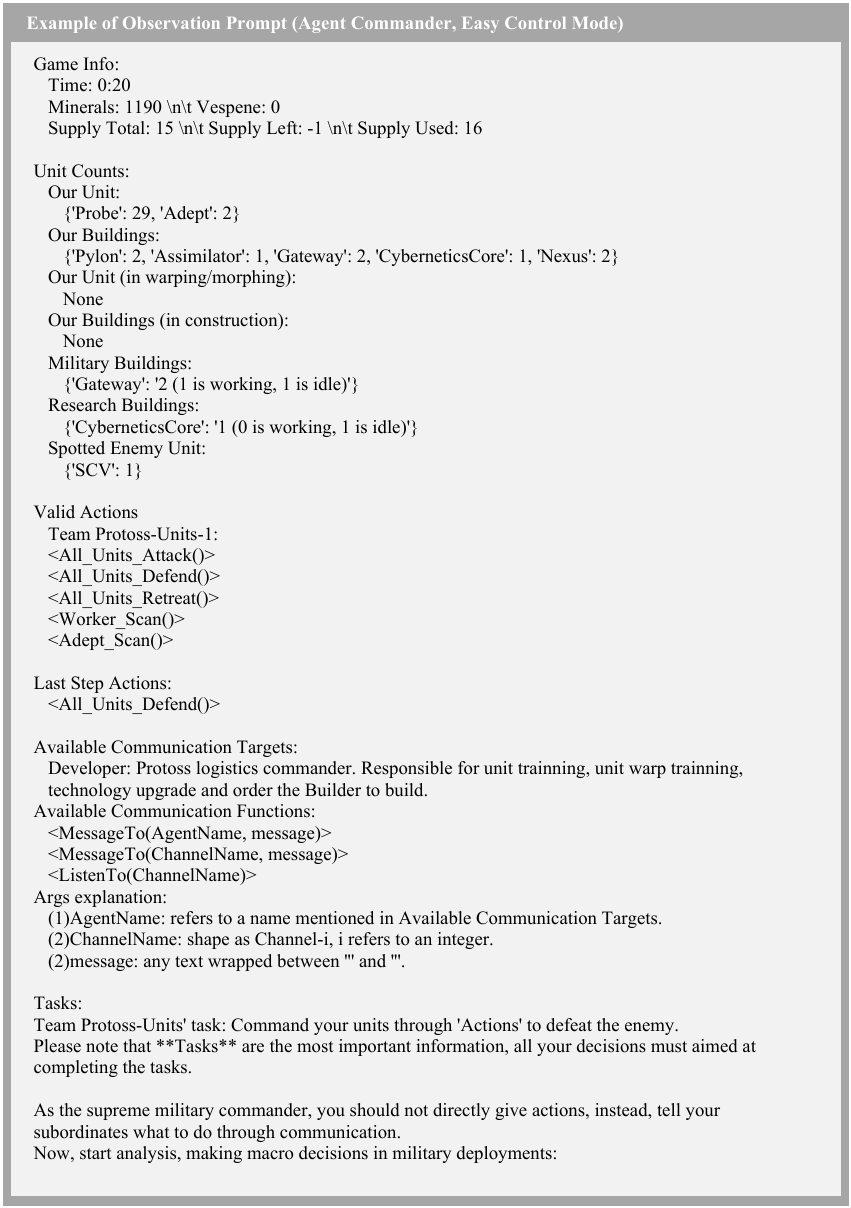}
    \caption{\textbf{Example textual observation of agent 'Commander' in easy control mode.}}
  \label{fig_prompt_c1}
\end{figure}

\begin{figure}[h]
  \centering
  \includegraphics[width=0.95\textwidth]{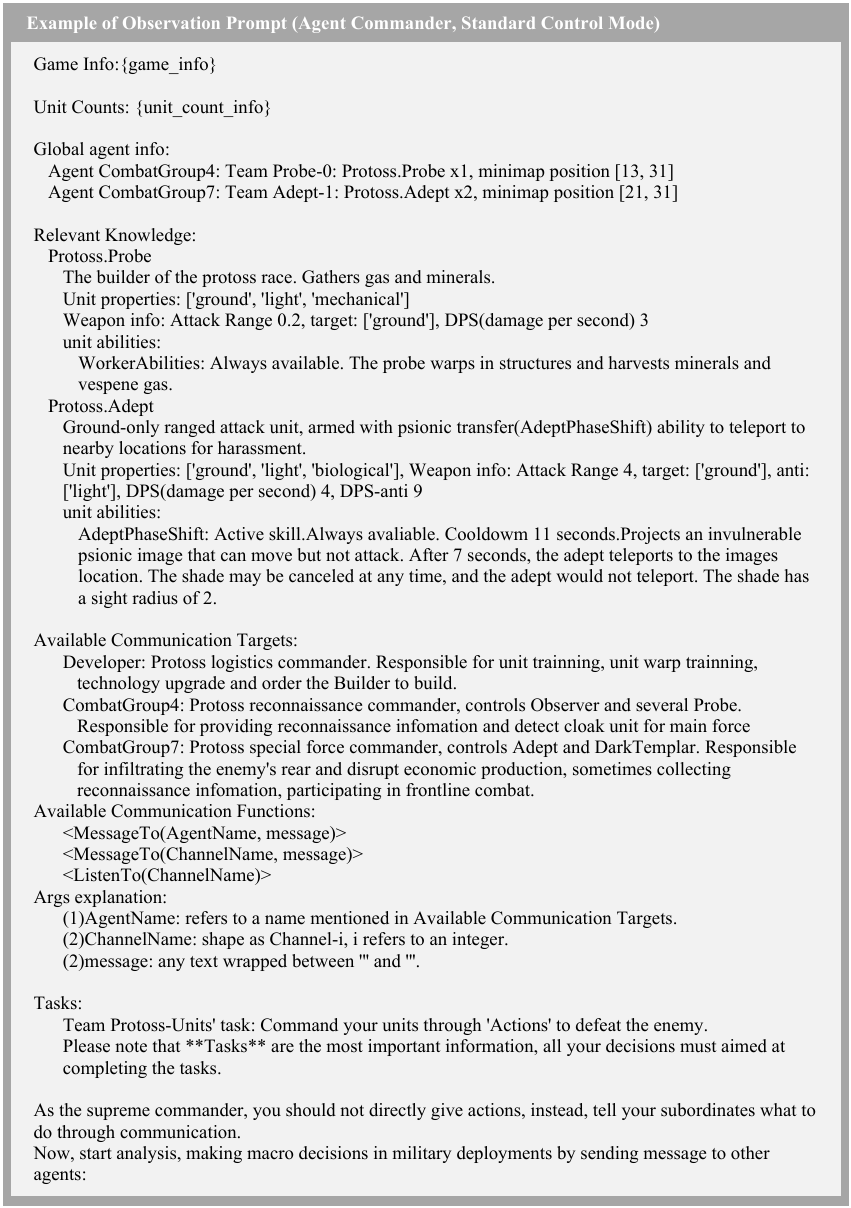}
    \caption{\textbf{Example textual observation of agent 'Commander' in standard control mode.}}
  \label{fig_prompt_c2}
\end{figure}

\begin{figure}[h]
  \centering
  \includegraphics[width=0.95\textwidth]{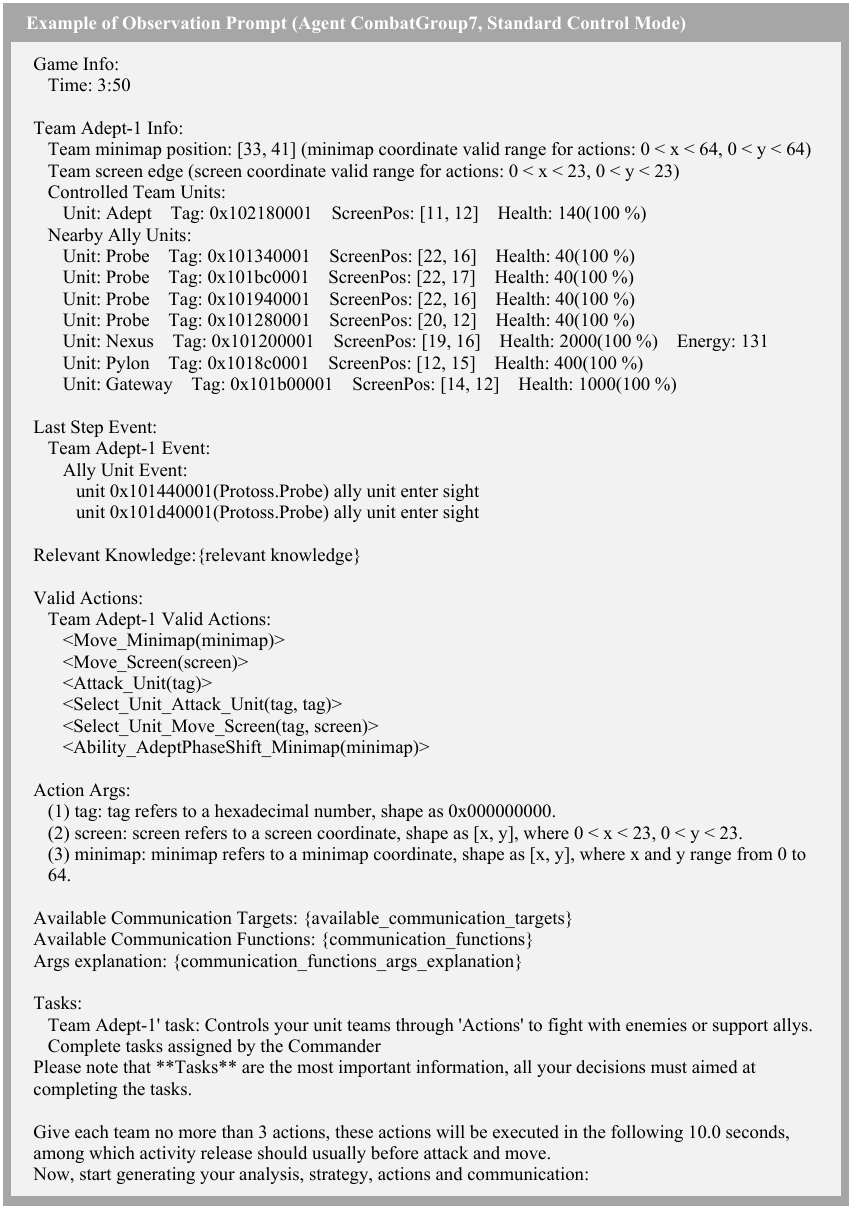}
    \caption{\textbf{Example textual observation of agent 'CombatGroup7' in standard control mode.}}
  \label{fig_prompt_cg}
\end{figure}

\begin{figure}[h]
  \centering
  \includegraphics[width=0.95\textwidth]{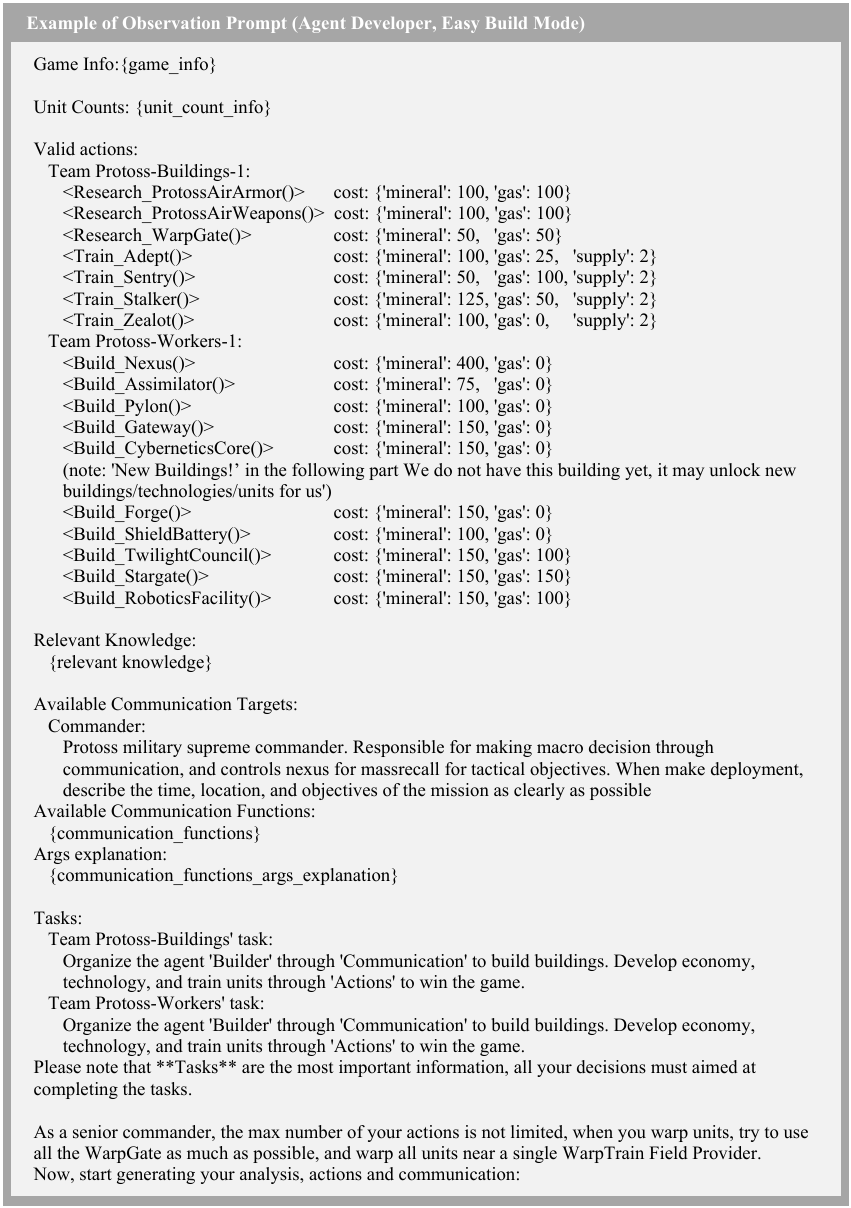}
    \caption{\textbf{Example textual observation of agent 'Developer' in easy build mode.}}
  \label{fig_prompt_d1}
\end{figure}

\begin{figure}[h]
  \centering
  \includegraphics[width=0.95\textwidth]{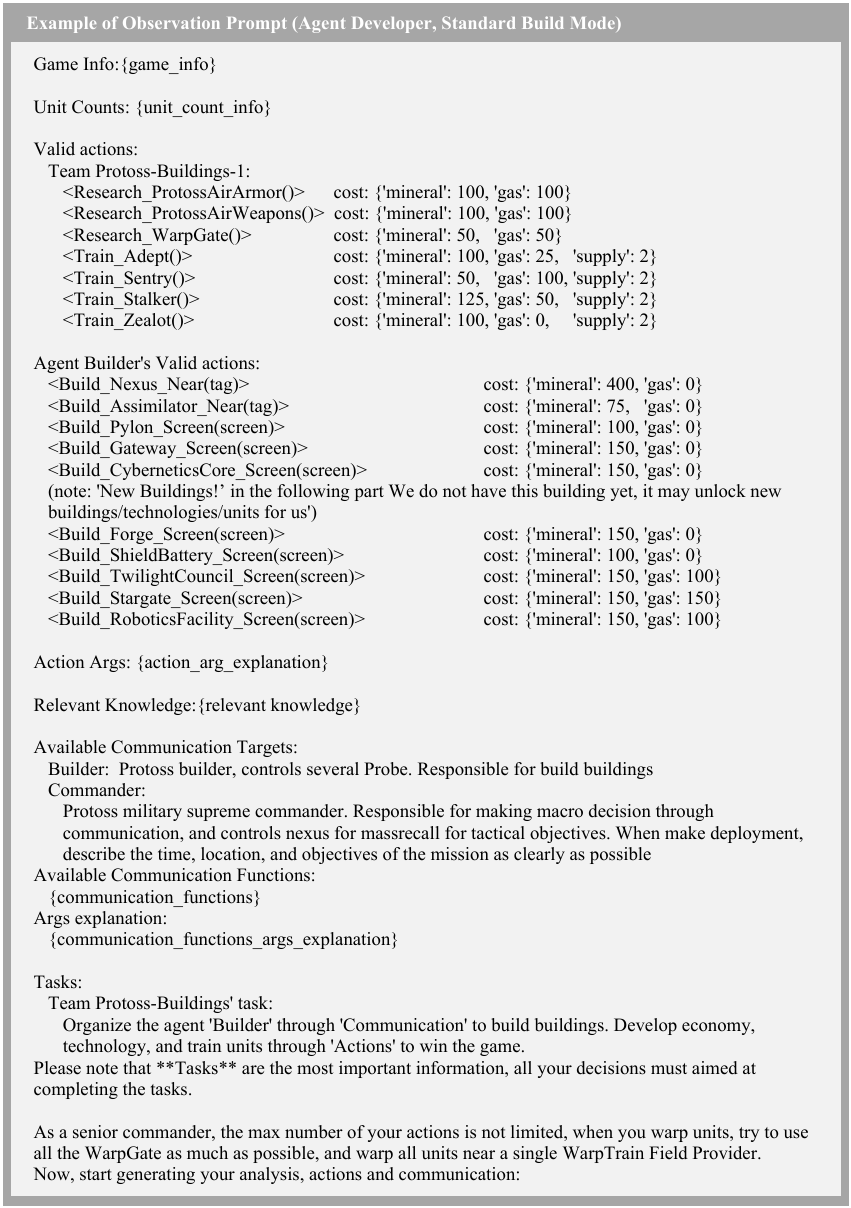}
    \caption{\textbf{Example textual observation of agent 'Developer' in standard build mode.}}
  \label{fig_prompt_d2}
\end{figure}

\begin{figure}[h]
  \centering
  \includegraphics[width=0.95\textwidth]{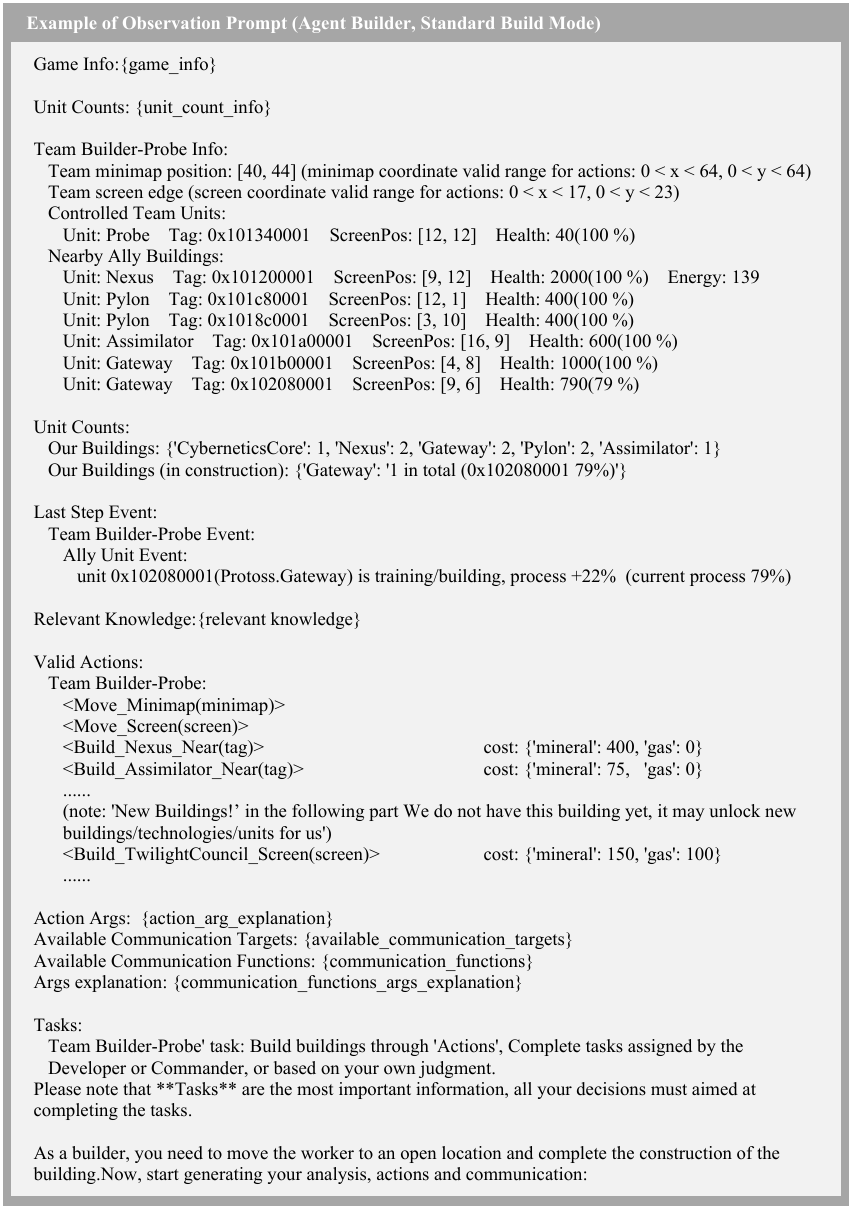}
    \caption{\textbf{Example textual observation of agent 'Builder' in standard build mode.}}
  \label{fig_prompt_b}
\end{figure}

\clearpage
\subsection*{C3. Examples of Image Observation}

\begin{figure}[h]
  \centering
  \includegraphics[width=0.92\textwidth]{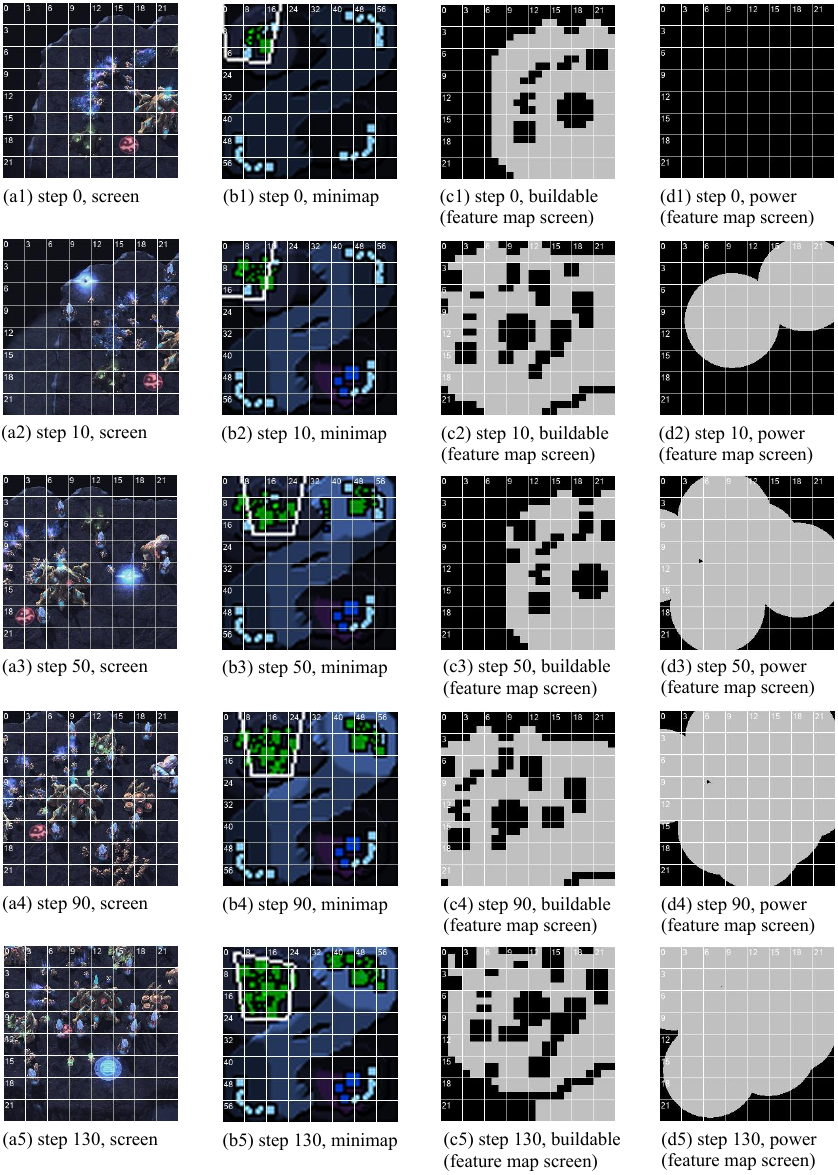}
    \caption{\textbf{Examples of image observation of agent 'Builder' in standard control mode.}}
  \label{fig-img-obs1}
\end{figure}

\begin{figure}[h]
  \centering
  \includegraphics[width=0.92\textwidth]{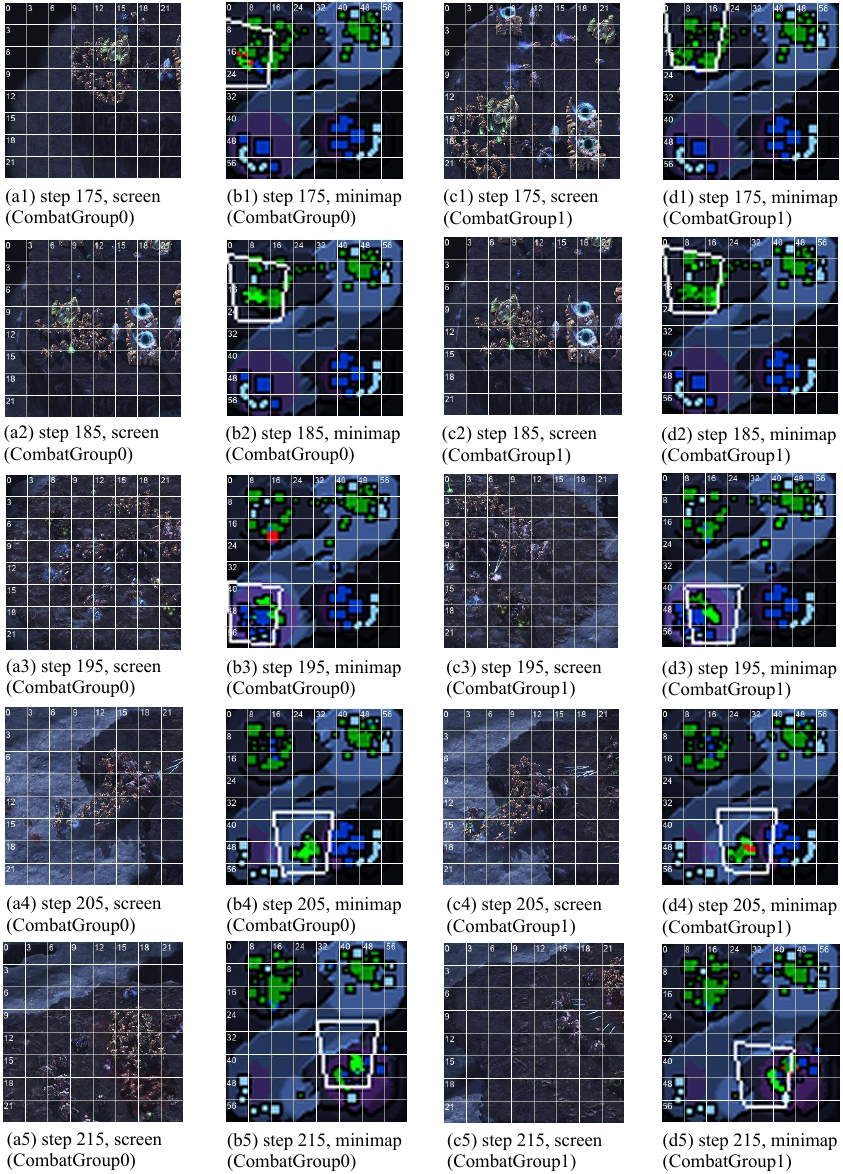}
    \caption{\textbf{Examples of image observation of agent 'CombatGroup0' (controls Zealots) and agent 'CombatGroup1' (controls Stalkers) in standard control mode.}}
  \label{fig-img-obs1}
\end{figure}

\clearpage
\subsection*{C4. Examples of LLM Responses}

\begin{figure}[h]
  \centering
  \includegraphics[width=0.95\textwidth]{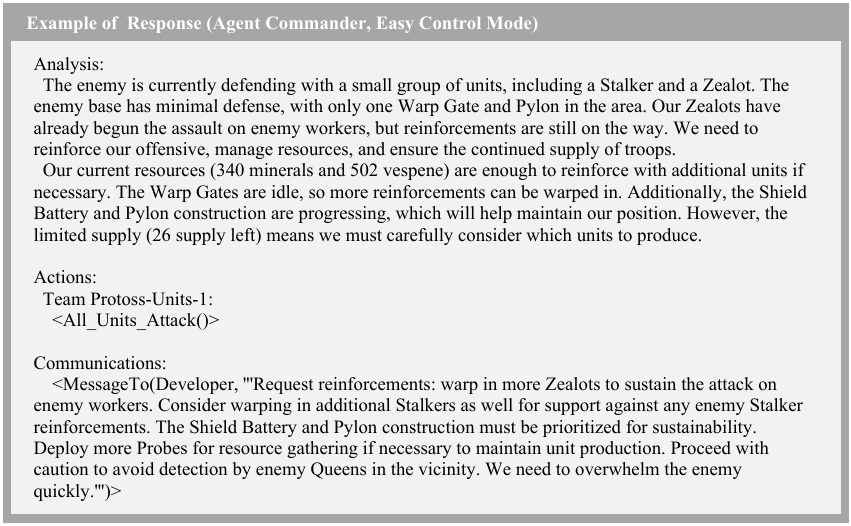}
    \caption{\textbf{Example response of agent 'Commander' in easy control mode.}}
  \label{fig_response_ECEB_1}
\end{figure}

\begin{figure}[h]
  \centering
  \includegraphics[width=0.95\textwidth]{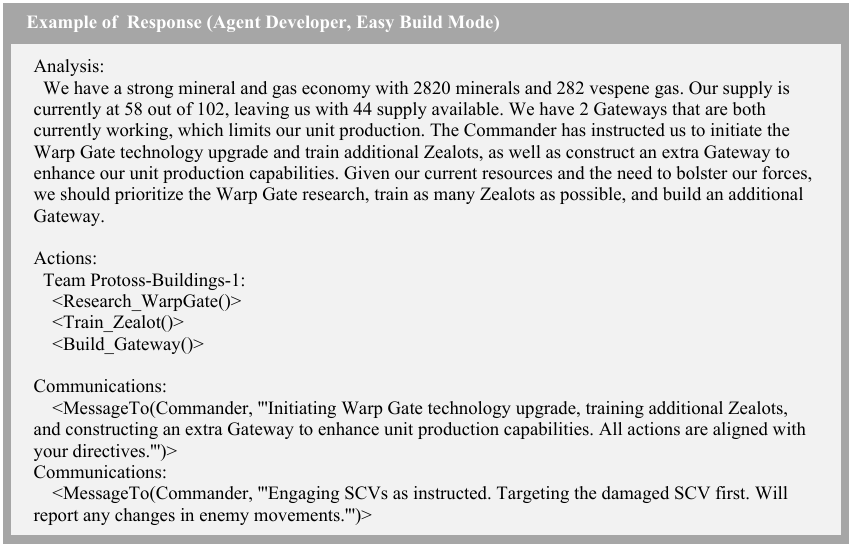}
    \caption{\textbf{Example response of agent 'Developer' in easy build mode.}}
  \label{fig_response_ECEB_1}
\end{figure}

\begin{figure}[h]
  \centering
  \includegraphics[width=0.95\textwidth]{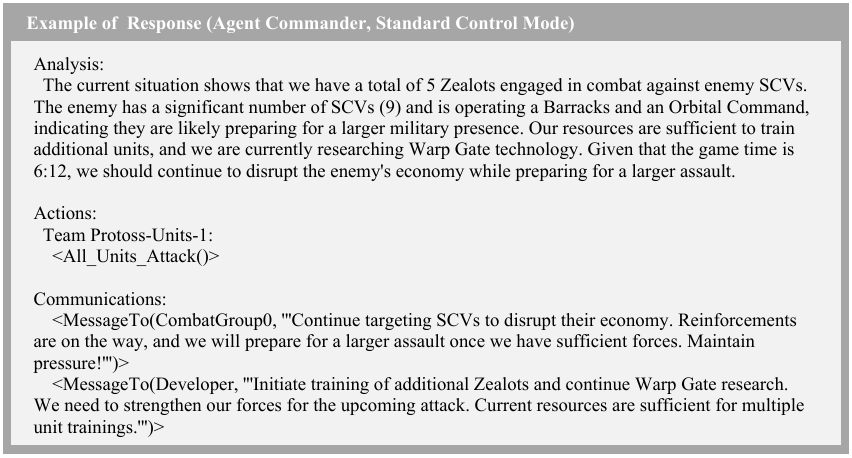}
    \caption{\textbf{Example response of agent 'Commander' in standard control mode.}}
  \label{fig_response_SCEB_1}
\end{figure}

\begin{figure}[h]
  \centering
  \includegraphics[width=0.95\textwidth]{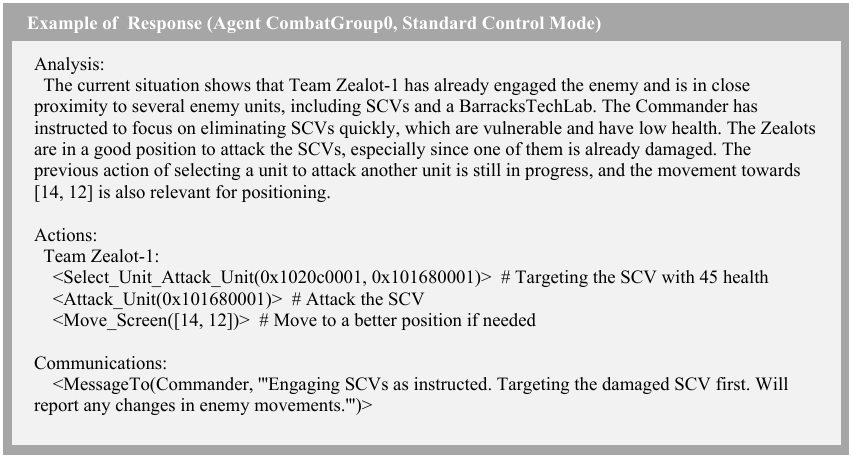}
    \caption{\textbf{Example response of agent 'CombatGroup0' in standard control mode.}}
  \label{fig_response_SCEB_1}
\end{figure}

\begin{figure}[h]
  \centering
  \includegraphics[width=0.95\textwidth]{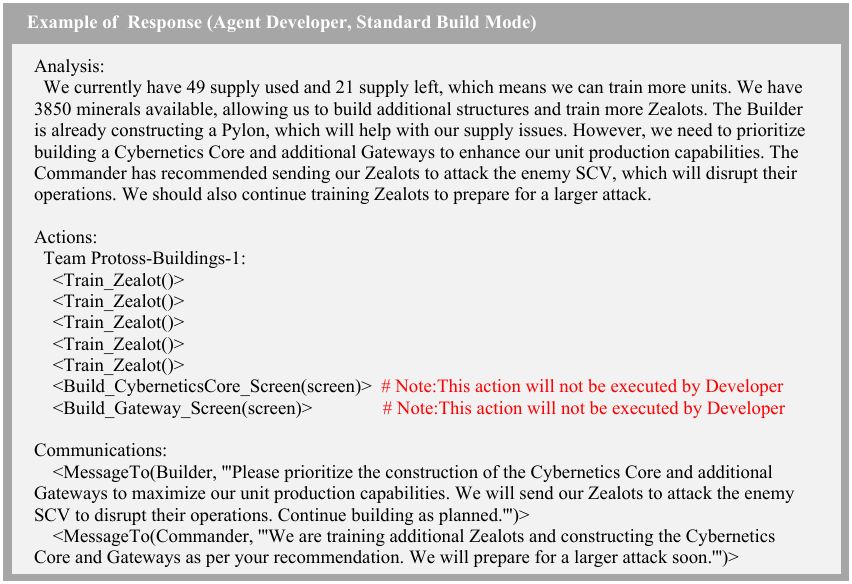}
    \caption{\textbf{Example response of agent 'Developer' in standard build mode.}}
  \label{fig_response_ECSB_1}
\end{figure}

\begin{figure}[h]
  \centering
  \includegraphics[width=0.95\textwidth]{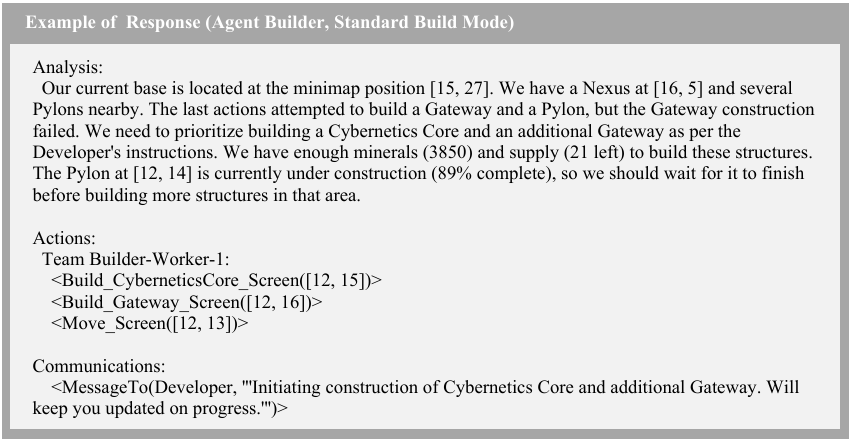}
    \caption{\textbf{Example response of agent 'Builder' in standard build mode.}}
  \label{fig_response_ECSB_2}
\end{figure}

\begin{figure}[h]
  \centering
  \includegraphics[width=0.95\textwidth]{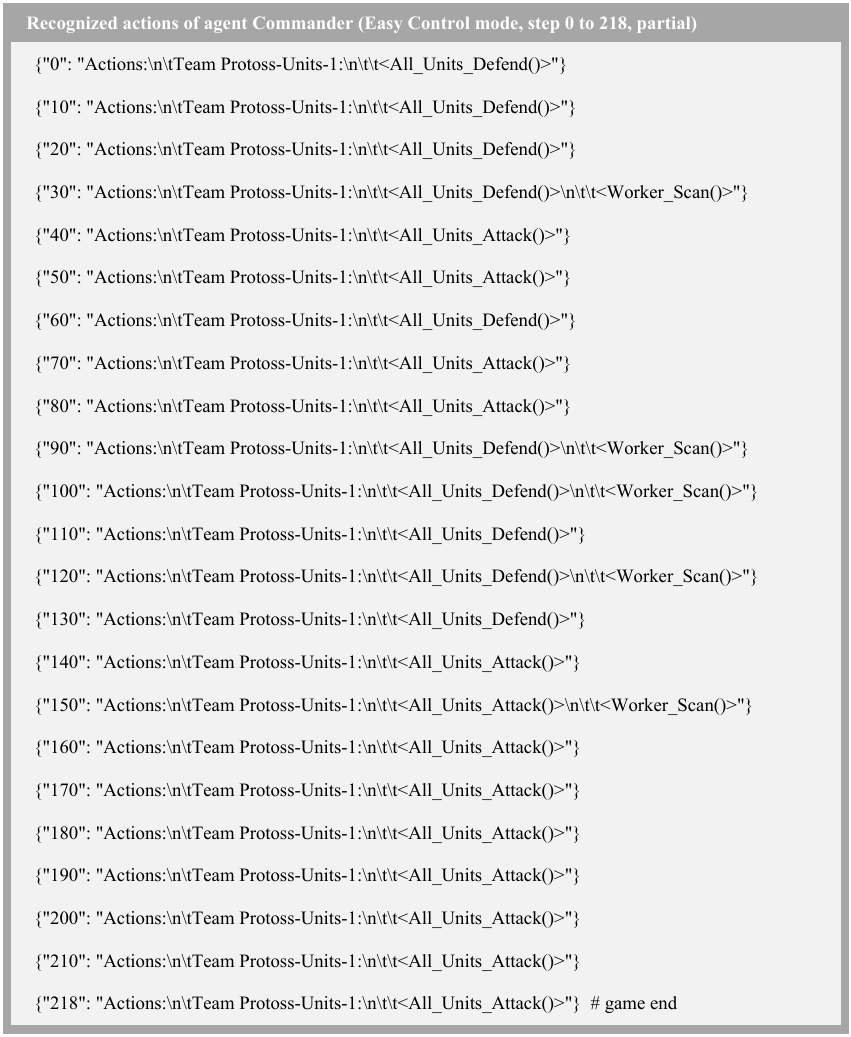}
    \caption{\textbf{Examples of recognized actions of agent Commander.}}
  \label{fig-rec-a-commander}
\end{figure}

\begin{figure}[h]
  \centering
  \includegraphics[width=0.95\textwidth]{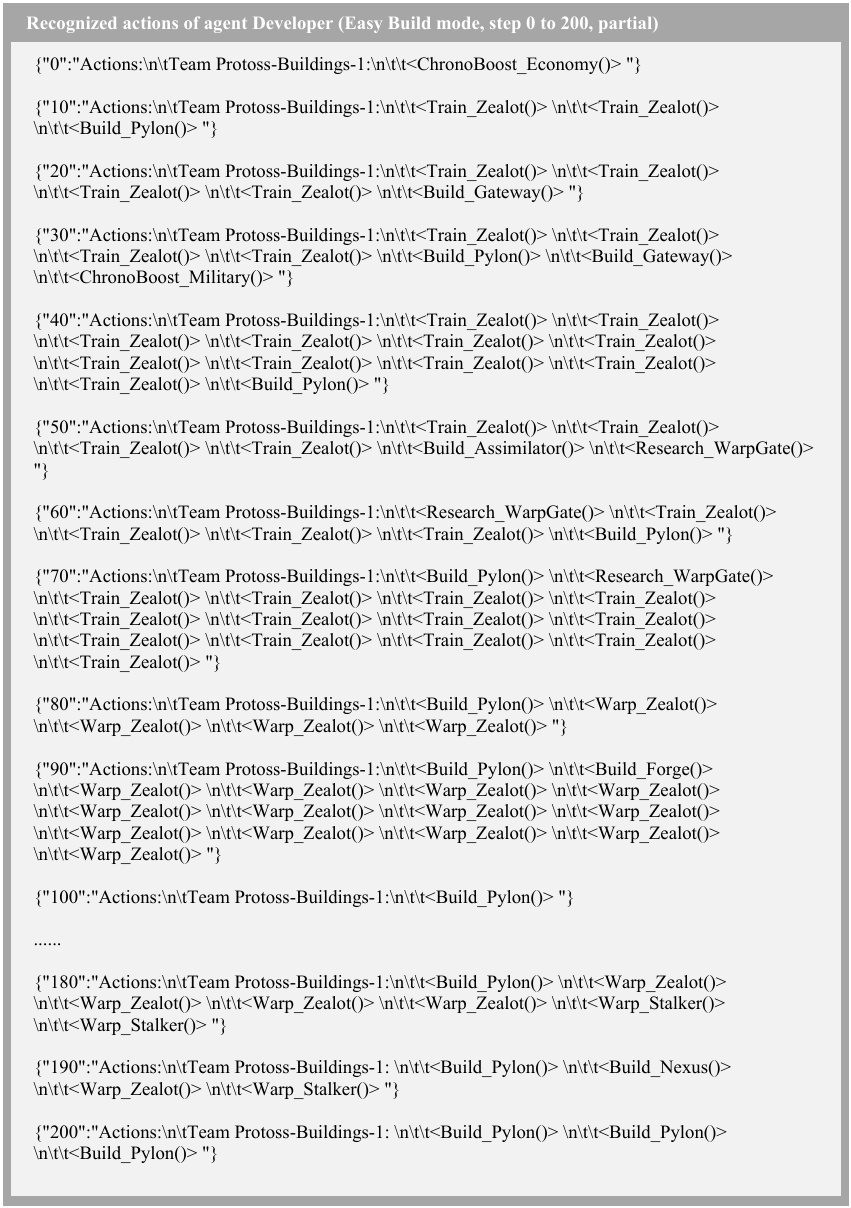}
    \caption{\textbf{Examples of recognized actions of agent Developer.}}
  \label{fig-rec-a-developer}
\end{figure}

\begin{figure}[h]
  \centering
  \includegraphics[width=0.95\textwidth]{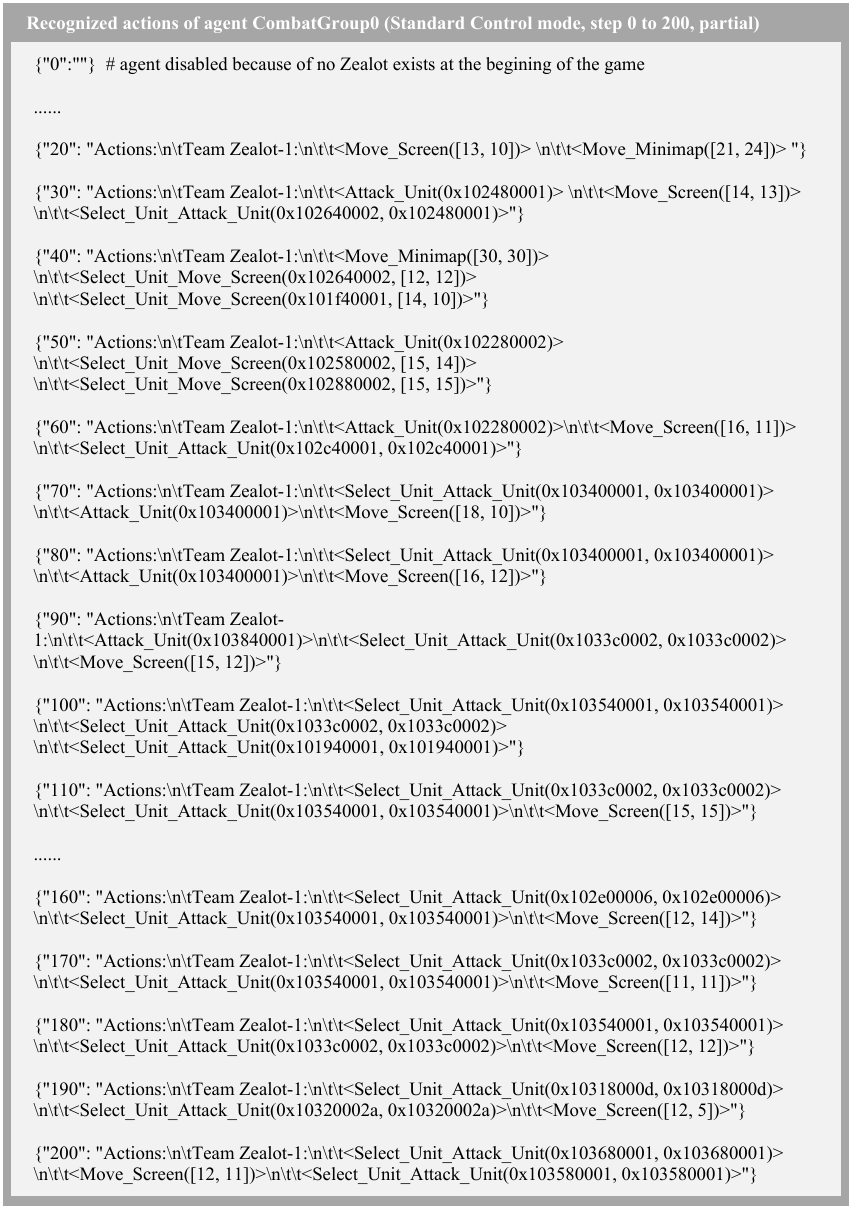}
    \caption{\textbf{Examples of recognized actions of agent Combatgroup0.}}
  \label{fig-rec-a-combatgroup0}
\end{figure}

\begin{figure}[h]
  \centering
  \includegraphics[width=0.95\textwidth]{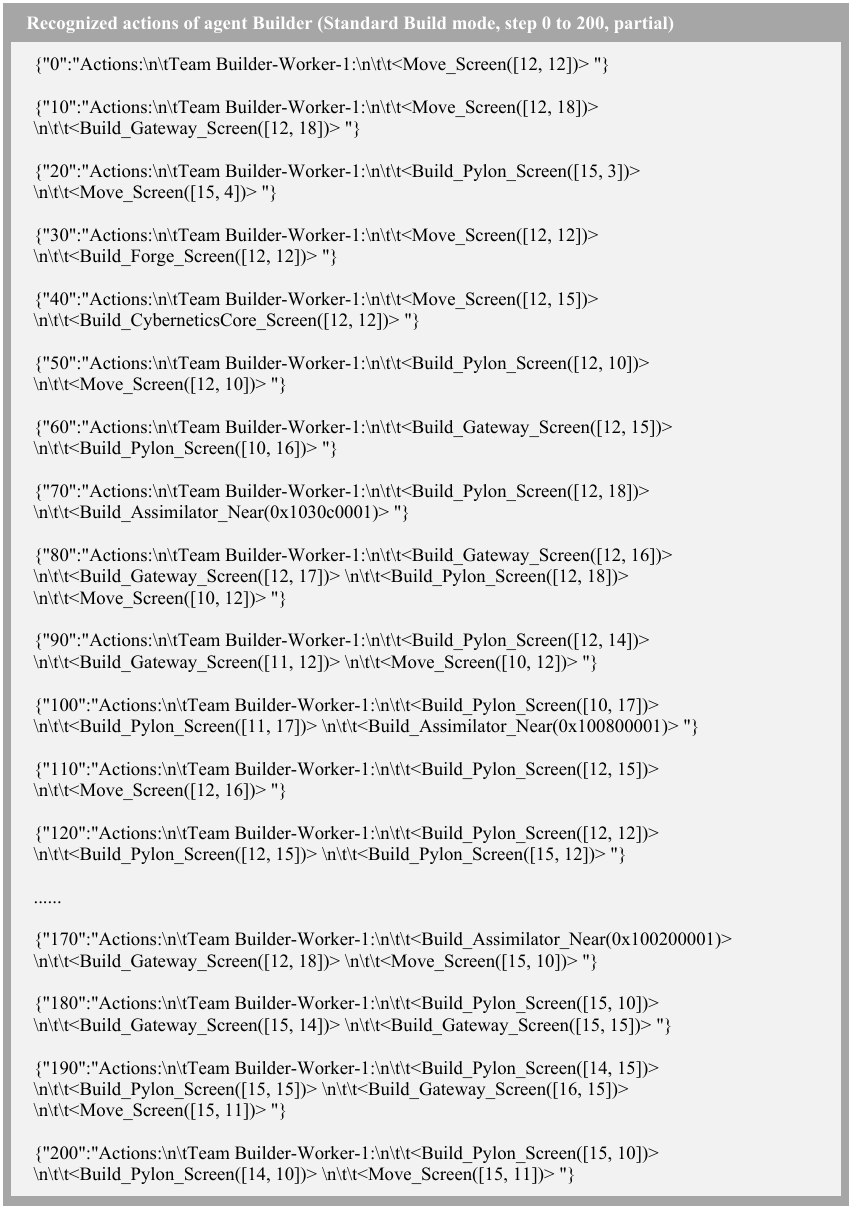}
    \caption{\textbf{Examples of recognized actions of agent Builder.}}
  \label{fig-rec-a-builder}
\end{figure}

\clearpage
\subsection*{C5. Examples of Received Communication Messages}

\begin{figure}[h]
  \centering
  \includegraphics[width=0.92\textwidth]{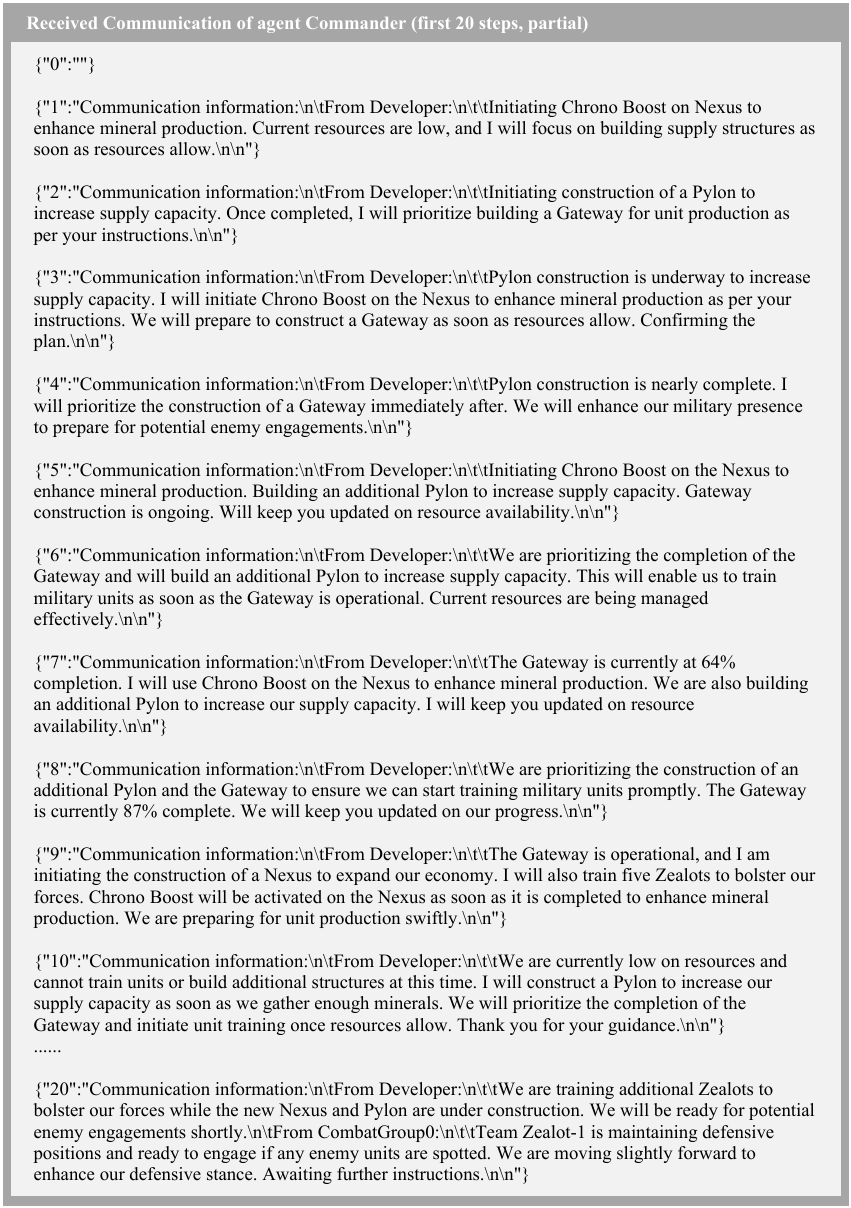}
    \caption{\textbf{Examples of received messages of agent 'Commander' in standard control mode. (part-1)}}
  \label{fig-c-in-commander1}
\end{figure}

\begin{figure}[h]
  \centering
  \includegraphics[width=0.95\textwidth]{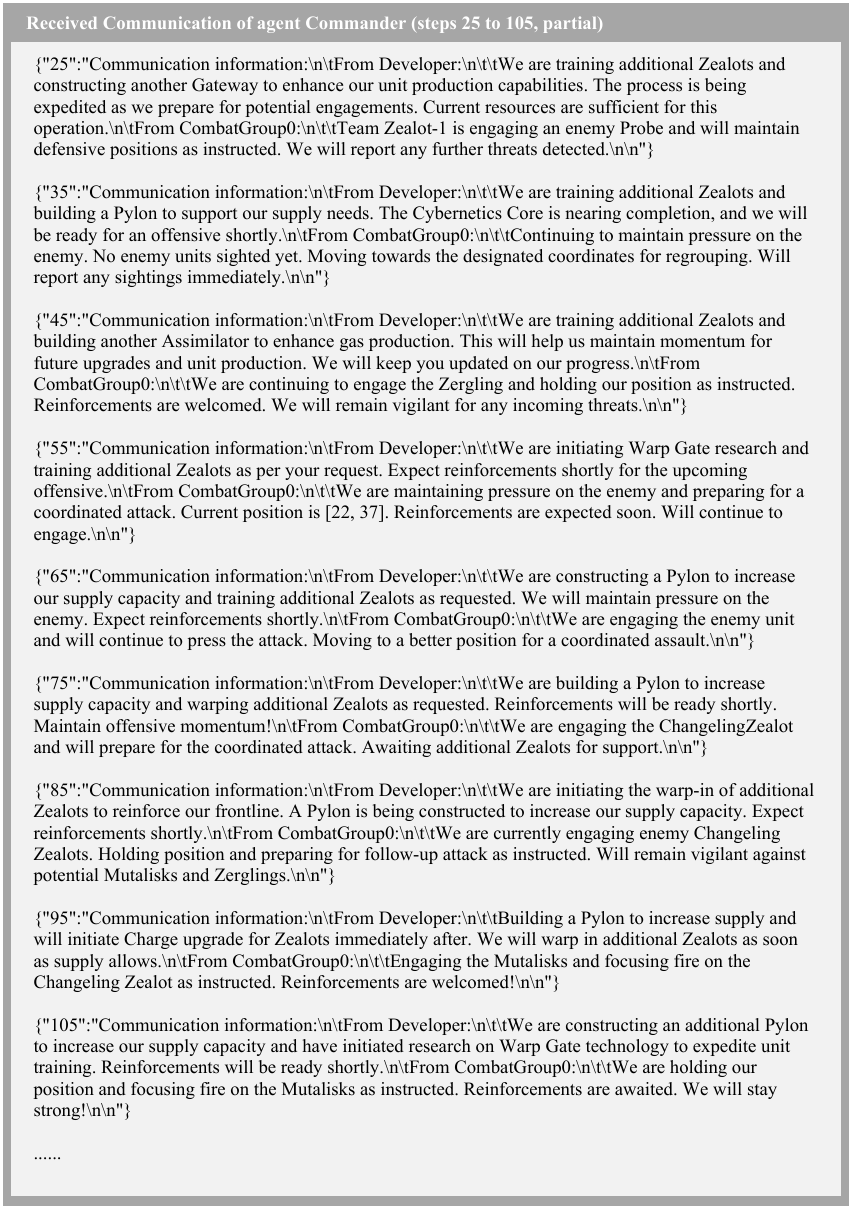}
    \caption{\textbf{Examples of received messages of agent 'Commander' in standard control mode. (part-2)}}
  \label{fig-c-in-commander2}
\end{figure}

\begin{figure}[h]
  \centering
  \includegraphics[width=0.95\textwidth]{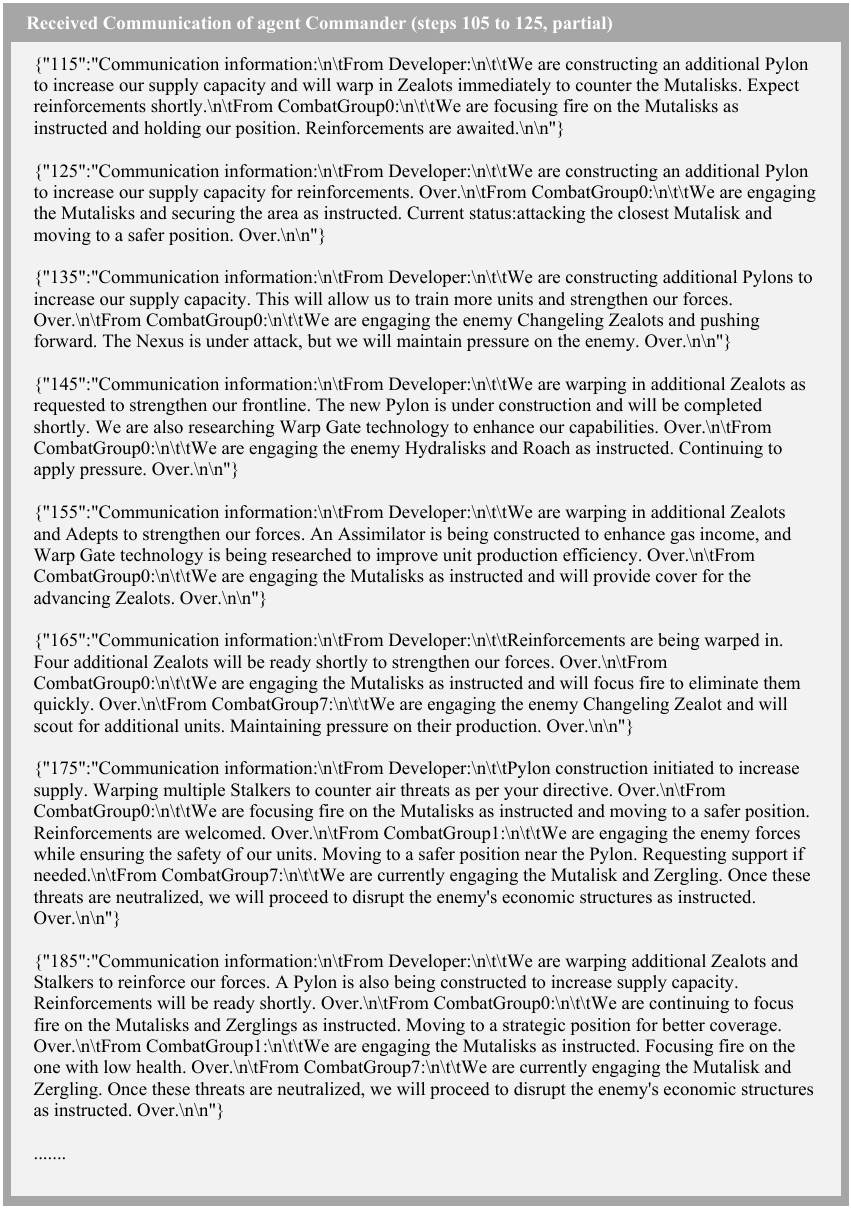}
    \caption{\textbf{Examples of received messages of agent 'Commander' in standard control mode. (part-3)}}
  \label{fig-c-in-commander3}
\end{figure}

\clearpage
\subsection*{C6. Examples of Communication Actions}

\begin{figure}[h]
  \centering
  \includegraphics[width=0.95\textwidth]{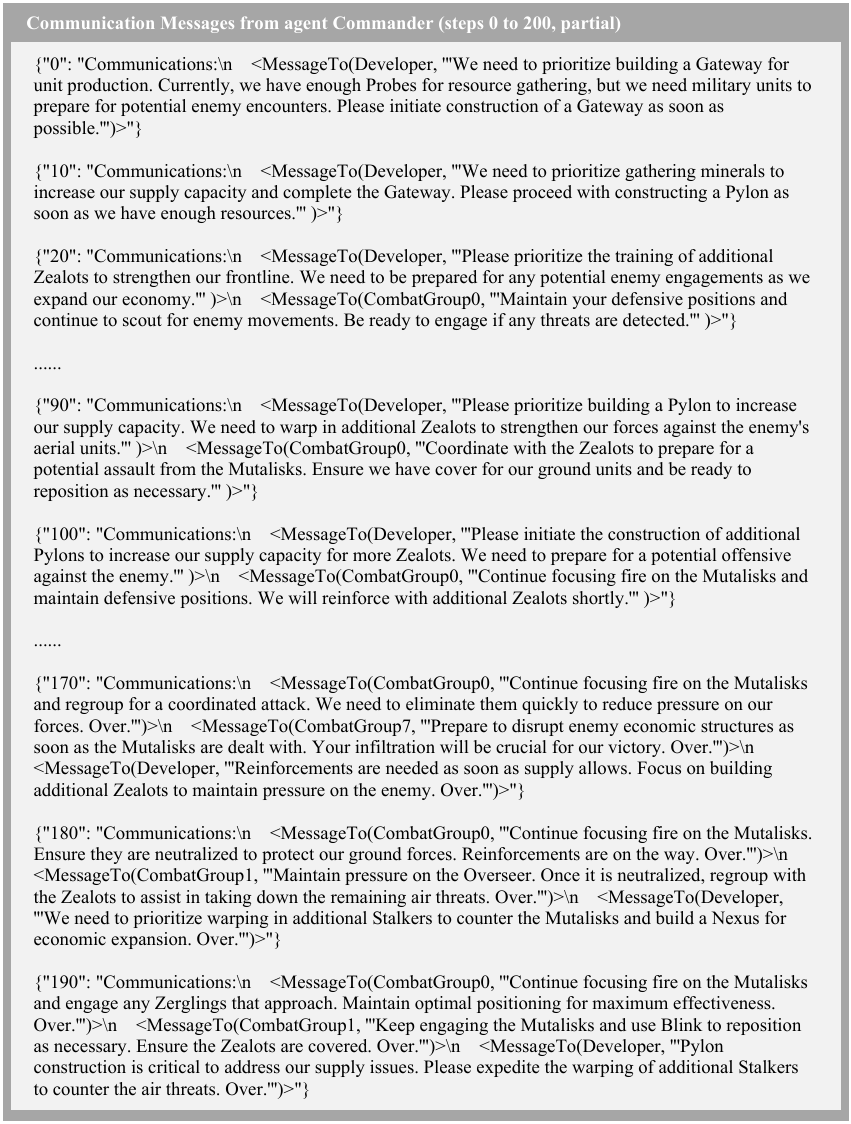}
    \caption{\textbf{Examples of sent messages of agent 'Commander' in standard control mode.}}
  \label{fig-c-out-commander}
\end{figure}

\begin{figure}[h]
  \centering
  \includegraphics[width=0.95\textwidth]{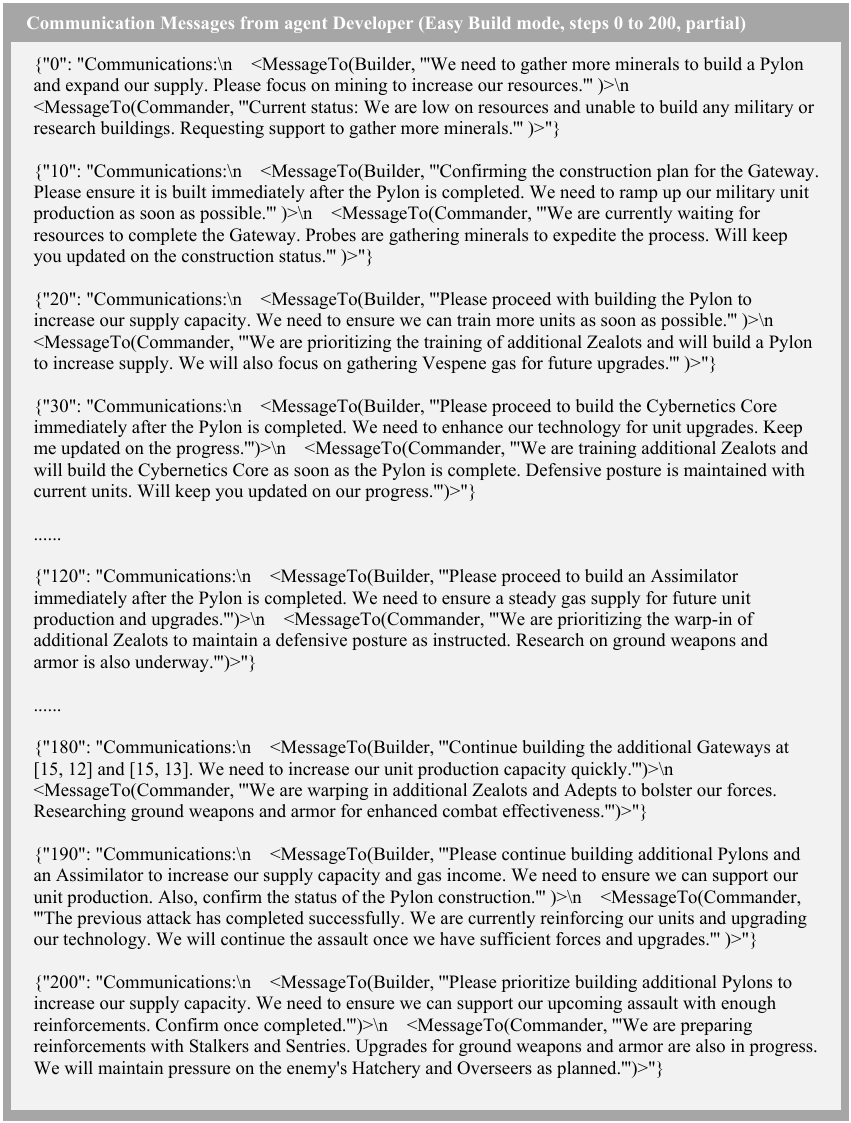}
    \caption{\textbf{Examples of sent messages of agent 'Developer' in standard control mode.}}
  \label{fig-c-out-developer}
\end{figure}

\clearpage
\section*{Appendix D.  Experimental Settings with Multi-Agent Settings}
\setcounter{figure}{0}
\setcounter{table}{0}
\renewcommand{\thefigure}{D\arabic{figure}}
\renewcommand{\thetable}{D\arabic{table}}

\begin{table}[h]
\vspace{0.2cm}
\caption{System settings}\label{tabd1}
\begin{center}
\vspace{-0.2cm}
\small
\renewcommand\arraystretch{1.2}
\begin{tabular}{p{3cm} p{5.2cm}  p{4.2cm}}
\toprule
Module & Recommand    & Minimum requirements\\
\midrule
System  & Windows-10 or 11 & Windows-10\\
CPU  & i9-14900, 24 cores 32 threads & 8 core\\
GPU  & GeForce RTX 4090, 24G & GeForce GTX 1080\\
Storage  & 64G RAM +2T SSD & 8G RAM + 100G SSD\\
Starcraft II & Version 9.0.14(93333) & Version 9.0.14(93333)\\
\bottomrule
\end{tabular}
\end{center}
\end{table}

\begin{table}[h]
\caption{Multi-agent settings for complete game and LLM-PySC task group}\label{tabd2}
\begin{center}
\small
\renewcommand\arraystretch{1.2}
\begin{tabular}{p{2.5cm} p{2.5cm}  p{7cm}}
\toprule
Agent name & Unit team names   & Details of each unit team\\
\midrule
Commander & Protoss-Units       & A virtual team, enable in easy control mode, directly control all combat units attack, defend or retreat, or call for scan. But unable to use skills or precise control. \\
\midrule
Developer & Protoss-Buildings   & A virtual team, always enable, available for unit training/warping and technology upgrade actions. In easy build mode, this team also available for build building. \\
\midrule
Builder   & Builder-Probe-1     & Enable in standard build mode. Controls Probes. \\
\midrule
CombatGroup0  & Zealot-1        & Enable in standard control mode. Controls Zealots. \\
\midrule
CombatGroup1  & Stalker-1       & Enable in standard control mode. Controls Stalkers. \\
\midrule
CombatGroup2  & Immortal-1      & Enable in standard control mode. Controls Immortal. \\
              & Colossus-1      & Enable in standard control mode. Controls Colossus. \\
              & Archon-1        & Enable in standard control mode. Controls Archon. \\
\midrule
CombatGroup3  & VoidRay-1       & Enable in standard control mode. Controls Void-Ray. \\
              & Carrier-1       & Enable in standard control mode. Controls Carrier. \\
              & Tempest-1       & Enable in standard control mode. Controls Tempest. \\
\midrule
CombatGroup4  & Observer-1      & Enable in standard control mode. Controls Observer. \\
\midrule
CombatGroup5  & HighTemplar-1   & Enable in standard control mode. Controls HighTemplar. \\
              & Disruptor-1     & Enable in standard control mode. Controls Disruptor. \\
\midrule
CombatGroup6  & Sentry-1        & Enable in standard control mode. Controls Sentry. \\
              & Mothership-1    & Enable in standard control mode. Controls Mothership. \\
\midrule
CombatGroup7  & Adept-1         & Enable in standard control mode. Controls Adept. \\
              & AdeptPhase-1    & Enable in standard control mode. Controls AdeptPhase. \\
              & DarkTemplar-1   & Enable in standard control mode. Controls DarkTemplar. \\
\midrule
CombatGroup8  & Oracle-1        & Enable in standard control mode. Controls Oracle. \\
              & Phoenix-1       & Enable in standard control mode. Controls Phoenix. \\
\midrule
CombatGroup9  & WarpPrism-1     & Enable in standard control mode. Controls WarpPrism. \\
  
\bottomrule
\end{tabular}
\end{center}
\end{table}

\begin{table}[h]
\caption{Agent settings in LLM-SMAC tasks}\label{tabd3}
\begin{center}
\small
\renewcommand\arraystretch{1.2}
\begin{tabular}{p{2.0cm} p{10.5cm}}
\toprule
Tasks    & Details of each agent\\
\midrule
3s\_vs\_nz  & 1 agent with 1 teams: Team Stalker-1(3x Stalker) \\
\midrule
2c\_vs\_64zg   & 1 agent with 2 teams: Team Colossus-1(1x Colossus), Team Colossus-2(1x Colossus) \\
\midrule
2s\_vs\_1sc  & 1 agent with 2 teams: Team Stalker-1(1x Stalker), Team Stalker-2(1x Stalker) \\
\midrule
2s3z  & 1 agent with 3 teams: Team Zealot-1 (2x Zealot), Team Zealot-2 (1x Zealot), Team Stalker-1 (2 Stalker) \\
\midrule
3s5z  & 1 agent with 4 teams: Team Zealot-1 (2x Zealot), Team Zealot-2 (2x Zealot), \\
3s5z\_vs\_3s6z & Team Zealot-3 (1x  Zealot), Team Stalker-1 (3 Stalker)\\
\midrule
1c3s5z   & 1 agent with 5 teams: Team Zealot-1 (2x Zealot), Team Zealot-2 (2x Zealot), Team Zealot-3 (1x  Zealot), Team Stalker-1 (3 Stalker), Team Colossus-1 (1x Colossus)\\
  
\bottomrule
\end{tabular}
\end{center}
\end{table}

\begin{table}[h]
\caption{Victory conditions and evaluated aspect of LLM-PySC2 tasks level-1}\label{tabd4}
\begin{center}
\small
\renewcommand\arraystretch{1.2}
\begin{tabular}{p{1.7cm} p{1.5cm} p{4.2cm} p{4.4cm}}
\toprule
Task name & max time &  victory condition & evaluated aspect\\
\midrule
task1      & 1min & kill at least 7 workers & task understanding, unit skills\\
task2      & 1min & kill at least 7 workers & task understanding, unit skills\\
task3      & 1min & defend all the airdrops and save more than 6 workers & task understanding, memory\\
task4      & 1min & defeat enemy units & unit skills, multi agent cooperation\\
task5      & 1min & defeat enemy units & unit skills, multi agent cooperation\\
task6      & 1min & defeat enemy units & unit skills, multi agent cooperation\\
task7      & 1min & defeat enemy units & unit skills, multi agent cooperation\\
task8      & 1.5min & defeat enemy units and kill at least 7 workers & unit skills, multi agent cooperation, communication, planning\\
\bottomrule
\end{tabular}
\end{center}
\end{table}

\begin{table}[h]
\caption{Unit settings of LLM-PySC2 tasks level-1}\label{tabd5}
\begin{center}
\vspace{-0.2cm}
\small
\renewcommand\arraystretch{1.2}
\begin{tabular}{p{2.5cm} p{5.0cm} p{5.0cm}}
\toprule
 Task name &  Controlled  & Enemy \\
\midrule
task1 (2a\_harass) & 2 Adapt   & 2 Queen + 12 Drone  \\
\midrule
task2 (3ph\_harass) & 3 Phoenix & 2 Queen + 12 Drone  \\
\midrule
task3 (6s\_defend)    & 6 Stalker  & 4x2 OverlordTransport with     \\
   &    & several Zergling / Baneling  \\
\midrule
task4 (12s\_combat) & 12 Stalker & 15 Roach\\

\midrule
task5                    & 2 Colossus + 3 Disruptor + 4 Sentry +  & 24 Roach + 9 Ravagers + 2 Queen\\
 (3d\_ma\_combat)     & 12 Stalkers &  \\
\midrule
task6    & 1 Archon + 6 HighTemplar + 4 Sentry + & 64 Zergling + 32 Banelings + 1 Ultralisk\\
(6h\_ma\_combat)                          & 12 Stalkers &  \\
\midrule
task7   & 1 Mothership + 3 Carrier + 3 Tempest + & 18 Hydralisk + 7 Corruptor +  \\ 
(1m\_ma\_combat) & 6 Void-Ray + 12 Stalkers  & 4 BoordLord + 3 Viper\\
\midrule
task8 & 2 Warpprism + 8 Warpgate + & 15 Roach + 3 Ravager + 4 Queen.\\
(8bg\_ma\_combat)    & 12 Stalker + 1600 minerals & \\

\bottomrule
\end{tabular}
\end{center}
\end{table}

\begin{table}[h]
\caption{Details of LLM-PySC2 tasks from level-1 to level-3 }\label{tabd4}
\begin{center}
\small
\renewcommand\arraystretch{1.2}
\begin{tabular}{p{3.0cm} p{1.5cm} p{9cm}}
\toprule
 Task name & Difficulty &  Important changes \\
\midrule
task1 (2a\_harass) & level-1  & Adept upgrade enabled (+45\% attack speed). \\
& & Enemy 2 Queens.  \\
 & level-2 & Adept upgrade enabled (+45\% attack speed). \\
& & Enemy 2 Queens with 4 Zerglings.  \\
 & level-3 & Adept upgrade disabled. \\
& & Enemy 2 Queens with 4 Zerglings.  \\
            
\midrule
task2 (3ph\_harass) & level-1 & Phoenix upgrade enabled (+2 attack range). \\
& & Enemy 2 Queens. \\
 & level-2 & Phoenix upgrade enabled (+2 attack range). \\
& & Enemy 2 Queens, with 1 Spore Crawler.\\
 & level-3 & Phoenix upgrade disabled. \\
& & Enemy 2 Queens, with 1 Spore Crawler.\\

\midrule
task3 (6s\_defend) & level-1 & One PhotonCannon helps for anti-air combat. \\
& & Enemy OverlordTransport no upgrade. \\
 & level-2 & One PhotonCannon helps for anti-air combat. \\
& & Enemy OverlordTransport upgrade enable (higher speed). \\
 & level-3 & No PhotonCannon helps for anti-air combat. \\
& & Enemy OverlordTransport upgrade enable (higher speed speed). \\

\midrule
task4 (12s\_combat) & level-1 & Enemy 15 Roach, 1 Ravager. \\
 & level-2 & Enemy 15 Roach, 2 Ravager, 1 Queen.\\
 & level-3 & Enemy 15 Roach. 3 Ravager, 2 Queen, 1 Overseer.\\

\midrule
task5 (3d\_ma\_combat) & level-1 & Enemy 24 Roach, 9 Ravager, 2 Queen. \\
 & level-2 & Enemy 24 Roach, 9 Ravager, 2 Queen, 1 Ultralisk.\\
 & level-3 & Enemy Enemy 24 Roach. 9 Ravager, 2 Queen, 1 Ultralisk, 2 SwarmHost. \\

\midrule
task6 (6h\_ma\_combat) & level-1 & Enemy 64 Zergling, 32 Banelings, 1 Ultralisk. \\
 & level-2 & Enemy 64 Zergling, 32 Banelings, 3 Ultralisk. \\
 & level-3 & Enemy 64 Zergling, 32 Banelings, 3 Ultralisk, 4 Queen. \\

\midrule
task7 (1m\_ma\_combat) & level-1 & Enemy 18 Hydralisk, 7 Corruptor, 4 BoordLord, 3 Viper. \\

 & level-2 & Enemy 18 Hydralisk, 7 Corruptor, 4 BoordLord, 3 Viper, \\
 & & 4 Queen, 2 Infestor.\\
 & level-3 & Enemy 21 Hydralisk, 9 Corruptor, 6 BoordLord, 3 Viper, \\
 & & 4 Queen, 2 Infestor.\\

\midrule
task8 (8bg\_ma\_combat) & level-1 & Controls 2 WarpPrism, 8 WarpGates, 1600 minerals. \\
& & Enemy 15 Roach. 3 Ravager, 4 Queen. \\
 & level-2 & 2 WarpPrism, 8 WarpGates, 1600 minerals. \\
& & Enemy 15 Roach. 3 Ravager, 4 Queen, 3 Spore Crawler.\\
 & level-3 & 1 WarpPrism, 8 WarpGates, 1600 minerals. \\ 
& & Enemy 15 Roach. 3 Ravager, 4 Queen, 3 Spore Crawler.\\

\bottomrule
\end{tabular}
\end{center}
\end{table}

\clearpage
\section*{Appendix E. Examples of the problems in LLM decision-making}
\setcounter{figure}{0}
\setcounter{table}{0}
\renewcommand{\thefigure}{E\arabic{figure}}
\renewcommand{\thetable}{E\arabic{table}}

\subsection*{E.1 Hallucination Examples in Complete StarCraft II Games (standard contorl mode)}

\begin{figure}[h]
  \centering
  \includegraphics[width=0.95\textwidth]{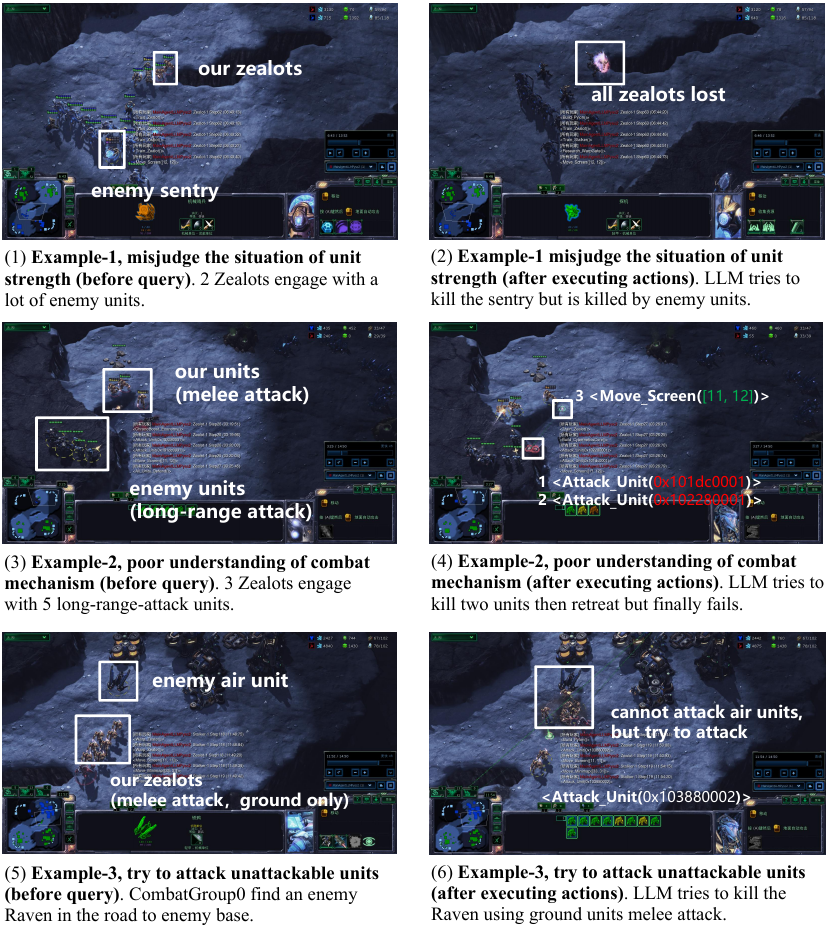}
    \caption{\textbf{Hallucination Examples in Complete StarCraft II Games (combat).}}
  \label{fig-example-hallucination1}
\end{figure}

\begin{figure}[h]
  \centering
  \includegraphics[width=0.95\textwidth]{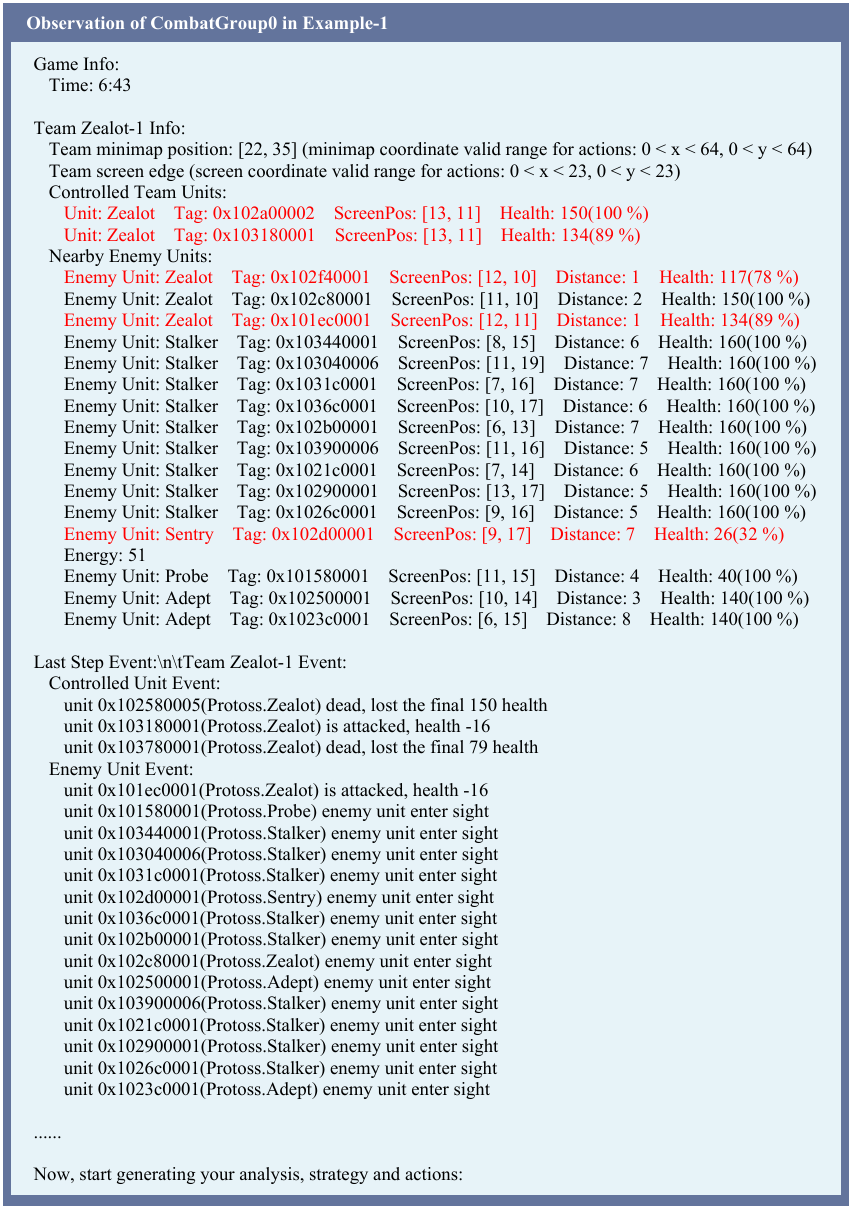}
    \caption{\textbf{Observation of CombatGroup0 in Example-1.}}
  \label{fig-example-hallucination-e1-p1}
\end{figure}

\begin{figure}[h]
  \centering
  \includegraphics[width=0.95\textwidth]{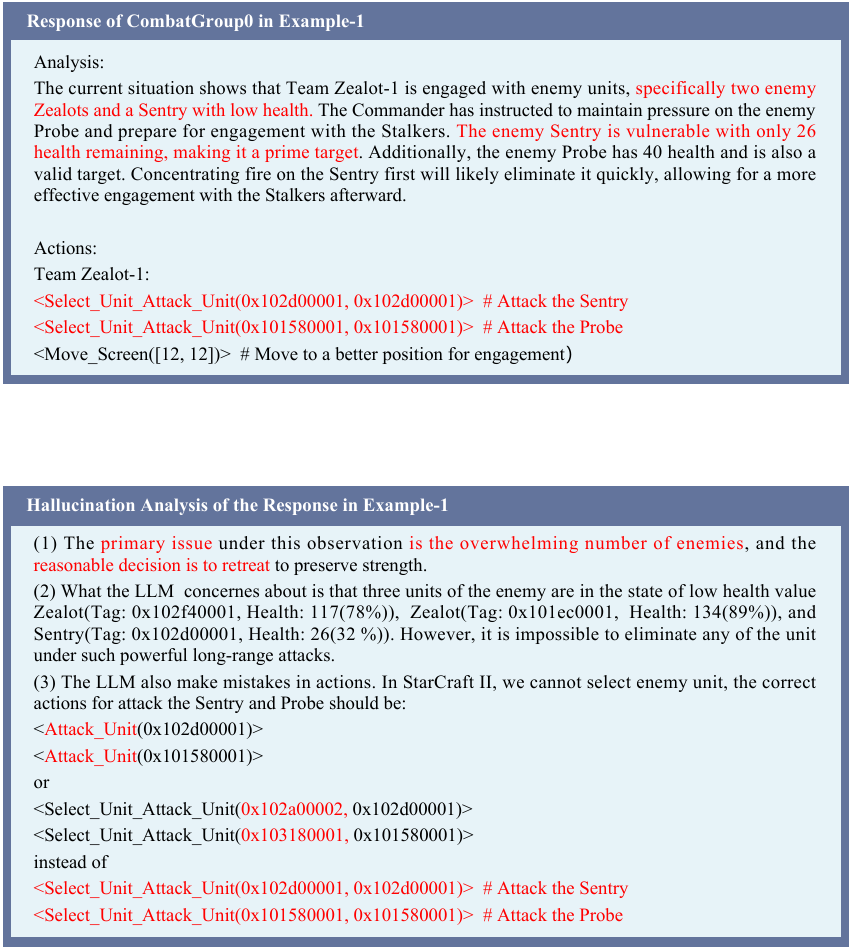}
    \caption{\textbf{Response and Hallucination Analysis of CombatGroup0 in Example-1}}
  \label{fig-example-hallucination-e1-p2}
\end{figure}

\clearpage
\subsection*{E.2 Hallucination Examples in Complete StarCraft II Games (develop and build)}

\begin{figure}[h]
  \centering
  \includegraphics[width=0.95\textwidth]{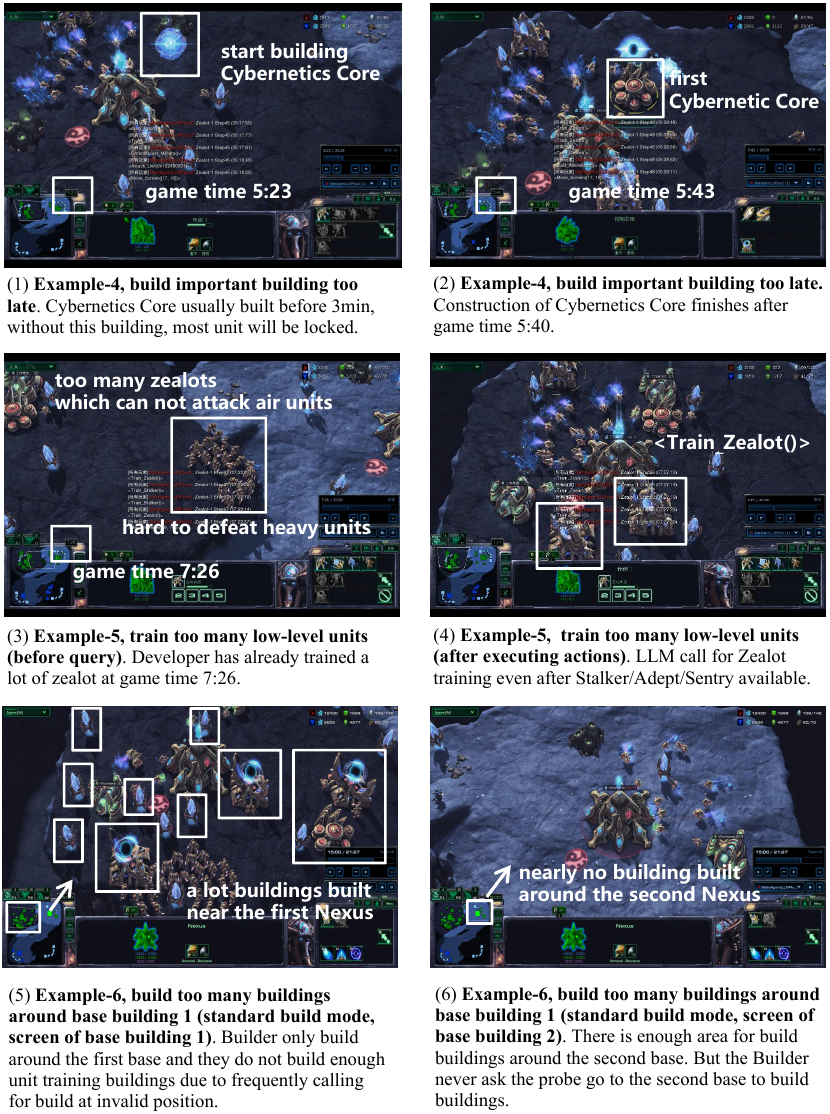}
    \caption{\textbf{Hallucination Examples in Complete StarCraft II Games (standard build mode).}}
  \label{fig-example-hallucination2}
\end{figure}

\begin{figure}[h]
  \centering
  \includegraphics[width=0.95\textwidth]{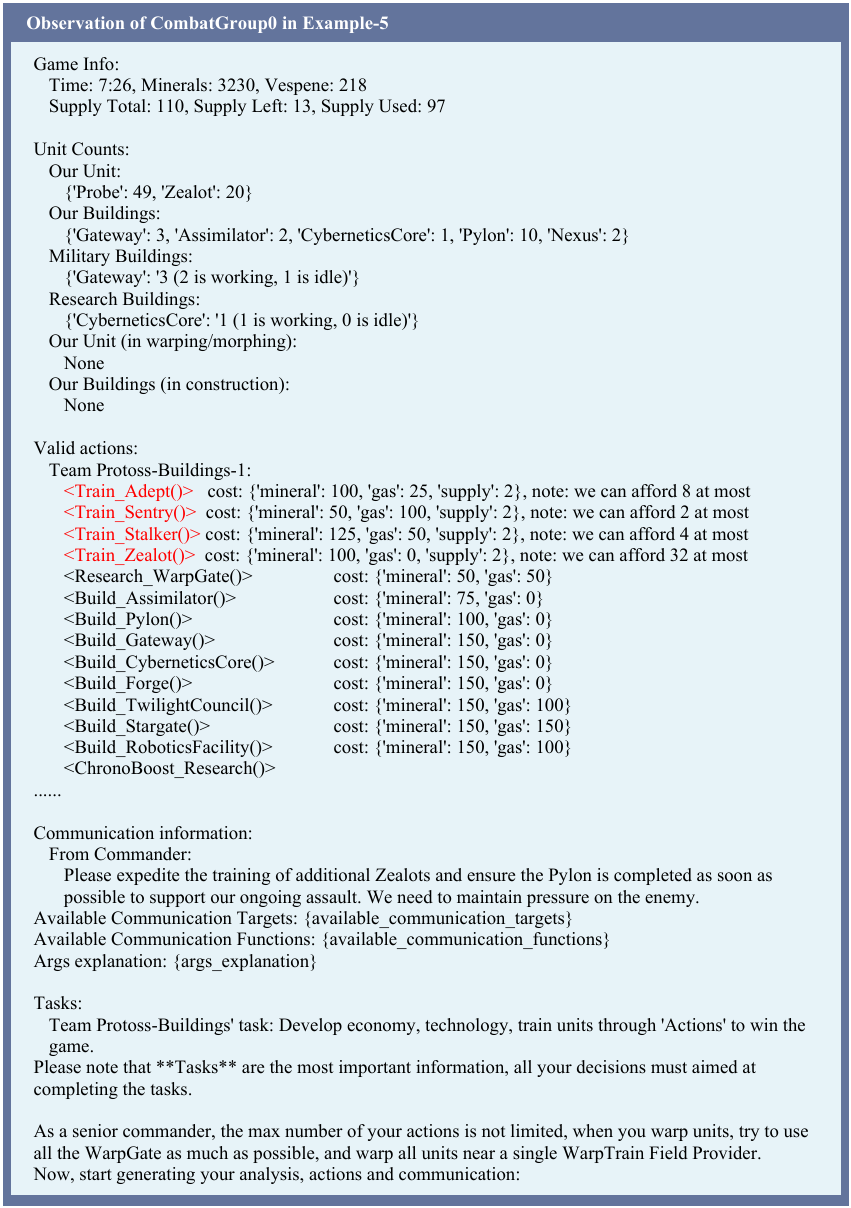}
    \caption{\textbf{Observation of Developer in Example-5.}}
  \label{fig-example-hallucination-e2-p1}
\end{figure}

\begin{figure}[h]
  \centering
  \includegraphics[width=0.95\textwidth]{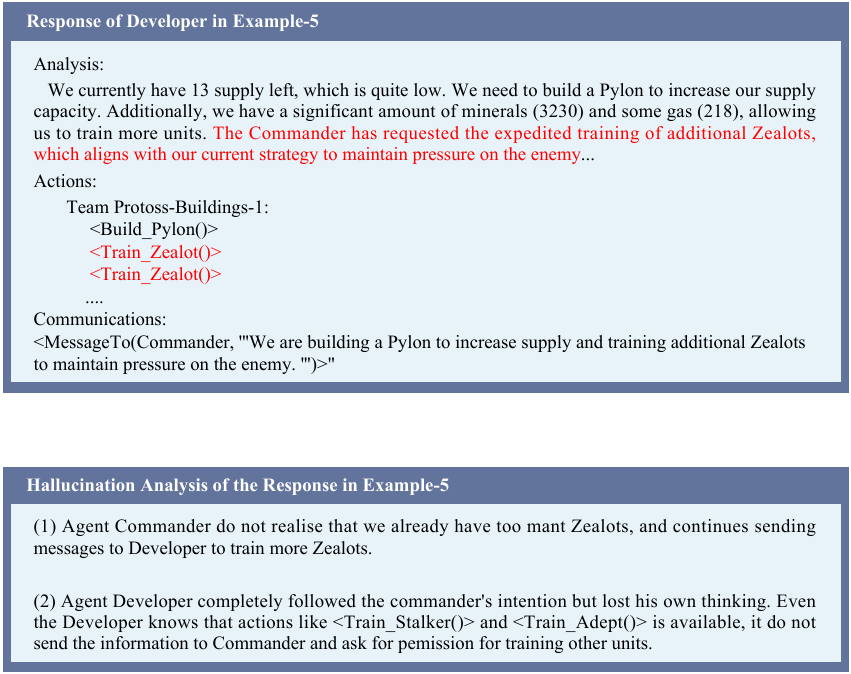}
    \caption{\textbf{Response and Hallucination Analysis of Developer in Example-5}}
  \label{fig-example-hallucination-e2-p2}
\end{figure}

\clearpage
\subsection*{E.3 Hallucination Examples in micro-operation scenarios}

\begin{figure}[h]
  \centering
  \includegraphics[width=0.87\textwidth]{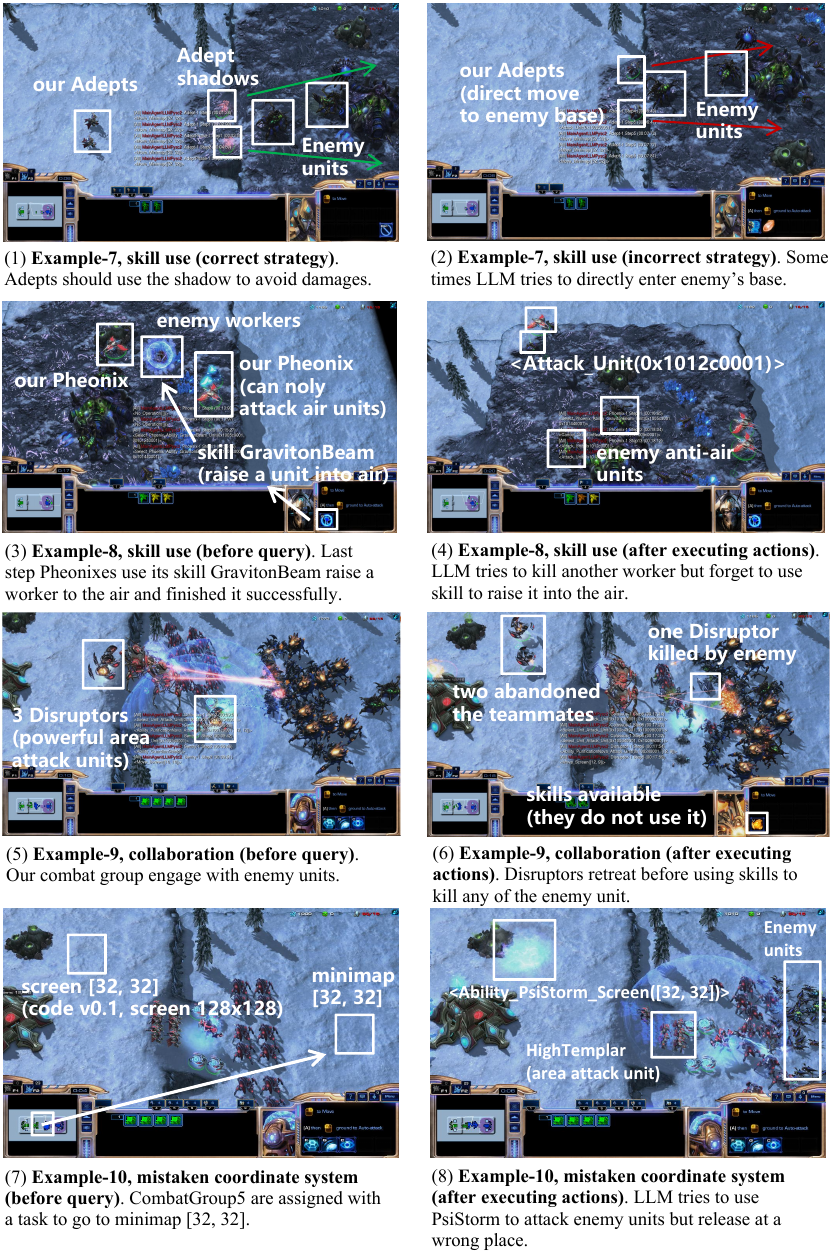}
    \caption{\textbf{Hallucination Examples in micro-operation scenarios}}
  \label{fig-example-hallucination3}
\end{figure}

\begin{figure}[h]
  \centering
  \includegraphics[width=0.95\textwidth]{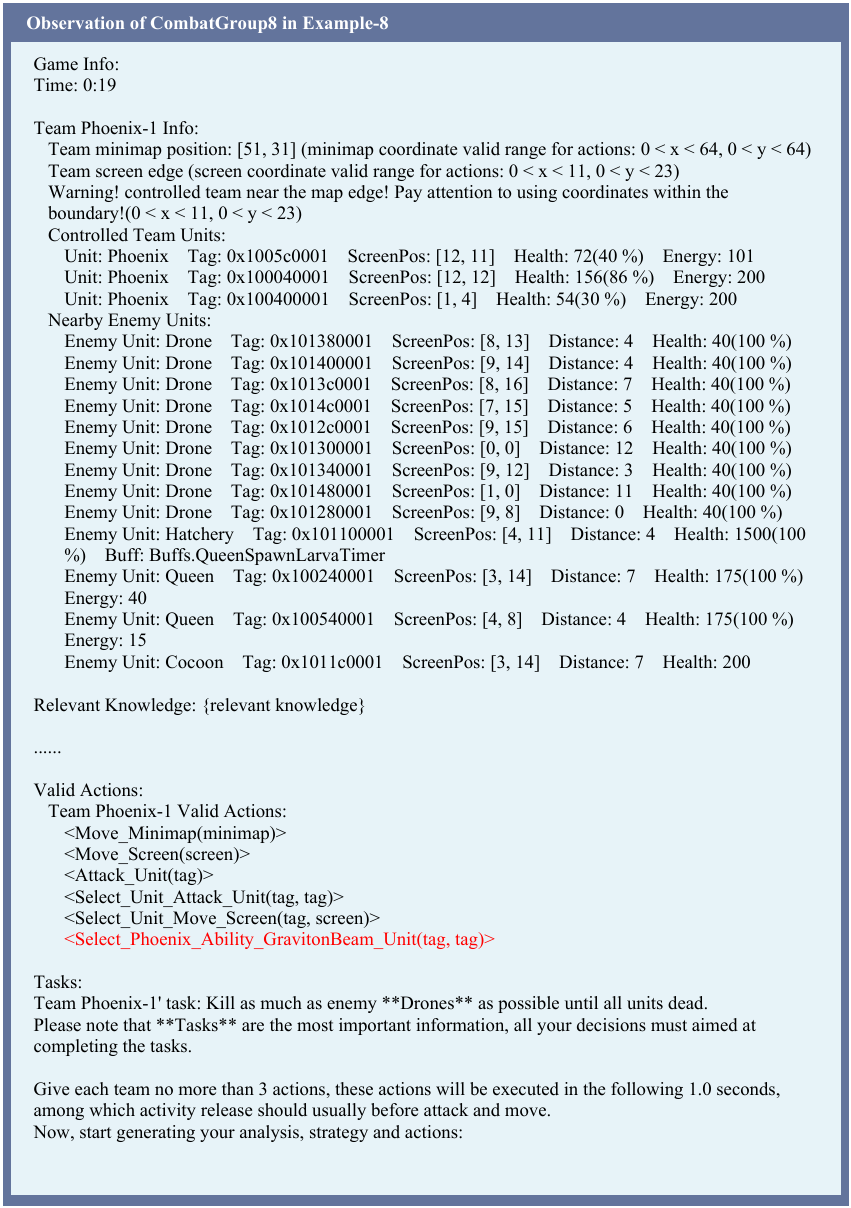}
    \caption{\textbf{Observation of CombatGroup8 in Example-8.}}
  \label{fig-example-hallucination-e3-p1}
\end{figure}

\begin{figure}[h]
  \centering
  \includegraphics[width=0.95\textwidth]{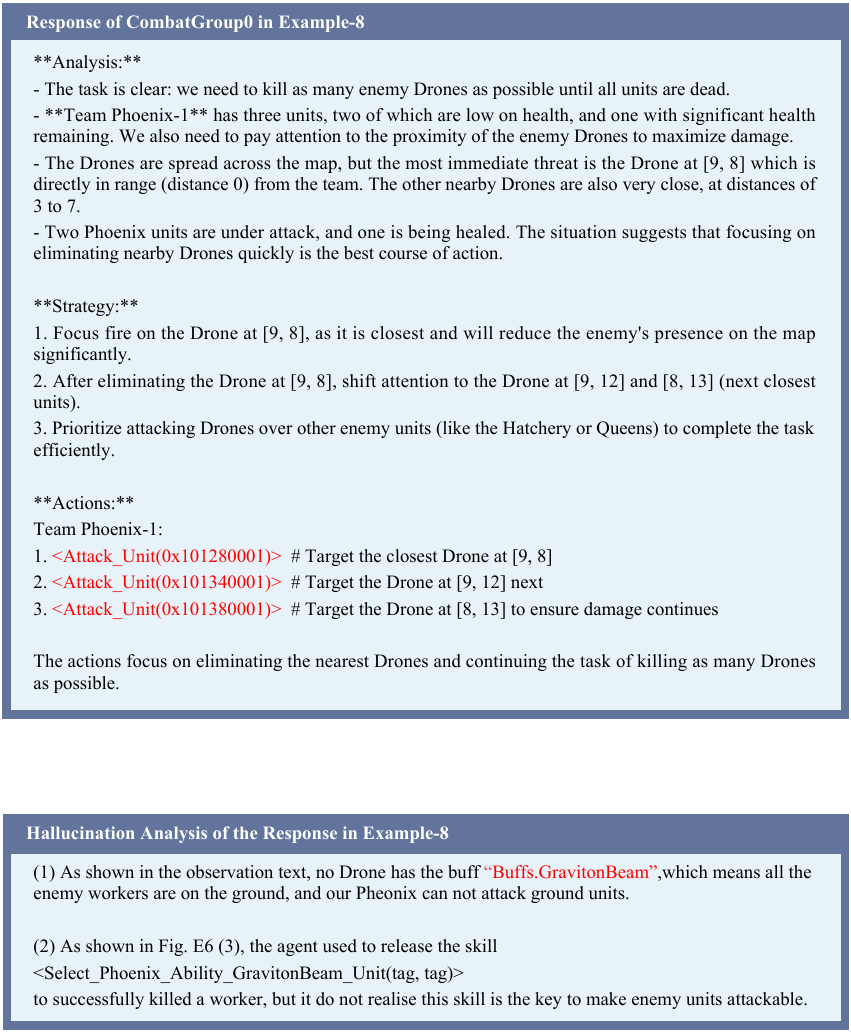}
    \caption{\textbf{Response and Hallucination Analysis of CombatGroup8 in Example-8}}
  \label{fig-example-hallucination-e3-p2}
\end{figure}

\end{document}